\documentclass[10pt, twocolumn]{article}

\usepackage[letterpaper,margin=0.8in]{geometry}
\usepackage{amsmath,amssymb}
\usepackage{bm}
\usepackage{graphicx}
\usepackage[hyphens]{url}
\usepackage[round,authoryear]{natbib}
\usepackage{enumerate}
\usepackage{algpseudocode}
\usepackage[linesnumbered,ruled,vlined,noend]{algorithm2e}
\usepackage{etoolbox}
\usepackage{multirow}
\usepackage{booktabs}
\usepackage{mathtools}
\usepackage{xfrac}
\usepackage{subcaption}
\usepackage{amsthm}
\usepackage{thmtools}
\usepackage{pifont}
\usepackage[hidelinks]{hyperref}

\usepackage{titlesec}

\titleformat{\section}
  {\normalfont\large\bfseries}
  {\thesection}{0.5em}{}

\titleformat{\subsection}
  {\normalfont\normalsize\bfseries}
  {\thesubsection}{0.5em}{}

\titleformat{\subsubsection}
  {\normalfont\normalsize\itshape}
  {\thesubsubsection}{0.5em}{}

\SetInd{0.5em}{0.5em}
\SetArgSty{textrm}
\SetAlgoSkip{}

\makeatletter
\patchcmd{\@algocf@start}{-1.5em}{0pt}{}{}
\def\thmhead@plain#1#2#3{%
  \thmname{#1}\thmnumber{ #2}\thmnote{ \normalfont(#3)}}
\makeatother

\declaretheorem{theorem}
\newtheorem{lemma}[theorem]{Lemma}

\theoremstyle{definition}
\newtheorem{definition}{Definition}

\renewcommand\qedsymbol{$\blacksquare$}
\declaretheoremstyle[
  headfont=\normalfont\itshape\bfseries,
  postheadhook={: },
  qed=\qedsymbol
]{my-proof-style}
\declaretheorem[name={Proof},style=my-proof-style,unnumbered]{myproof}

\allowdisplaybreaks
\newenvironment{sketch_proof}[1]
  {\begin{myproof}[{sketch; full version in #1}]}
  {\end{myproof}}

\newenvironment{funding}{\section*{Funding}\noindent}{\par}

\hypersetup{
  pdftitle={Spatiotemporal and Multi-Robot Motion Planning on Graphs of Space-Time Convex Sets},
  pdfauthor={Jingtao Tang, Zining Mao, Lufan Yang, and Hang Ma}
}

\title{Spatiotemporal and Multi-Robot Motion Planning on Graphs of Space-Time Convex Sets}
\author{
Jingtao Tang\thanks{Corresponding author email: jingtao\_tang@sfu.ca}, Zining Mao, Lufan Yang, and Hang Ma\\
Simon Fraser University, Canada\\
}
\date{}

\begin{document}

\maketitle

\begin{abstract}
Decoupled multi-robot motion planning avoids the high-dimensional joint configuration space but requires repeated single-robot planning among moving obstacles, where route choice and timing are coupled. We develop a framework based on graphs of space-time convex sets (ST-GCSs) for continuous spatiotemporal planning and multi-robot coordination. An ST-GCS represents collision-free space-time by convex sets, allowing the sequence of regions and the geometry and timing of piecewise-linear trajectories to be optimized jointly under velocity limits. To incorporate moving obstacles and previously planned robots, we develop Exact Convex Decomposition (ECD), which removes their swept occupancies while maintaining a convex decomposition for subsequent planning queries. Building on path-based GCS search, we develop a best-first solver tailored to the ST-GCS motion model, exploiting temporal reachability for dominance pruning and using admissible heuristics with separate or joint search-key evaluation. Precomputed heuristic bounds can also be reused across reservation-updated graphs. We integrate ST-GCS planning and ECD into prioritized multi-robot planning and introduce windowed coordination with dynamic window adjustment to limit reservation-induced graph growth. Experiments demonstrate improved scalability and competitive solution quality against optimization-based and sampling-based planners on challenging spatiotemporal and multi-robot problems. Large-scale simulations coordinate up to $100$ robots within a minute, while real-robot demonstrations showcase the practical applicability of the framework.
Project homepage: \url{https://sites.google.com/view/stgcs}.
\end{abstract}


\section{Introduction}

Multi-Robot Motion Planning (MRMP) aims to compute collision-free trajectories that move multiple robots from their starts to goals in a shared environment. A direct approach plans in the joint configuration space of all robots, but its dimensionality and the complexity of representing collision-free configurations grow rapidly with the number of robots. Decoupled approaches such as Prioritized Planning (PP)~\citep{erdmann1987multiple} and Priority-Based Search (PBS)~\citep{ma2019searching} instead coordinate single-robot plans through priority decisions. In prioritized planning, higher-priority trajectories are treated as moving obstacles for lower-priority robots. This avoids direct joint-space planning but requires repeatedly solving single-robot motion-planning problems in which feasibility depends on both where and when a robot moves.
This decoupled approach has been particularly successful in Multi-Agent Path Finding (MAPF)~\citep{stern2019multi} on discrete graphs. For example, Cooperative A$^*$ extends spatial A$^*$ to space-time using reservations of previously planned agents~\citep{silver2005cooperative}. Similar principles underlie methods that separate high-level coordination from low-level space-time planning. Extending them to continuous MRMP is more involved: geometric paths, velocities, and timing must be determined jointly, without prescribed graph transitions or discrete time steps, while moving obstacles can open or close passages for short intervals. Sampling-based spatiotemporal planners~\citep{huppi2022t,grothe2022st} offer modeling flexibility, but can struggle with such transient feasible regions and require repeated collision checks in space-time.

Graphs of Convex Sets (GCSs) provide another approach by representing collision-free regions as convex sets and jointly optimizing the selected set sequence and the trajectory within them~\citep{marcucci2023motion}. 
Joint configuration-space GCSs can account for inter-robot collisions and have been used for fixed multi-arm systems, including dual-arm planning~\citep{marcucci2023motion}, multi-arm assembly~\citep{natarajan2024ixg}, and temporal-logic task and motion planning~\citep{wei2025hierarchical}. 
While such decompositions can be reused for a fixed robot team, our setting considers varying numbers of mobile robots sharing the same environment. A spatial GCS can be precomputed once and reused independently of team size, whereas a joint-space GCS depends on the number of robots and their inter-robot collision constraints, with dimensionality growing with team size. It therefore cannot serve as a single precomputed representation across arbitrary team sizes. This motivates a PP/PBS-based architecture, as studied in MRMP~\citep{erdmann1987multiple,van2005prioritized,vcap2015prioritized} and MAPF~\citep{silver2005cooperative,ma2019searching}, to coordinate robot trajectories while reusing a precomputed spatial GCS across different team sizes.

In this work, we develop a decoupled MRMP framework based on a \textit{Graph of Space-Time Convex Sets} (ST-GCS), extending spatial GCSs with time. Individual planning queries incorporate selected trajectories of other robots as reserved space-time occupancies. For $n$ robots with $m$-dimensional positions, each low-level problem operates in $(m+1)$-dimensional space-time rather than the $(nm)$-dimensional joint configuration space, with inter-robot coordination handled separately.

\subsection{Space-Time Representation and Reservation}

An ST-GCS represents collision-free space-time by convex sets connected through nonempty intersections. It can be initialized by extruding a spatial convex decomposition along time and then excluding known moving-obstacle occupancies. A trajectory consists of a path through the ST-GCS and piecewise-linear motions within its selected sets. With time monotonicity and component-wise velocity limits, the geometric route, transition positions and times, and goal arrival time can be optimized jointly without discretizing time. We require the robot to remain safely at its goal after arrival through the planning horizon, allowing its subsequent occupancy to be considered during coordination.

To incorporate moving objects and planned robots, we introduce \textit{Exact Convex Decomposition} (ECD). Under our piecewise-linear motion and translational hypercube occupancy model, a trajectory's swept occupancy is a finite union of space-time parallelotopes. ECD removes these occupancies by subdividing affected convex sets and updating their adjacency, preserving a convex decomposition of the remaining collision-free space-time. Thus, a trajectory produced by one planning query can be reserved in the ST-GCS used by subsequent queries. The construction applies to arbitrary-dimensional Euclidean position spaces under the stated occupancy model.

\begin{figure}[t]
\centering
\includegraphics[width=\linewidth]{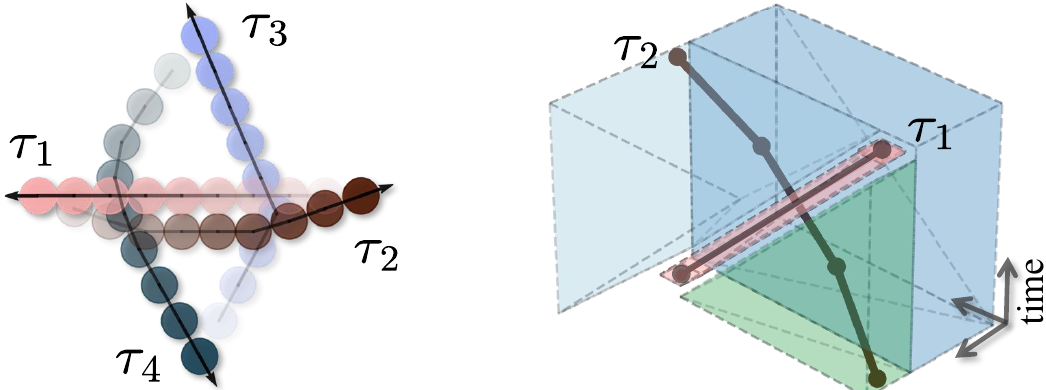}
\caption{MRMP with ST-GCS: robots $1$ and $3$ exchange positions with $2$ and $4$, respectively. Left: Spatial trajectories, with increasing transparency indicating later times. Right: Trajectories $\tau_1$ and $\tau_2$ in space-time, where $\tau_2$ is planned through convex sets (colored polyhedra) excluding $\tau_1$'s reserved occupancy.}
\label{fig:mrmp}
\end{figure}

\subsection{Search on ST-GCSs}

Solving a spatiotemporal planning problem on ST-GCS as a mixed-integer convex program can be expensive, particularly for repeated queries on ECD-updated graphs. IxG$^*$~\citep{natarajan2024ixg} and GCS$^*$~\citep{chia2024gcs} address large GCS problems by searching over partial paths of convex sets and optimizing continuous trajectories along them. Unlike ordinary shortest-path search, multiple paths reaching the same vertex may need to be retained because they induce different feasible continuous states and continuation costs. We build on this path-based search approach to develop a best-first search (BFS) solver for time-optimal planning on ST-GCSs. Tab.~\ref{tab:gcs_search_comparison} compares its main design choices with those of IxG$^*$ and GCS$^*$.

Time introduces not only additional constraints but also useful structure. For a fixed start time, the elapsed-time cost of reaching a space-time state is determined by its time coordinate. Under our velocity-bounded piecewise-linear motion model, convex space-time sets also support straight-line connections characterized by forward reachability cones. These properties allow us to exclude revisitation of the same space-time convex set and develop a sound set-containment dominance check, together with approximate arrival-state- and position-based checks for faster pruning. 
We combine them with admissible heuristics and optional incumbent upper-bound pruning, retaining bounded-suboptimality guarantees under sound dominance pruning.

For search guidance, we consider separate and joint evaluation of the prefix cost $g$ and admissible cost-to-go bounds $h$. Separate $g+\epsilon h$ evaluation (with heuristic inflation factor $\epsilon\geq 1$), as in IxG$^*$, accommodates motion-only ($h_{\mathrm{mot}}$), triplet-relaxation ($h_{\mathrm{tri}}$), and interface-to-set ($h_{\mathrm{tab}}$) heuristics, which can be combined through their maximum $h_{\mathrm{max}}$. For the motion-only bound, joint $g+\epsilon h_{\mathrm{mot}}$ evaluation, as in GCS$^*$, can tighten the search key by optimizing the prefix and heuristic over a shared interface state, at the cost of an additional convex optimization. The triplet-relaxation and interface-to-set heuristics exploit information precomputed on a base ST-GCS and reused across ECD-updated queries, which is particularly useful for repeated low-level planning in MRMP.

\subsection{Multi-Robot Coordination}

We combine ST-GCS planning and ECD into PP~\citep{erdmann1987multiple} and PBS~\citep{ma2019searching}. PP plans robots in a fixed priority order, while PBS searches over pairwise priority constraints. Both use ECD to reserve higher-priority trajectories in the ST-GCS for each lower-priority query, allowing the same low-level solver to optimize individual routes and timing.

Full-horizon reservations may substantially enlarge ST-GCSs by introducing occupancy-induced subdivisions. Inspired by windowed MAPF methods~\citep{silver2005cooperative,li2021lifelong}, we develop windowed coordination to restrict reservations and conflict resolution to a finite planning window while each low-level query still seeks the original goal. Only an execution prefix, possibly shorter than the planning window, is committed before replanning from the resulting space-time states. 
This limits reservation-induced graph growth, while dynamic window adjustment exposes longer-range interactions when coordination fails or makes insufficient progress.

\subsection{Contributions and Scope}

Our formulation considers translational motion in Euclidean position spaces, with piecewise-linear trajectories, component-wise velocity limits, and translational hypercube occupancies. These assumptions enable convex path-conditioned optimization, ECD under the adopted occupancy model, and our reachability analysis. More general nonholonomic or kinodynamic constraints, including acceleration and orientation dynamics, are not covered by these guarantees. The low-level search bounds do not imply global MRMP optimality, and finite-window coordination does not guarantee completion in general; trajectory postprocessing is also outside the stated planning guarantees.

We summarize our main contributions as follows:
\begin{enumerate}
    \item We develop the ST-GCS framework for time-optimal spatiotemporal motion planning, jointly optimizing the sequence of collision-free space-time convex sets and the geometry and timing of piecewise-linear trajectories under velocity limits.

    \item We present ECD for reserving any piecewise-linear trajectory with a hypercube occupancy model in an arbitrary-dimensional space-time, enabling unified treatment of moving obstacles and inter-robot interactions in repeated queries.

    \item Based on path-based GCS search, we develop a BFS solver specialized to time-optimal ST-GCS planning. We establish set-revisitation exclusion, design sound and approximate dominance checks, and support complementary admissible heuristics with separate or joint search-key evaluation.

    \item We integrate ST-GCS planning and ECD into prioritized MRMP methods and develop windowed coordination with dynamic adjustment to limit reservation-induced graph growth while retaining continuous route and timing optimization.

    \item We evaluate the planning and coordination components through extensive numerical benchmarks against optimization-based and sampling-based planners, large-scale simulations with up to $100$ robots, and real-robot demonstrations with trajectory-optimization postprocessing.
\end{enumerate}

This work substantially extends our conference paper~\citep{tang2025stgcs}, which introduced ST-GCS, the original ECD construction for 2D spatial motion with time, and prioritized MRMP integration. The journal version develops path-based BFS with ST-GCS-specific dominance checks, heuristics, and theoretical analysis, generalizes ECD to arbitrary-dimensional space-time, introduces windowed coordination and dynamic window adjustment, and substantially expands the empirical evaluation with additional ablations, large-scale scenarios, and real-robot experiments.

\section{Related Work}
\label{sec:related_work}

This section positions our work relative to spatiotemporal planning, MRMP, and planning on GCSs.

\subsection{Spatiotemporal Motion Planning}
Spatiotemporal motion planning treats time as part of the planning domain to handle moving obstacles with known trajectories~\citep{erdmann1987multiple}.
Grid-based methods reserve discrete space-time states~\citep{silver2005cooperative}; Safe Interval Path Planning (SIPP) reasons over safe time intervals for discrete spatial occupancies~\citep{phillips2011sipp}; any-angle variants such as Zeta$^*$-SIPP combine safe intervals with any-angle search for time-optimal planning in dynamic environments~\citep{zou2024zeta}; and state-lattice methods plan over spatiotemporal lattices built from continuous space and fixed motion primitives~\citep{mcnaughton2011motion}.
Sampling-based planners extend Rapidly-exploring Random Tree (RRT), Probabilistic Roadmap (PRM), and RRT$^*$ methods to configuration--time space, validating sampled time-parameterized states or edges against moving obstacles over their execution intervals~\citep{sintov2014time,huppi2022t,grothe2022st}; they are flexible, but can struggle when feasible regions are narrow, short-lived, or repeatedly changing, and their runtime can be dominated by collision checking in the time-augmented state space.
Recent work mitigates this issue by bringing safe-interval ideas from MAPF~\citep{phillips2011sipp} into sampling-based planning~\citep{sim2024safe, kerimov2025safe}.
In contrast, we plan on ST-GCSs: collision-free space-time is represented by convex sets connected through nonempty intersections, so reserved occupancies can be removed from an updatable free-space decomposition while continuous trajectory optimization remains available over remaining regions.

\subsection{MRMP}

MRMP methods differ mainly in how they represent and resolve inter-robot conflicts.
Coupled methods reason in composite configuration spaces. 
Coordinated PRMs explicitly construct products of individual roadmaps~\citep{vsvestka1998coordinated}, dRRT-style methods explore implicit tensor-product roadmaps without materializing the full joint graph~\citep{solovey2016finding,shome2020drrt}, and partially coupled methods such as M$^*$ plan independently when possible while expanding into higher-dimensional coupled search near conflicts~\citep{wagner2015subdimensional}.
Decoupled and conflict-based methods improve scalability by using priorities, reservations, or constraints. 
Prioritized planning treats planned robots as moving obstacles in configuration--time space~\citep{erdmann1987multiple,van2005prioritized}, cooperative pathfinding uses discrete space-time reservations~\citep{silver2005cooperative}, Conflict-Based Search (CBS) separates high-level conflict resolution from low-level planning~\citep{sharon2015conflict}, and PBS searches over priority constraints rather than a fixed ordering~\citep{ma2019searching}.
These ideas underlie MAPF variants including continuous-time MAPF~\citep{andreychuk2022multi}, safe-interval planning on continuous-time roadmaps~\citep{kasaura2022psipp}, and representation-optimal MRMP for heterogeneous robots in continuous spaces~\citep{solis2021representation}.

Recent MRMP work connects conflict-based coordination with sampling-based, kinodynamic, optimization, and control-based planning.
Kinodynamic CBS incorporates motion primitives into CBS~\citep{kottinger2022conflict}, while adaptive methods vary coupling online through adaptive robot coordination~\citep{solis2024adaptive}, kinodynamic adaptive coordination~\citep{qin2025k}, or guidance-informed grouping~\citep{mcbeth2026scalable}.
Other approaches use decentralized trajectory optimization~\citep{tordesillas2021mader}, path retiming~\citep{mao2024collision}, CBS-guided Model Predictive Control (MPC)~\citep{tajbakhsh2024conflict}, or mixed-integer continuous formulations~\citep{zhao2025mixed,ren2025cp}.
Our MRMP planner is closest to prioritized planning, but each low-level query plans on an ST-GCS; after a robot is planned, ECD updates the ST-GCS to reserve its swept occupancy, so later robots plan in the remaining collision-free space-time and jointly reason about route choice, timing, and collision avoidance.

\subsection{Planning on GCSs}

GCSs combine graph structure with convex optimization for mixed discrete--continuous planning~\citep{marcucci2023motion,marcucci2024shortest}.
They have been applied beyond cluttered Euclidean motion planning to non-Euclidean configuration spaces~\citep{cohn2023non}, temporal-logic and precedence-constrained planning~\citep{kurtz2023temporal,you2025motion}, contact-rich manipulation~\citep{graesdal2024towards}, and guidance for downstream nonconvex trajectory optimization~\citep{von2024using}.
These works demonstrate the breadth of GCS modeling, but mainly consider static, task-augmented, or contact-mode domains rather than a free-space representation repeatedly updated by moving obstacles and planned robot trajectories.

GCS performance depends strongly on the convex decomposition of free space.
Iterative Regional Inflation by Semidefinite programming (IRIS) grows large obstacle-free convex regions~\citep{deits2015computing}; visibility-graph clique covers and certified polyhedral decompositions improve coverage, certification, and scalability~\citep{werner2024approximating,dai2024certified}; and GPU-accelerated methods compute collision-free configuration-space convex sets online in changing environments~\citep{werner2025superfast}.
Our ECD scheme is complementary. It updates an ST-GCS by removing swept occupancies while maintaining a convex decomposition of the remaining collision-free space-time.

Efficient GCS solving has also received substantial attention.
Large safe-box planners exploit offline preprocessing with fast runtime shortest-path and convex-control subproblems~\citep{marcucci2024fast}.
A$^*$-GCS~\citep{sundar2024graphs} incrementally expands subgraphs and solves convex relaxations to obtain lower bounds on solution costs.
In contrast, IxG$^*$~\citep{natarajan2024ixg} and GCS$^*$~\citep{chia2024gcs} search over partial paths of convex sets, evaluating them through continuous trajectory optimization.
Our BFS solver builds on this path-based search framework, with dominance checks and heuristics tailored to time-optimal spatiotemporal planning.
Other variants address multi-query planning~\citep{morozov2024multiquery}, shortest walks with repeated vertices~\citep{morozov2025shortestwalks}, nonconvex-cost~\citep{clark2025nonconvexgcs} or parametrized-space objectives~\citep{garg2025undistort}, fixed-sequence minimum-time motion through convex sets~\citep{marcucci2025biconvex}, and routing problems that couple discrete visitation order with continuous trajectories~\citep{philip2024mixed,bhat2025complete,tang2026ghost}.
Closest to our setting, \cite{osburn2025stgcs} studies ST-GCS planning in dynamic environments and constructs GCS-compatible constraints, while \cite{zhao2025cbgcs} combines CBS with time-augmented GCS under a fixed time-step representation.
Recent GCS work has expanded modeling scope and solver efficiency, but has not jointly addressed path-indexed search on ST-GCSs, occupancy reservation by convex-decomposition updates, and windowed multi-robot coordination.

\section{Preliminaries}\label{sec:preliminaries}
This section introduces the notation and basic GCS formulation used in this paper. We first review the standard definition of (spatial) GCSs, where vertices correspond to collision-free convex sets and edges indicate nonempty intersections. We then describe the standard motion-planning problem on GCSs and the path-conditioned convex subproblem used later by our search-based solver for planning on ST-GCSs. Vector inequalities $\preceq$ and $\succeq$ are interpreted component-wise.

\subsection{GCS Definition}
We consider a robot with an $m$-dimensional state space, whose collision-free region is decomposed into a collection of convex sets. These spatial convex sets need not be disjoint. A GCS, denoted as $G=(V, E, \mathcal{X})$, is a connected graph representing such a convex decomposition, where $V$ is the vertex set, $E$ is the edge set, and $\mathcal{X}=\{X_v\}_{v\in V}$ is the collection of convex sets. Each vertex $v\in V$ corresponds to a convex set $X_v =\{\mathbf{x}\,|\,\mathbf{A}_v\mathbf{x}\preceq \mathbf{b}_v\}\subseteq\mathbb{R}^m$ of states. Each edge $e=(u,v)\in E$ indicates that $X_u\cap X_v\neq\emptyset$. 
Our use of GCS is slightly more restrictive than the original formulation of~\cite{marcucci2024shortest}, where constraints may be attached more generally to vertices and edges.

A \textit{path} on $G$ is an ordered sequence of vertices $\pi=\langle v_1,v_2,\ldots,v_{l} \rangle$ where $(v_{i-1},v_i)\in E$ for all $i=2,\ldots,l$. We write $E(\pi)=\{(v_{i-1},v_i)\}_{i=2}^{l}$ for its consecutive edges and use $\oplus$ for sequence concatenation. Unless otherwise stated, we consider simple paths, i.e., paths without repeated vertices. A \textit{trajectory} associated with $\pi$ is a continuous curve obtained by concatenating local trajectory segments inside the convex sets along $\pi$.

\subsection{Motion Planning on GCSs}

Given a start state $\mathbf{x}_s$ and a goal state $\mathbf{x}_g$, motion planning on a GCS $G=(V,E,\mathcal{X})$ asks for a trajectory from $\mathbf{x}_s$ to $\mathbf{x}_g$.
Since the start and goal states may lie in multiple convex sets, we introduce auxiliary source and target vertices $v_s$ and $v_g$, with $X_{v_s}=\{\mathbf{x}_s\}$ and $X_{v_g}=\{\mathbf{x}_g\}$, connected to all start- and goal-containing vertices, respectively, and use these auxiliary vertices as the unique endpoints of the graph path.
We introduce two sets of variables. 
The binary edge-selection variables $\Phi=\{\phi_e\}_{e\in E}$ parameterize a path $\pi_\Phi$, where $\phi_e=1$ if and only if edge $e$ is selected by the path $\pi_\Phi$. The continuous variables $\mathbf{x}_{v}, \mathbf{y}_{v}\in X_v$ represent the entry and exit states of the local trajectory segment inside $X_v$. For each vertex $v\in V$, let $\ell_v(\mathbf{x}_v,\mathbf{y}_v)$ denote a given local cost associated with vertex $v$. 
The motion planning problem on GCS can be written as:
\begin{subequations}
\begin{align}
  \min_{\Phi,\mathbf{x},\mathbf{y}}\;
  & \sum_{v\in \pi_\Phi} \ell_v(\mathbf{x}_v,\mathbf{y}_v)\label{eqn:gcs:obj}\\
  \textbf{s.t.}\;
  & \pi_\Phi = \langle v_s,\ldots,v_g \rangle \in \mathcal{P}(v_s, v_g;G),\label{eqn:gcs:path}\\
  &\mathbf{x}_v, \mathbf{y}_v\in X_v,&\makebox[36pt][r]{$\forall v\in  \pi_\Phi,$}\label{eqn:gcs:in_set}\\
  &\mathbf{x}_v=\mathbf{y}_u,&\makebox[36pt][r]{$\forall (u,v)\in E(\pi_\Phi),$}\label{eqn:gcs:continuity}\\
  &\mathbf{x}_{v_s}=\mathbf{x}_s\text{ and }\mathbf{y}_{v_g}=\mathbf{x}_g,\label{eqn:gcs:st_condition}
\end{align}
\end{subequations}
where $\mathcal{P}(v_s, v_g;G)$ denotes the set of simple paths in $G$ that start from $v_s$ and end at $v_g$.

The objective in Eqn.~(\ref{eqn:gcs:obj}) sums the local costs of the trajectory segments $\mathbf{x}_v \rightarrow \mathbf{y}_v$ along the selected path. Eqn.~(\ref{eqn:gcs:path}) enforces that the selected edges form a valid simple path from the auxiliary source vertex to the auxiliary target vertex. This path constraint can be encoded using standard flow-conservation and degree constraints for routing problems~\citep{miller1960integer,bertsimas1997introduction}, yielding a Mixed-Integer Convex Program (MICP). Eqn.~(\ref{eqn:gcs:in_set}) enforces that each local segment lies inside the corresponding convex set. Since $X_v$ is convex, the straight-line segment $(\mathbf{x}_v,\mathbf{y}_v)$ is collision-free. Eqn.~(\ref{eqn:gcs:continuity}) enforces continuity between consecutive local segments for every selected edge. Eqn.~(\ref{eqn:gcs:st_condition}) enforces the trajectory starts from and ends at the given start state and goal state, respectively.

\begin{figure}[t]
  \centering
  \includegraphics[width=0.92\linewidth]{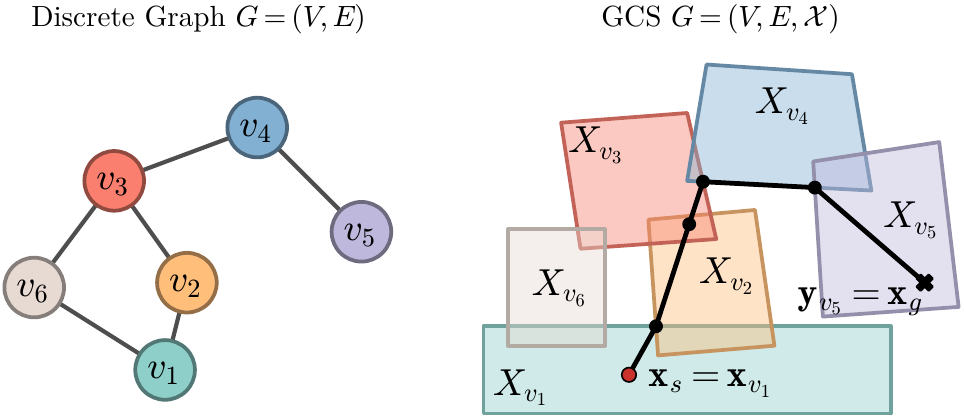}
  \caption{A motion planning problem on a GCS, where each vertex denotes a collision-free convex set and a feasible solution reconstructs a continuous piecewise-linear trajectory from $\mathbf{x}_s$ to $\mathbf{x}_g$ along a selected graph path.}
  \label{fig:gcs}
\end{figure}

As illustrated in Fig.~\ref{fig:gcs}, a feasible solution reconstructs a continuous, collision-free, piecewise-linear trajectory from $\mathbf{x}_s$ to $\mathbf{x}_g$ by chaining the local segments $\mathbf{x}_v \rightarrow\mathbf{y}_v$ along $v\in\pi_\Phi$. The above formulation focuses on kinematic feasibility, which aligns with the standard convention in the literature. Differential or kinodynamic constraints can be incorporated by augmenting the state space and adding suitable constraints, such as in~\cite{marcucci2023motion}. 
The formulation can also be adapted to a convex start set $X_s$ or a convex goal set $X_g$ by replacing the point constraints in Eqn.~(\ref{eqn:gcs:st_condition}) with set-membership constraints.

\subsection{Path-Conditioned Optimization}\label{subsec:convex_restriction}

Solving Eqns.~(\ref{eqn:gcs:obj}--\ref{eqn:gcs:st_condition}) directly as an MICP can be computationally expensive \citep{marcucci2024shortest}. A useful subproblem is obtained by fixing the graph path and optimizing only the continuous variables along that path. This operation is commonly referred to as \textit{convex restriction}~\citep{diamond2018general} and is used as a subroutine in several search-based solvers for graph optimization problems on GCSs~\citep{marcucci2024graphs,natarajan2024ixg,tang2026ghost}.

Fixing a graph path determines which convex sets are visited and in what order. The entry and exit states remain continuous optimization variables, which are optimized jointly subject to set-membership and continuity constraints.
Given a path $\pi=\langle v_1,v_2,\ldots,v_l \rangle$ on $G$, we say that a trajectory $\tau$ is \emph{conditioned on} $\pi$ if it is represented by an ordered sequence $\tau=\langle(\mathbf{x}_{v_i},\mathbf{y}_{v_i})\rangle_{i=1}^l$ of entry and exit states, where $\mathbf{x}_{v_i},\mathbf{y}_{v_i}\in X_{v_i}$ with continuity enforced between consecutive sets. Conditioning on a path $\pi$ fixes the binary variables $\Phi$ so that $\pi_\Phi = \pi$. 
The remaining optimization is the convex program over the continuous entry and exit states:
\begin{subequations}
\begin{align}
  \min_{\mathbf{x},\mathbf{y}}\quad
  & \sum_{i=1}^l \ell_{v_i}(\mathbf{x}_{v_i},\mathbf{y}_{v_i})\\
  \textbf{s.t.}\quad
  &\mathbf{x}_{v_i},\mathbf{y}_{v_i}\in X_{v_i},&i=1,\ldots,l,\\
  &\mathbf{x}_{v_i}=\mathbf{y}_{v_{i-1}},&i=2,\ldots,l,\\
  &\mathbf{x}_{v_1}=\mathbf{x}_s\text{ and }\mathbf{y}_{v_l}=\mathbf{x}_g.
\end{align}
\end{subequations}
This is a convex program that off-the-shelf optimizers can efficiently solve whenever the local costs and set constraints are convex. If feasible, its solution defines the optimal trajectory conditioned on the fixed path $\pi$. For example, in Fig.~\ref{fig:gcs}, fixing $\pi=\langle v_1,v_2,v_3,v_4,v_5 \rangle$ removes the discrete path-selection variables and optimizes only the entry and exit state variables along that path.

In practice, we merge adjacent exit--entry variables along a fixed path, which eliminates redundant variables and continuity equalities to improve solving efficiency.

\section{Graph of Space-Time Convex Sets}\label{sec:stgcs}
We extend GCS described in Sec.~\ref{sec:preliminaries} to spatiotemporal motion planning. The key concept is a Graph of Space-Time Convex Sets (ST-GCS), where each convex set lies in the joint space of position and time.
Planning on an ST-GCS allows dynamic-obstacle avoidance, variable arrival times, velocity limits, and time optimality to be handled within a unified graph-optimization problem.

\begin{figure}[t]
    \centering
    \includegraphics[width=\linewidth]{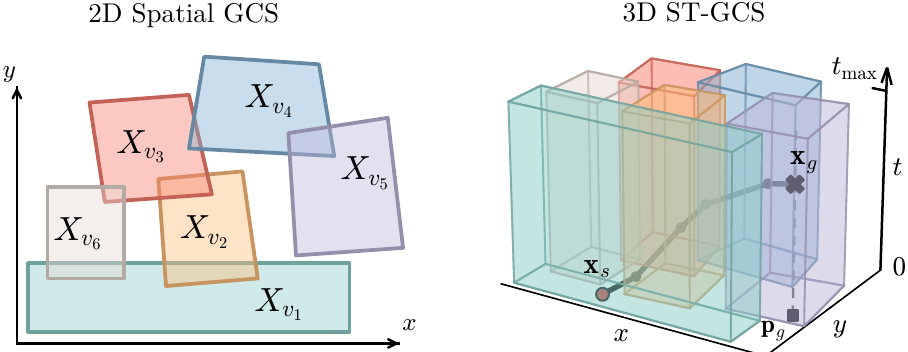}
    \caption{A 3D ST-GCS (right) constructed by extruding each set of a 2D spatial GCS (left) along the time dimension $t \in [0, t_{\max}]$. A feasible trajectory is planned from the start state $\mathbf{x}_s$ to the goal position $\mathbf{p}_g$, with the dashed vertical line indicating the possible goal states $\mathbf{x}_g=(\mathbf{p}_g,t)$ across time.}
    \label{fig:stgcs}
\end{figure}

Let $\mathbf{x}=(\mathbf{p},t)\in\mathbb{R}^{m+1}$ denote a space-time state, where $\mathbf{x}.\mathbf{p}\in\mathbb{R}^m$ is the spatial position and $\mathbf{x}.t\in\mathbb{R}$ is the time.
An input ST-GCS $G=(V,E,\mathcal{X})$ associates each vertex $v\in V$ with a convex set $X_v\subset\mathbb{R}^{m+1}$ of states that are collision-free from both static and dynamic obstacles, where $\mathcal{X}=\{X_v\}_{v\in V}$. Each edge $(u,v)\in E$ indicates that $X_u\cap X_v\neq\emptyset$.
As shown in Fig.~\ref{fig:stgcs}, an ST-GCS can be initialized by extruding each given spatial collision-free convex set from time $0$ to an arbitrarily large but finite time limit $t_{\text{max}}$.
The ECD scheme introduced in Sec.~\ref{sec:ecd} can further remove the space-time occupancies of dynamic obstacles or previously planned robots, potentially subdividing the extruded sets.

\subsection{Trajectory Representation on ST-GCSs}\label{subsec:st_trajectory_representation}
A trajectory $\tau$ conditioned on a vertex path $\pi$ on ST-GCS $G=(V,E,\mathcal{X})$ is represented by concatenating straight space-time segments along that path. For each occurrence of a vertex $v$ in $\pi$, the local segment $\mathbf{x}_v\rightarrow\mathbf{y}_v$ joins entry and exit states in $X_v$. 
Convexity ensures that the entire segment lies in $X_v$, and continuity between consecutive segments gives $\mathbf{y}_u=\mathbf{x}_v$ for each transition $(u,v)$ along $\pi$.
Each segment has constant spatial velocity, which may change between segments. 
We write $\Delta t_v=\mathbf{y}_v.t-\mathbf{x}_v.t$ and $\Delta\mathbf{p}_v=\mathbf{y}_v.\mathbf{p}-\mathbf{x}_v.\mathbf{p}$ for its elapsed time and spatial displacement, respectively.

The local motion constraints enforce time monotonicity and a component-wise velocity bound $\mathbf{v}_{\mathrm{lim}}\in\mathbb{R}^{m}_{>0}$:
\begin{align}\label{eqn:st_local_motion}
\Delta t_v\geq 0\, \text{ and }\,
-\mathbf{v}_{\mathrm{lim}}\Delta t_v\preceq\Delta\mathbf{p}_v\preceq\mathbf{v}_{\mathrm{lim}}\Delta t_v,
\end{align}
which permit waiting as a vertical space-time segment with $\Delta\mathbf{p}_v=\mathbf{0}$ and $\Delta t_v>0$ when its endpoints lie in $X_v$.
Optionally, an isotropic $L_2$ speed limit $v_{\max}$ can be imposed using the second-order cone constraint $\|\Delta\mathbf{p}_v\|_2\leq v_{\max}\Delta t_v$, which preserves convexity.

We define the trajectory cost as its total elapsed time:
\begin{align}\label{eqn:traj_cost}
c(\tau)=\sum_{v\in\pi}\Delta t_v.
\end{align}

This formulation specializes GCS to translational planning in Euclidean position--time space.
The chosen trajectory model keeps optimization along candidate graph paths convex and supports geometric reasoning to reduce the search space.
Together with the hypercube collision model, it also yields polyhedral trajectory occupancies, enabling exact convex decomposition of the remaining free space after trajectory reservation.

\begin{table*}[t]
\centering
\caption{Comparison of search-based solvers for motion planning problems on (ST-)GCSs, which interleave discrete search over convex set sequences with continuous trajectories optimizations.}
\label{tab:gcs_search_comparison}
\small
\setlength{\tabcolsep}{4.5pt}
\setlength{\aboverulesep}{0pt}
\setlength{\belowrulesep}{0pt}
\renewcommand{\arraystretch}{1.2}
\begin{tabular}{@{}|c|c|c|c|@{}}
\toprule
\textbf{}
& \textbf{IxG$^*$}\citep{natarajan2024ixg}
& \textbf{GCS$^*$}\citep{chia2024gcs}
& \textbf{BFS Solver on ST-GCS (Ours)} \tabularnewline
\midrule
\begin{tabular}[c]{@{}c@{}}\textbf{Search-key}\\\textbf{Evaluation}\end{tabular}
& \begin{tabular}[c]{@{}c@{}}Separate prefix cost\\and heuristic\end{tabular}
& \begin{tabular}[c]{@{}c@{}}Joint prefix--heuristic\\optimization\end{tabular}
& \begin{tabular}[c]{@{}c@{}}Separate evaluation; optional joint\\optimization with a motion-only bound\end{tabular} \\
\midrule[0.25pt]

\begin{tabular}[c]{@{}c@{}}\textbf{Set}\\\textbf{Revisitation}\end{tabular}
& Allowed
& Allowed
& \begin{tabular}[c]{@{}c@{}}Unnecessary under our motion\\model specified in Sec.~\ref{subsec:st_trajectory_representation}\end{tabular} \\
\midrule[0.25pt]

\begin{tabular}[c]{@{}c@{}}\textbf{Dominance}\\\textbf{Checks}\end{tabular}
& No pairwise checks
& \begin{tabular}[c]{@{}c@{}}Cost/reachability checks via\\containment or sampling\end{tabular}
& \begin{tabular}[c]{@{}c@{}}Sound cone containment;\\approximate state/position checks\end{tabular} \\
\midrule[0.25pt]

\textbf{Heuristics}
& \begin{tabular}[c]{@{}c@{}}Triplet-based\\lower-bound graph\end{tabular}
& \begin{tabular}[c]{@{}c@{}}Pointwise\\cost-to-go bound\end{tabular}
& \begin{tabular}[c]{@{}c@{}}Motion-only, triplet, and\\interface-to-set bounds\end{tabular} \\
\midrule[0.25pt]

\begin{tabular}[c]{@{}c@{}}\textbf{Heuristic}\\\textbf{Reuse}\end{tabular}
& \begin{tabular}[c]{@{}c@{}}Across queries on the\\same spatial GCS\end{tabular}
& \begin{tabular}[c]{@{}c@{}}
No graph-based bounds reuse;\\
online joint optimization
\end{tabular}
& \begin{tabular}[c]{@{}c@{}}Across queries and ECD reservations\\on the same base ST-GCS\end{tabular} \\
\bottomrule
\end{tabular}
\end{table*}

\subsection{Spatiotemporal Planning on ST-GCSs}\label{subsec:st_problem_formulation}
Using the trajectory representation and cost in Sec.~\ref{subsec:st_trajectory_representation}, we seek a minimum-time trajectory from a given start space-time state $\mathbf{x}_s$ to a goal spatial position $\mathbf{p}_g$, with the arrival time determined by the planner. We encode this query by augmenting $G$ with a single auxiliary start vertex $v_s$ and a single auxiliary goal vertex $v_g$~\citep{bertsimas1997introduction}. These auxiliary vertices provide unique graph endpoints without preselecting the initial and terminal convex sets.

For the given start state, the auxiliary start set is
\[
X_{v_s}=\{\mathbf{x}_s\}.
\]
More generally, $X_{v_s}$ can be any convex set of admissible start states.
For the goal, we require the robot to remain at $\mathbf{p}_g$ after arrival until the time limit $t_{\max}$. The auxiliary goal set therefore contains exactly the arrival states that satisfy this goal-staying convention:
\begin{equation*}
\begin{aligned}
X_{v_g} = \Big\{&(\mathbf{p}_g,t)\Bigm\vert 0\leq t\leq t_{\max} \text{ and }\\
&(\mathbf{p}_g,t')\in \bigcup_{v\in V}X_v\ \text{for all } t'\in[t,t_{\max}]\Big\}.
\end{aligned}
\end{equation*}
Membership in $X_{v_g}$ ensures that the robot can remain at the goal throughout $[t,t_{\max}]$ within the space-time collision-free region $\bigcup_{v\in V}X_v$ represented by $G$. If the query instead specifies a full goal state, or only requires reaching $\mathbf{p}_g$ without remaining there, the definition of $X_{v_g}$ can be modified while the rest of the formulation remains unchanged.

\begin{figure}[t]
    \centering
    \includegraphics[width=\linewidth]{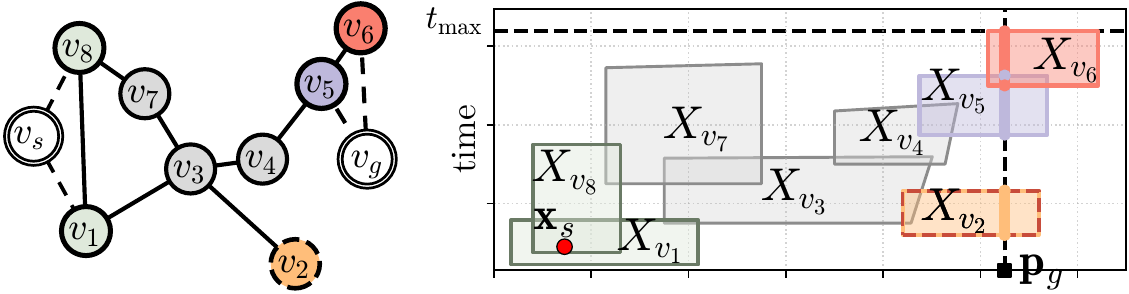}
    \caption{A query-augmented 2D ST-GCS with auxiliary start $v_s$ and goal $v_g$. Input-graph edges are solid while augmented edges are dashed. Vertices $v_1$ and $v_8$ are adjacent to $v_s$ because their sets contain $\mathbf{x}_s$. Vertices $v_5$ and $v_6$ are adjacent to $v_g$ because their sets intersect $X_{v_g}$, the set of goal states from which the robot can remain at $\mathbf{p}_g$ until $t_\text{max}$. If the robot is only required to reach $\mathbf{p}_g$, then $v_2$ is also adjacent to $v_g$.}
    \label{fig:start_goal_vertices}
\end{figure}

For clarity, we define the \textit{query-augmented} ST-GCS as $G^\sharp=(V^\sharp,E^\sharp,\mathcal{X}^\sharp)$, where $V^\sharp=V\cup\{v_s,v_g\}$, $\mathcal{X}^\sharp=\mathcal{X}\cup\{X_{v_s},X_{v_g}\}$, and $E^\sharp = \{(u,v)\mid u,v \in V^\sharp, u\neq v, X_u \cap X_v \neq \emptyset\}$. We denote this construction by $(G^\sharp,v_s,v_g)=\textsc{QueryAugment}(G,\mathbf{x}_s,\mathbf{p}_g)$. Fig.~\ref{fig:start_goal_vertices} illustrates a query-augmented ST-GCS from the original $G=(V,E,\mathcal{X})$.

With the query encoded in $G^\sharp$, we jointly optimize the vertex path and its entry and exit states using the following time-optimal planning program:
\begin{subequations}\label{eqn:stgcs}
\begin{align}
\min_{\Phi,\mathbf{x},\mathbf{y}}\quad
& c(\tau)=\sum_{v\in\pi_\Phi}\Delta t_v\label{eqn:stgcs:obj}\\*
\textbf{s.t.}\quad
&\pi_\Phi\in \mathcal{P}(v_s,v_g;G^\sharp),\label{eqn:stgcs:path}\\*
&\mathbf{x}_v,\mathbf{y}_v\in X_v, &&\forall v\in\pi_\Phi,\label{eqn:stgcs:in_set}\\*
&\mathbf{x}_v=\mathbf{y}_u, &&\makebox[36pt][r]{$\forall (u,v)\in E(\pi_\Phi),$}\label{eqn:stgcs:continuity}\\*
&\Delta t_v\geq 0, &&\forall v\in\pi_\Phi,\label{eqn:stgcs:forward_time}\\*
&-\mathbf{v}_{\mathrm{lim}}\Delta t_v\preceq\Delta\mathbf{p}_v\preceq\mathbf{v}_{\mathrm{lim}}\Delta t_v,
&&\forall v\in\pi_\Phi.\label{eqn:stgcs:vel_bound}
\end{align}
\end{subequations}
Eqn.~(\ref{eqn:stgcs:obj}) minimizes the trajectory cost $c(\tau)$ as defined in Eqn.~\eqref{eqn:traj_cost}. 
Eqns.~(\ref{eqn:stgcs:path}--\ref{eqn:stgcs:continuity}) specialize the path, set-membership, and continuity constraints extended from Eqns.~(\ref{eqn:gcs:path}--\ref{eqn:gcs:continuity}) to $G^\sharp$. Eqns.~(\ref{eqn:stgcs:forward_time}--\ref{eqn:stgcs:vel_bound}) apply the local motion constraints in Eqn.~(\ref{eqn:st_local_motion}) to every segment along the selected path. The auxiliary sets enforce the start state and admissible goal arrival times, so no additional endpoint constraints are needed.

\subsection{Solution Approaches}\label{subsec:solution_approaches}
The formulation in Eqns.~\eqref{eqn:stgcs} can be solved directly as an MICP, jointly selecting a sequence of space-time convex sets and optimizing the continuous trajectory within them. Alternatively, one can solve its continuous relaxation and recover candidate paths by stochastic rounding~\citep{marcucci2023motion,tang2025stgcs}, which highlights a useful observation: once a graph path is fixed, the remaining trajectory optimization is a convex program, as described in Sec.~\ref{subsec:convex_restriction}.

This key observation motivates search-based solvers such as IxG$^*$~\citep{natarajan2024ixg} and GCS$^*$~\citep{chia2024gcs} to interleave graph search and continuous trajectory optimization, which explore partial paths of convex sets and optimize trajectories along them. 
Since extending a path may change its optimal continuous states, these methods retain multiple partial paths reaching the same vertex. 
In Sec.~\ref{sec:st_search}, we build our best-first search (BFS) solver on this established paradigm.

We specialize our BFS solver to the time-optimal ST-GCS model in Sec.~\ref{subsec:st_trajectory_representation}. Its time and velocity structure supports excluding set revisitation and constructing reachability-based dominance checks. Our heuristic design supports reuse of precomputed bounds after ECD reservations. Tab.~\ref{tab:gcs_search_comparison} summarizes the major distinctions across existing search-based solvers and our solver.

\section{Graph Search on ST-GCSs}\label{sec:st_search}

We now develop the search space, pruning rules, and heuristics for the BFS solver introduced in Sec.~\ref{subsec:solution_approaches}.

\subsection{Search Space}\label{subsec:search_space}
A \textit{prefix path} is a partial vertex path starting at the auxiliary start vertex $v_s$.
In standard cost-minimal-path search on a discrete graph, such as A$^*$ search~\citep{hart1968formal}, a search state is typically identified by the current vertex. This is valid because the graph has a vertex-level optimal substructure.
Once the minimum cost-to-come to a vertex $v$ is known, any more expensive prefix ending at $v$ can be discarded, since all future costs depend only on $v$ and the remaining path.

This vertex-level optimal substructure does not generally hold for GCS planning, as discussed in~\cite{natarajan2024ixg} and~\cite{chia2024gcs}. The cost of a path is not the sum of fixed edge costs. Instead, once a vertex sequence is chosen, the continuous trajectory variables along the entire sequence are optimized jointly. Consequently, when a prefix path $\pi = \langle v_s,\ldots,v \rangle$ is extended to $\pi'= \langle v_s,\ldots,v,w \rangle$, the optimal continuous states along the earlier part of the path, including the states at $v$, may change. Thus, the optimal trajectory conditioned on $\pi'$ is not obtained by simply appending a locally optimal transition from $v$ to $w$ to the optimal trajectory conditioned on $\pi$. A prefix that is more expensive when considered only up to $v$ may still result in a lower-cost full trajectory after extension.

Thus, search nodes for planning on ST-GCSs must be indexed by the full prefix path, rather than only by the arrival vertex. Let $\Pi$ denote the set of partial vertex paths on the query-augmented $G^\sharp$ that start from the auxiliary start vertex $v_s$. As justified in Sec.~\ref{subsec:set_transition}, we restrict $\Pi$ to paths that do not revisit a vertex.
\begin{definition}[Path-Indexed Search Node]
A search node, identified by its prefix path, is a record
\[N=(\pi,v,\tau,g,f),\]
where $N.\pi \in \Pi$ is the prefix path, $N.v$ is the last vertex of $N.\pi$, $N.\tau$ is the optimal trajectory conditioned on $N.\pi$, $N.g = c(N.\tau)$ is the cost-to-come, and $N.f=N.g+\epsilon h(N.\pi)$ is the default search key (see Sec.~\ref{subsec:joint_key} for an alternative joint evaluation). For convenience, let $N.\mathbf{x}$ and $N.\mathbf{y}$ denote the entry and exit states of the terminal segment of $N.\tau$ in $X_{N.v}$, respectively.
\end{definition}

Here, $h:\Pi\rightarrow\mathbb{R}_{\geq 0}\cup\{+\infty\}$ is an \textit{admissible} heuristic defined in Definition~\ref{defn:admissible_heuristic}, and $\epsilon \geq 1$ is an inflation factor.

For a prefix $N.\pi=\langle v_s,\ldots,v\rangle$, we compute $N.\tau$ and $N.g$ by minimizing the elapsed-time objective in Eqn.~(\ref{eqn:stgcs:obj}), subject to Eqns.~(\ref{eqn:stgcs:in_set}--\ref{eqn:stgcs:vel_bound}) along that fixed prefix. This replaces the goal-reaching path constraint in Eqn.~(\ref{eqn:stgcs:path}), and all continuous states along the prefix are optimized jointly. If $v\neq v_g$, the terminal state is optimized within $X_v$, without a goal-position or goal-staying requirement. If $v=v_g$, membership in the auxiliary goal set enforces the goal conditions.

The states $N.\mathbf{x}$ and $N.\mathbf{y}$ are not part of the search-node identity; they are stored only as witnesses from the optimized trajectory conditioned on $N.\pi$. Since the terminal segment has no successor-continuity constraint, time-minimization implies $\mathbf{x}_{N.v}=\mathbf{y}_{N.v}$ in an optimal solution.
Therefore, $N.\mathbf{x}$ can be viewed as the stored witness state at $N.v$. If the path-conditioned program is infeasible, no search node is created for $N.\pi$.

To reason about which prefixes can lead to better solutions, we consider complete trajectories whose vertex paths extend $N.\pi$, allowing the continuous states along the prefix to be re-optimized.
\begin{definition}[Prefix-Consistency and Prefix-Conditioned Solution Cost]\label{defn:prefix_consistent}
For a generated search node $N$, a solution trajectory $\bar\tau$ is \emph{$N.\pi$-consistent} if its vertex path has $N.\pi$ as a prefix. Let $\mathcal{T}_\text{sol}(N.\pi)$ denote the set of all $N.\pi$-consistent solution trajectories. The \textit{prefix-conditioned optimal solution cost} is $J^*(N.\pi) = \inf_{\bar{\tau}\in \mathcal{T}_\text{sol}(N.\pi)} c(\bar{\tau})$, with $J^*(N.\pi) = +\infty$ if $\mathcal{T}_\text{sol}(N.\pi)=\emptyset$.
\end{definition}

Any $N.\pi$-consistent solution restricts to a feasible prefix of no larger cost, so $N.g\leq J^*(N.\pi)$, with equality at goal nodes. We thus define an admissible heuristic as a nonnegative lower bound on $J^*(N.\pi)-N.g$.
\begin{definition}[Admissible Heuristic]\label{defn:admissible_heuristic}
A heuristic $h:\Pi\rightarrow\mathbb{R}_{\geq 0}\cup\{+\infty\}$ is \textit{admissible} if, for every search node $N$, we have $N.g + h(N.\pi) \leq J^*(N.\pi)$, and $h(N.\pi)=0$ when $N.v = v_g$. Equivalently, $h(N.\pi)$ lower-bounds the \emph{optimal remaining solution cost} $J^*(N.\pi) - N.g$ from its prefix $N.\pi$, and this remaining cost is zero if $N.v=v_g$.
\end{definition}

Computing $J^*(N.\pi)$ requires solving the remaining planning problem, so we use $h(N.\pi)$ to lower-bound the additional cost beyond $N.g$.
Through $N.f$, this estimate guides expansion toward promising prefixes and supports upper-bound pruning (Sec.~\ref{subsubsec:upper_bound_pruning}).

\subsection{Set Transition and Revisitation Exclusion}\label{subsec:set_transition}

The candidate graph transition follows the topology of the query-augmented ST-GCS $G^\sharp$. For a vertex $v$, let 
\begin{align*}
\operatorname{adj}(v) = \{w \in V^\sharp \mid (v,w)\in E^\sharp\}
\end{align*}
denote its adjacent vertices. By construction of $G^\sharp$, $w\in \operatorname{adj}(v)$ implies $X_v\cap X_w\neq\emptyset$. The temporal direction and velocity feasibility of an extended path are not imposed at the graph-transition level, but are instead enforced by the path-conditioned convex program.

Unlike a monolithic GCS optimization, which needs additional machinery to explicitly allow repeated visits to the same convex set, graph search can naturally represent such paths by allowing the same vertex to appear multiple times in a prefix path. 
However, the ST-GCS trajectory representation and cost in Sec.~\ref{subsec:st_trajectory_representation} makes revisitation of the same space-time convex set unnecessary.
By Lemma~\ref{lemma:no_set_revisit}, it is sufficient to search over simple partial vertex paths without cycles. In Alg.~\ref{alg:st_search}, candidate successors are generated from $\operatorname{adj}(N.v)$, and successors already appearing in $N.\pi$ are skipped.
Notably, this result concerns repeated visits to the same space-time convex set instead of repeated visits to the same spatial region.
Therefore, a trajectory may leave and later re-enter a spatial region through distinct space-time convex sets.

\begin{lemma}\label{lemma:no_set_revisit}
Consider a feasible trajectory conditioned on a vertex path $\pi=\langle v_1,\ldots,v_i,\ldots,v_j,\ldots, v_l\rangle$, where $v_i=v_j=v$ for some $i<j$. Under the ST-GCS trajectory representation with motion constraints in Eqn.~(\ref{eqn:st_local_motion}) and cost in Eqn.~(\ref{eqn:traj_cost}), there exists a feasible trajectory conditioned on the shortened path $\pi'=\langle v_1,\ldots,v_i,v_{j+1},\ldots,v_l\rangle$, where the suffix $\langle v_{j+1},\ldots,v_l\rangle$ is omitted when $j=l$, with no higher cost.
\end{lemma}

\begin{sketch_proof}{App.~\ref{app:no_set_revisit}}
Convexity allows the first entry and a later exit to be joined by a straight segment inside $X_v$. This shortcut preserves the endpoint times and satisfies the same velocity bounds, so it remains feasible without increasing the time cost.
\end{sketch_proof}

\begin{algorithm}[t]
\DontPrintSemicolon
\small
\caption{Best-First Search (BFS) on ST-GCS}\label{alg:st_search}
\SetKwInOut{Input}{Input}
\SetKwInOut{Output}{Output}
\SetKwInOut{Param}{Param}
\Input{ST-GCS $G=(V,E,\mathcal{X})$, query $(\mathbf{x}_s,\mathbf{p}_g)$}
\Param{admissible heuristic $h$, inflation factor $\epsilon \geq 1$, valid incumbent pair $(c_\text{ub},\tau_\text{ub})\gets(+\infty,\emptyset)$, dominance check $\delta\gets\delta_{\emptyset}$}
\Output{solution trajectory if one exists}
$(G^\sharp,v_s,v_g)\gets \textsc{QueryAugment}(G,\mathbf{x}_s,\mathbf{p}_g)$\;
$\tau_r\gets$ zero length trajectory at $\mathbf{x}_s$\;\label{alg:st_search:root_trajectory}
$N_r\gets(\langle v_s\rangle,v_s,\tau_r,0,\textsc{EvaluateKey}(G^\sharp,\langle v_s\rangle,0))$\;\label{alg:st_search:create_root}
OPEN $\gets [N_r]$\Comment{min-heap prioritized by $N.f$}\;\label{alg:st_search:init_open}
initialize $S(v) \gets \emptyset$ for each vertex $v\in V^\sharp$\label{alg:st_search:def_S}\;
$S(v_s) \gets \{N_r\}$\;
\While{\normalfont OPEN $\neq\emptyset$}{\label{alg:st_search:loop_st}
    $N\gets$ OPEN.pop()\;\label{alg:st_search:pop_OPEN}
    \If{$N.v = v_g$}{\label{alg:st_search:goal_cond}
        \Return $N.\tau$\;\label{alg:st_search:return_sol}
    }
    \ForEach{$w\in \operatorname{adj}(N.v)$ with $w\notin N.\pi$}{\label{alg:st_search:expand_node_start}
        $\pi\gets N.\pi \oplus \langle w \rangle$\;\label{alg:st_search:extend_pi}
        $(\tau,g)\gets\textsc{PathOptimize}(G^\sharp,\pi)$\label{alg:st_search:partial_traj_opt}\;
        \If{$\tau$ is feasible}{
        $N_{\text{\normalfont child}}\gets(\pi,w,\tau,g,\textsc{EvaluateKey}(G^\sharp,\pi,g))$\;\label{alg:st_search:create_child}
    \If{$N_{\text{\normalfont child}}.f<c_\text{ub}$ \textbf{and} \textbf{not} $\delta(N_{\text{\normalfont child}},S(w))$}{\label{alg:st_search:dc}
                Update $S(w)$ with $N_{\text{\normalfont child}}$\;
                OPEN.add($N_{\text{\normalfont child}})$\;\label{alg:st_search:add_child_to_OPEN}
            }
        }
    }
}
\eIf{$\tau_\text{ub}\neq\emptyset$}{
\Return $\tau_\text{ub}$ \textbf{and} ``\textit{no better solution found}''\;\label{alg:st_search:return_incumbent}}{ \Return ``\textit{no solution found}'';\label{alg:st_search:fail}
}
\vspace{3pt}

\renewcommand{\texttt}[1]{\scshape{\textsc{#1}}}
\SetKwFunction{FMain}{PathOptimize}
\SetKwProg{Fn}{Function}{:}{}
\Fn{\FMain{$G^\sharp$, $\pi$}}
{\label{func:cvx_opt}
construct the convex program on the fixed path $\pi$, using objective~(\ref{eqn:stgcs:obj}) and constraints~(\ref{eqn:stgcs:in_set}--\ref{eqn:stgcs:vel_bound})\;
solve the resulting convex program\;
\Return the optimal trajectory $\tau$ conditioned on $\pi$ and its cost $g=c(\tau)$, or report infeasibility
}
\end{algorithm}

\subsection{Best-First Search (BFS) Solver}
Alg.~\ref{alg:st_search} follows the general path-based best-first search scheme used in IxG$^*$~\citep{natarajan2024ixg} and GCS$^*$~\citep{chia2024gcs}, specialized here to time-optimal ST-GCS planning. Given an ST-GCS $G=(V,E,\mathcal{X})$ and a query $(\mathbf{x}_s,\mathbf{p}_g)$, it first applies \textsc{QueryAugment} to construct the query-augmented ST-GCS $G^\sharp=(V^\sharp,E^\sharp,\mathcal{X}^\sharp)$ from Sec.~\ref{subsec:st_problem_formulation}. It then searches over path-indexed nodes and returns a solution trajectory if one is found. Optionally, it can take a feasible incumbent trajectory $\tau_\text{ub}$ with cost $c_\text{ub}$; if the search exhausts OPEN without finding a better trajectory, this incumbent is returned. The optimality guarantee depends on the heuristic inflation factor, the validity of the incumbent upper bound, and whether the selected dominance check is sound.
The default dominance check $\delta_\emptyset$ always returns false, so no nodes are pruned by dominance.

Alg.~\ref{alg:st_search} summarizes the procedure. \textsc{EvaluateKey} returns
$g+\epsilon h(\pi)$ under separate evaluation, or the optimal
value of Eqn.~(\ref{eqn:joint_key}) under joint evaluation
with $h_{\mathrm{mot}}$. The solver creates the root node $N_r$ corresponding to the trivial path $\langle v_s \rangle$, inserts into OPEN, and stores it in $S(v_s)$  (Lines~\ref{alg:st_search:root_trajectory}--\ref{alg:st_search:def_S}). The map $S$ associates each vertex $v$ with a set $S(v) = \{N\,|\,N.v=v\}$ of accepted search nodes. As in GCS$^*$~\citep{chia2024gcs}, this set may contain multiple nodes with different prefix paths ending at $v$, which are retained for dominance comparisons.
At each iteration, the solver expands the node $N$ with the smallest $f$-value in OPEN (Lines~\ref{alg:st_search:loop_st}--\ref{alg:st_search:pop_OPEN}).
If $N.v = v_g$, the stored trajectory $N.\tau$ is returned as a solution (Lines~\ref{alg:st_search:goal_cond}--\ref{alg:st_search:return_sol}).
Otherwise, the solver expands $N$ by considering each adjacent vertex $w$ of $N.v$ that does not already appear in $N.\pi$ (Lines~\ref{alg:st_search:expand_node_start}--\ref{alg:st_search:add_child_to_OPEN}). For each extended path $\pi = N.\pi \oplus \langle w \rangle$, the solver calls \textsc{PathOptimize} to solve the corresponding path-conditioned convex program. 
If it is feasible, a child node is created (Lines~\ref{alg:st_search:partial_traj_opt}--\ref{alg:st_search:create_child}). The child node is discarded if its key is no smaller than the input cost upper bound $c_\text{ub}$, or if it is rejected by the dominance check $\delta$ against some nodes in $S(w)$. 
Otherwise, the child node is inserted into both $S(w)$ and OPEN, and previously accepted nodes in $S(w)$ that the check rejects in favor of the child node are removed (Lines~\ref{alg:st_search:create_child}--\ref{alg:st_search:add_child_to_OPEN}). If OPEN becomes empty, the solver returns the incumbent trajectory $\tau_\text{ub}$ if one was provided; otherwise, it reports failure (Lines~\ref{alg:st_search:return_incumbent}--\ref{alg:st_search:fail}).

\subsubsection{Upper-Bound Pruning}\label{subsubsec:upper_bound_pruning}

Inspired by~\cite{natarajan2024ixg}, Alg.~\ref{alg:st_search} prunes nodes with $N.f\geq c_\text{ub}$, where $c_\text{ub}=c(\tau_\text{ub})$ for a feasible incumbent trajectory $\tau_\text{ub}$ for the same query.
We obtain the incumbent using a vertex-indexed best-first search on $G^\sharp$ that keeps only the lowest-cost node found at each arrival vertex and validates each candidate path with \textsc{PathOptimize}.
This vertex-indexed pruning does not guarantee optimality (Sec.~\ref{subsec:search_space}), so the auxiliary search serves only to find a feasible incumbent, which Alg.~\ref{alg:st_search} returns if OPEN is exhausted.
If such procedure failed, we simply set $(c_\text{ub},\tau_\text{ub})=(+\infty,\emptyset)$.

For $\epsilon=1$, admissibility gives $N.f\leq J^*(N.\pi)$, ensuring that pruned prefixes cannot improve the incumbent, although upper-bound pruning alone generally does not reduce expansions except possibly for ties when the incumbent is optimal.
With heuristic inflation ($\epsilon>1$), pruning can reduce expansions by cutting off branches that the search would otherwise explore, but may also discard improving prefixes. Alg.~\ref{alg:st_search} nevertheless retains $\epsilon$-optimality with an admissible heuristic and sound dominance checks (Theorem~\ref{thm:cost_optimality}).

\subsubsection{Pruning with Dominance Checks}\label{subsubsec:define_dc} 
We now define the dominance relation used by the optional check $\delta$ on Line~\ref{alg:st_search:dc} of Alg.~\ref{alg:st_search}. Since search nodes are path-indexed, dominance is defined through the path-conditioned optimal solution cost $J^*(\cdot)$ (Definition~\ref{defn:prefix_consistent}). Consider two search nodes $N$ and $N'$ with the same arrival vertex $v$. We say that $N'$ \textit{dominates} $N$ if $J^*(N'.\pi)\leq J^*(N.\pi)$.

A dominance check $\delta(N,S(N.v))$ is \emph{sound} if it returns true only when there exists a retained node $N'\in S(N.v)$ that dominates $N$. Sound checks are sufficient but may be conservative.
They need not detect all dominated nodes, and only compare nodes pairwise rather than detecting whether a set of nodes jointly dominate another node. In contrast, \emph{approximate} dominance checks may prune without certifying this relation thus do not by themselves preserve the theoretical guarantee.

\subsubsection{Optimality with Sound Pruning}
The following theorem states the optimality guarantee for Alg.~\ref{alg:st_search} if the pruning operations in Line~\ref{alg:st_search:dc} are sound.
\begin{theorem}[$\epsilon$-Optimality]\label{thm:cost_optimality}
For any $\epsilon\geq 1$, with an admissible heuristic $h$, a valid incumbent $(c_\text{ub},\tau_\text{ub})$, and a sound dominance check, Alg.~\ref{alg:st_search} returns an $\epsilon$-optimal trajectory whenever a feasible one exists. 
The same guarantee holds with the joint search-key defined in Sec.~\ref{subsec:joint_key}. 
If $\epsilon=1$, the returned solution is cost-optimal.
\end{theorem}
\begin{sketch_proof}{App.~\ref{app:cost_optimality}}
We maintain an OPEN representative of an optimal continuation with key at most $\epsilon c^*$, where $c^*$ is the optimal cost, unless the incumbent already satisfies the same bound. Admissibility and sound pruning preserve this representative, so expanding the minimum-key goal gives the stated cost guarantee.
\end{sketch_proof}

\subsection{Pairwise Dominance Checks}\label{subsec:dc}
We now present three pairwise dominance checks as the optional check $\delta$ used in Alg.~\ref{alg:st_search}.
Throughout this subsection, let $N$ be a newly generated node and $N'\in S(N.v)$ be a retained node with the same arrival vertex $v=N.v=N'.v$. Write $\mathcal S=S(v)$ for this set of retained nodes.
The trivial case is $v = v_g$, where both nodes already represent solution trajectories, and exact dominance reduces to directly comparing their trajectory costs. The checks below are therefore described for non-goal vertices $v \neq v_g$.
Each check compares $N$ only with each retained node $N'$, rather than testing whether several retained nodes jointly dominate $N$. This pairwise restriction affects pruning power but not correctness when the check is sound.

Recall from Sec.~\ref{subsubsec:define_dc} that $N'$ dominates $N$ if $J^*(N'.\pi)\leq J^*(N.\pi)$. A dominance check is sound if it returns true only when this dominance relation is guaranteed to hold for some $N'\in S(N.v)$. Among the three checks below, $\delta_\text{set}$ is sound, while $\delta_\text{state}$ and $\delta_\text{pos}$ are approximate checks that trade optimality guarantees for stronger empirical pruning.

\subsubsection{Sound Set-Containment Check}
The sound check $\delta_\text{set}$ asks whether a retained node $N'$ can reach every feasible exit state of the shared set $X_v$ that an $N.\pi$-consistent trajectory could use.
Reaching the same exit at the same time allows it to follow the same remaining trajectory without increasing cost. Let $N.\pi = \langle v_s,\ldots,u,v\rangle$ and define the arrival set
\begin{align}
\label{eqn:R_N}
R_N=X_u\cap X_v\cap\{\mathbf{x} \mid \mathbf{x}.t\geq N.\mathbf{x}.t\}.
\end{align}
The set $R_N$ contains all states that any $N.\pi$-consistent solution trajectory may use to enter $v$. Continuity (Eqn.~(\ref{eqn:gcs:continuity})) requires membership in the predecessor interface $X_u\cap X_v$, and $N.\mathbf{x}.t$ is the earliest arrival time achieved by the path-conditioned optimum for $N.\pi$. We check all possible entry states in $R_N$ because extending the prefix may change its optimized entry state.

For a space-time state $\mathbf{x}$, define its forward reachable generalized cone under the velocity limits by
\begin{equation*}
\begin{aligned}
C(\mathbf{x})=&\{\mathbf{z}\in\mathbb{R}^{m+1}\mid \mathbf{z}.t\geq \mathbf{x}.t,\\
&-\mathbf{v}_{\mathrm{lim}}(\mathbf{z}.t-\mathbf{x}.t) \preceq \mathbf{z}.\mathbf{p}-\mathbf{x}.\mathbf{p} \preceq
\mathbf{v}_{\mathrm{lim}}(\mathbf{z}.t-\mathbf{x}.t)\}.
\end{aligned}
\end{equation*}
Here, $\mathbf{z}=(\mathbf{z}.\mathbf{p},\mathbf{z}.t)$ is a candidate destination state.
The cone $C(\mathbf{x})$ describes reachability by a single segment under the time-monotonicity and velocity constraints.

The set $X_v\cap\bigcup_{\mathbf{x}\in R_N}C(\mathbf{x})$ is an outer approximation of states reachable through $N$ within $X_v$, since $R_N$ may include entries infeasible along the full prefix. In contrast, $X_v\cap C(N'.\mathbf{x})$ is an inner approximation for $N'$, obtained from its stored feasible prefix. Containment of the former in the latter is therefore sufficient, but not necessary, for dominance.
The check $\delta_\text{set}$ is formally defined by
\begin{equation}\label{eqn:delta_esc}
\begin{aligned}
\delta_\text{set}(N,\mathcal S) \equiv {} &
\exists\, N' \in \mathcal S, \\
& \textbf{s.t. } \forall\, \mathbf{x} \in R_N,\;
C(\mathbf{x}) \subseteq C(N'.\mathbf{x}).
\end{aligned}
\end{equation}
By transitivity of cone reachability, requiring $C(\mathbf{x})\subseteq C(N'.\mathbf{x})$ for every $\mathbf{x}\in R_N$ is equivalent to $R_N\subseteq C(N'.\mathbf{x})$. We evaluate this containment by maximizing each cone halfspace's linear form over $R_N$ and comparing the result with its bound at $N'.\mathbf{x}$. These maxima are computed lazily and cached for reuse across comparisons with retained nodes.

\begin{lemma}[$\delta_\text{set}$ Soundness]\label{lem:esc_sound}
If $\delta_\text{set}(N,S(N.v))$ is true, then $N$ is dominated by a retained node $N'\in S(N.v)$.
\end{lemma}
\begin{sketch_proof}{App.~\ref{app:dominance_soundness}}
Containment and convexity let $N'$ reach the same exit state and follow the same suffix without increasing cost.
\end{sketch_proof}

\subsubsection{Approximate Arrival-State-Containment Check}
The check $\delta_\text{state}$ approximates $\delta_\text{set}$ by testing only the stored witness state $N.\mathbf{x}$, rather than all states in $R_N$. 
Formally, we define the check $\delta_\text{state}$ as
\begin{equation*}
\begin{aligned}
\delta_\text{state}(N,\mathcal S) \equiv& \,\exists\, N'\in \mathcal S,\, \textbf{s.t. } C(N.\mathbf{x})\subseteq C(N'.\mathbf{x}).
\end{aligned}
\end{equation*}

$\delta_\text{state}$ is cheaper than $\delta_\text{set}$ as it only compares two cones. 
It does not certify dominance as $N.\mathbf{x}$ is only the stored witness state of the prefix-optimal trajectory for $N.\pi$, while an optimal $N.\pi$-consistent solution trajectory may use a different state to enter $X_{N.v}$.

\subsubsection{Approximate Position-Based Check}
The set-containment check $\delta_\text{set}$ can be expensive because it reasons over all states in $R_N$. The check $\delta_\text{pos}$ offers a cheaper alternative by comparing two prefixes only at a sampled spatial position $\mathbf{p}\in\{\mathbf{x}.\mathbf{p}\,|\,\mathbf{x}\in X_{N.v}\}$.
This adapts the sampling-based cost comparison in GCS$^*$~\citep{chia2024gcs} to minimum elapsed-time costs at sampled spatial positions.

For a generated search node $N$ ending at $v=N.v$, we compute $J_\mathbf{p}(N.\pi)$ using the path-conditioned form of Eqns.~(\ref{eqn:stgcs}). We replace the path constraint in Eqn.~(\ref{eqn:stgcs:path}) with $\pi_\Phi=N.\pi$ and add $\mathbf{y}_v.\mathbf{p}=\mathbf{p}$, leaving the arrival time $\mathbf{y}_v.t$ free.
The objective in Eqn.~(\ref{eqn:stgcs:obj}) and the constraints in Eqns.~(\ref{eqn:stgcs:in_set}--\ref{eqn:stgcs:vel_bound}) are retained along the fixed prefix.
Thus, $J_\mathbf{p}(N.\pi)$ is the minimum elapsed time to reach $\mathbf{p}$ in $X_v$, with value $+\infty$ if the program is infeasible.
The sampled position serves as an intermediate endpoint, so no goal-staying requirement is imposed. Formally, we define
\begin{equation*}
\begin{aligned}
\delta_\text{pos}(N,\mathcal S) \equiv {} & \exists\,N'\in \mathcal S,\\
&\textbf{s.t. }\;
J_\mathbf{p}(N'.\pi)\leq J_\mathbf{p}(N.\pi)<+\infty.
\end{aligned}
\end{equation*}

\begin{figure}[t]
\centering
\includegraphics[width=\linewidth]{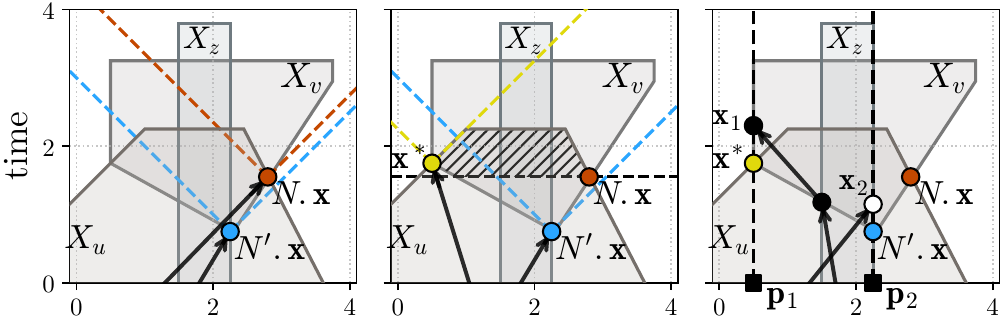}
\caption{
Dominance checks comparing $N$ with a retained node $N'$ at $X_v$. Left: $\delta_\text{state}$ prunes $N$ using stored-witness cone containment. Middle: $\delta_\text{set}$ keeps $N$ because the arrival set $R_N$ (hatched) contains $\mathbf{x}^*$ outside $C(N'.\mathbf{x})$. Right: $\delta_\text{pos}$ compares optimized arrival times at a sampled position, giving different pruning decisions at $\mathbf{p}_1$ and $\mathbf{p}_2$.
}
\label{fig:dc_demo}
\end{figure}

$\delta_\text{pos}$ can be strengthened by using multiple samples. In the limiting case of covering all relevant positions in $X_{N.v}$, it can certify a pairwise dominance relation between $N$ and a retained node $N'$. One can further extend the idea to certify joint dominance by allowing different retained nodes in $S(N.v)$ to dominate $N$ at different positions. These extensions are more expensive. Experimentally, we use only one sampled position, so $\delta_\text{pos}$ does not certify dominance.

\subsubsection{Comparison and Use with Upper-Bound Pruning}\label{subsec:seq_dc}
Fig.~\ref{fig:dc_demo} compares $\delta_\text{set}$ over all relevant states in $R_N$, $\delta_\text{state}$ over stored witness states, and $\delta_\text{pos}$ at sampled spatial positions.
We first prune nodes with $N.f\geq c_\text{ub}$ and then apply the chosen dominance check.
With $\delta_\emptyset$ or $\delta_\text{set}$, the guarantee in Theorem~\ref{thm:cost_optimality} applies.
Using $\delta_\text{state}$ or $\delta_\text{pos}$ can substantially accelerate search in practice, but forfeits this guarantee and may discard prefixes leading to a solution.
Together, the two pruning steps can increase the likelihood that no surviving node reaches the goal, in which case Alg.~\ref{alg:st_search} returns the incumbent $\tau_\text{ub}$ if available.

The check $\delta_\text{pos}$ is usually more aggressive and more sensitive to upper-bound pruning since sampling may miss useful unsampled states.
The check $\delta_\text{state}$ compares stored witness states through cone containment and tends to be more conservative in practice.

\subsection{Admissible Heuristics}\label{subsec:st_planning_heur}

\begin{figure}[t]
\centering
\includegraphics[width=\linewidth]{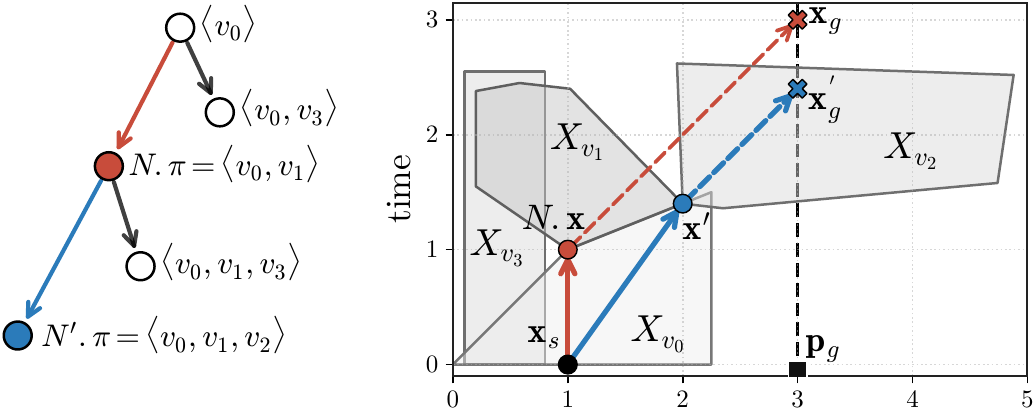}
\caption{Separate prefix-cost and fixed-witness heuristic evaluation versus joint search-key optimization. Expanding $N$ to $N'$ re-optimizes the prefix through $\mathbf{x}'$ instead of $N.\mathbf{x}$, so a valid cost-to-go lower bound at $N.\mathbf{x}$ (red dashed) can overestimate the remaining cost for $N.\pi$. With a unit speed limit and $\epsilon=1$, the joint optimizer (Sec.~\ref{subsec:joint_key}) coincides with the optimal $N.\pi$-consistent trajectory (blue).}
\label{fig:counter-example-point-to-set-heur}
\end{figure}

We now describe admissible heuristics for Alg.~\ref{alg:st_search}. All heuristics below are zero when $N.v = v_g$, since $N.\tau$ is already a solution trajectory. Thus, we consider non-goal nodes $N.v \neq v_g$. Throughout this subsection, a minimum over an empty candidate set is $+\infty$.
For separate prefix-cost and heuristic evaluation, Definition~\ref{defn:admissible_heuristic} requires the heuristic to lower-bound the optimal remaining solution cost from a prefix path $N.\pi$, rather than the cost-to-go from the stored witness state $N.\mathbf{x}$ alone. This distinction matters since extending $N.\pi$ re-optimizes its trajectory, which may then pass through a different state in $X_{N.v}$. Thus, a heuristic evaluated from $N.\mathbf{x}$ can overestimate the optimal remaining solution cost for the path-indexed node (Fig.~\ref{fig:counter-example-point-to-set-heur}).

For a non-goal node $N$, we define the prefix interface
\begin{equation}\label{eqn:prefix_interface}
I_N =
\begin{cases}
\{\mathbf{x}_s\}, & \text{if } N.\pi=\langle v_s\rangle,\\
X_u\cap X_v, & \text{if } N.\pi=\langle v_s,\ldots,u,v\rangle,
\end{cases}
\end{equation}
which contains all states from which an $N.\pi$-consistent solution can enter its terminal set. It serves as the source set for the heuristics, allowing the entry state to vary while retaining the interface imposed by the prefix. 

We construct $h_{\mathrm{mot}}$, $h_{\mathrm{tri}}$, and $h_{\mathrm{tab}}$ by relaxing different subsets of the path-conditioned ST-GCS constraints on graph topology, set membership, continuity between adjacent sets, motion, and the query-specific goal position.
Each relaxation yields a lower bound on the optimal remaining solution cost from $N.\pi$ and hence an admissible heuristic under Definition~\ref{defn:admissible_heuristic}.
Their pointwise maximum $h_{\mathrm{max}}$ also remains admissible and retains the strongest of these bounds.
The admissibility lemmas and their proofs are provided in App.~\ref{app:heuristic_admissibility}.

To reduce online computation, some bounds can be precomputed using the static environment and any dynamic obstacles known before the query. Query-specific information, such as $\mathbf{x}_s$, $\mathbf{p}_g$, and $v_g$, is incorporated online after query augmentation. In MRMP, previously planned robot trajectories are unknown offline. 
Their ECD updates only remove feasible space, so the heuristics computed on the initial ST-GCS remain admissible, as discussed in Sec.~\ref{subsec:mrmp_heuristic_reuse}.

For the ST-GCS trajectory representation in Sec.~\ref{subsec:st_trajectory_representation}, define the velocity-only travel-time lower bound from a set $Y$ of space-time states to the goal position $\mathbf{p}_g$ as
\begin{align*}
\rho(Y, \mathbf{p}_g)=\min_{\mathbf{x}\in Y}\left\|(\mathbf{p}_g-\mathbf{x}.\mathbf{p})\oslash\mathbf{v}_{\mathrm{lim}}\right\|_\infty 
\end{align*}
where $\oslash$ denotes element-wise division. Since the arrival time of the query goal is unconstrained, all our heuristics estimate the cost to $\mathbf{p}_g$ rather than to a fixed goal state.

\subsubsection{Motion-Only Heuristic}
The heuristic $h_{\mathrm{mot}}$ minimizes travel time (Eqn.~(\ref{eqn:stgcs:obj})) from $I_N$ to $\mathbf{p}_g$ under only the time-monotonicity and velocity constraints in Eqns.~(\ref{eqn:stgcs:forward_time}--\ref{eqn:stgcs:vel_bound}).
Dropping the graph-path, set-membership, and inter-set continuity constraints for the remaining trajectory (Eqns.~(\ref{eqn:stgcs:path}--\ref{eqn:stgcs:continuity})) gives the lower bound
\begin{align*}
h_{\mathrm{mot}}(N.\pi)= \rho(I_N, \mathbf{p}_g).
\end{align*}
The minimization covers all of $I_N$, and its minimizer need not be a vertex.
Restricting the source set from $X_{N.v}$ to $I_N$ gives an at least as tight motion-only bound by preserving the predecessor--current-set interface.

\subsubsection{Triplet-Relaxation Heuristic}
The heuristic $h_{\mathrm{tri}}$ retains graph topology and local set-membership and motion constraints but relaxes continuity between consecutive local transitions, following an idea used by spatial GCS solvers~\citep{natarajan2024ixg,tang2026ghost}.
For each triplet $(u,v,w)\in V^3$ with $u,w$ adjacent to $v$ in $G$, an offline local convex program computes the local lower-bound cost $q(u,v,w)$ as the minimum cost of a straight space-time segment inside $X_v$ from $X_u\cap X_v$ to $X_v\cap X_w$, subject to time monotonicity and velocity limits.
If no such segment exists, $q(u,v,w)=+\infty$.
At query time, the same program computes the analogous costs $q(u,v,v_g)$ for the auxiliary goal vertex $v_g$.

For $N.\pi=\langle\ldots,u,v\rangle$, let $\Pi^\text{suf}_N$ be the set of suffixes $\pi=\langle u,v=w_0,\ldots,w_l=v_g\rangle$ with predecessor $u$, and define
\begin{equation*}
\begin{aligned}
h_{\mathrm{tri}}(N.\pi) = &\min_{\pi\in \Pi^\text{suf}_N}\Big[q(u, w_0, w_1)+
\sum_{i=1}^{l-1} q(w_{i-1},w_{i},w_{i+1})\Big],
\end{aligned}
\end{equation*}
where the first term is omitted if $u=v_s$, and the summation is omitted if $l=1$.
For the root $N.\pi=\langle v_s\rangle$, the first two triplets are omitted, and the minimum is also taken over the first input-graph vertex $w_1$ after $v_s$.
Independent triplet optimization relaxes shared-interface continuity, making this minimum a lower bound on the remaining cost.

After adding the query-specific goal costs, Dijkstra's algorithm~\citep{dijkstra2022note} finds the minimum-cost relaxed suffix online on a directed graph with ordered adjacent pairs $(u,v)$ as vertices and edges $(u,v)\rightarrow(v,w)$ weighted by $q(u,v,w)$.
Pairs ending at $v_g$ are destinations, infinite-cost edges are excluded, and omitted start terms have zero cost.


\subsubsection{Interface-to-Set Cost Table Heuristic}
The heuristic $h_\text{tab}$ precomputes stronger lower bounds by solving interface-to-set problems offline and tabulating their costs. For any input-graph interface $I=X_u\cap X_v$ with $u, v \in V$ and any input-graph vertex $w\in V$, let $d(I,w)$ denote the minimum trajectory cost of reaching any state in $X_w$ from any state in $I$, over all graph paths, subject to the same graph, set-membership, continuity, and motion constraints. These values can be precomputed offline for all interfaces and vertices using Alg.~\ref{alg:st_search}. They are query independent because they start from input-graph interfaces and end at input-graph sets.

For a non-goal node $N$, we define the relaxed table-start value $\hat{d}_N(v_p)$ as
\begin{align*}
\begin{dcases}
\min_{v\in\operatorname{adj}(v_s)\cap V \atop w\in\operatorname{adj}(v)\cap V} d(X_v\cap X_w, v_p),
& \text{if } N.v = v_s,\\
\min_{w\in\operatorname{adj}(N.v)\cap V} d(X_{N.v}\cap X_w, v_p),
& \text{if } N.\pi = \langle v_s, N.v\rangle,\\
d(I_N, v_p), & \text{otherwise}.
\end{dcases}
\end{align*}
The first two cases handle query-specific start interfaces by omitting the nonnegative cost from $\mathbf{x}_s$ to the first input-graph interface used by the cost table. Thus, for a non-goal $N$ with $N.v\notin\operatorname{adj}(v_g)$ and either $N.v\ne v_s$ or $\operatorname{adj}(v_s)\cap \operatorname{adj}(v_g)\cap V=\emptyset$, we define the heuristic by
\begin{equation*}
h_{\mathrm{tab}}(N.\pi)=
\min_{\substack{
v_p\in \operatorname{adj}(v_g)\cap V\\
u\in\operatorname{adj}(v_p)\cap V}}
\left[
\hat{d}_N(v_p)+\rho(X_u\cap X_{v_p}, \mathbf{p}_g)
\right].
\end{equation*}
The two terms lower-bound travel to a goal-adjacent set and the final segment to $\mathbf{p}_g$, respectively.
If $N.v \in \operatorname{adj}(v_g)$, or $N.v=v_s$ and $\operatorname{adj}(v_s)\cap
\operatorname{adj}(v_g)\cap V\ne\emptyset$, we instead set $h_{\mathrm{tab}}(N.\pi)=\rho(I_N,\mathbf{p}_g)$, the motion-only bound.

\subsubsection{Comparison and Maximum Heuristic}
\begin{figure}[t]
    \centering
    \includegraphics[width=\linewidth]{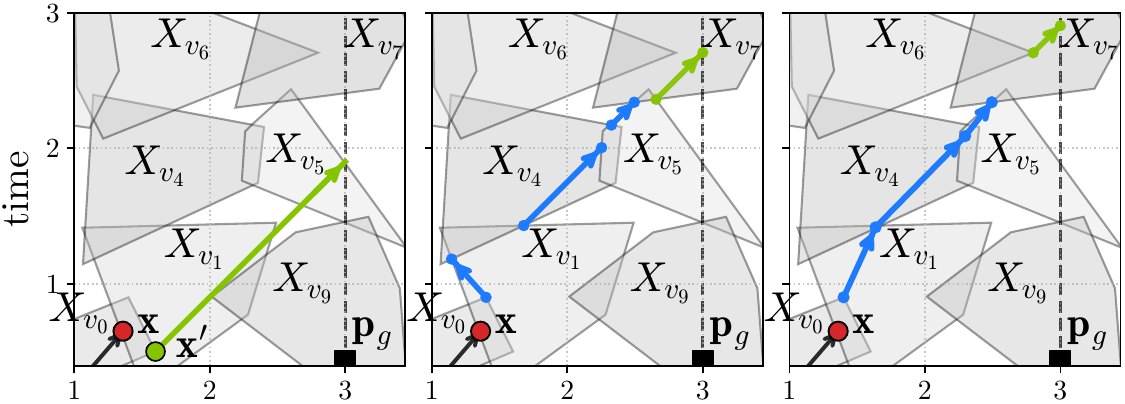}
    \caption{Comparisons of the heuristics from the prefix interface $X_{v_0}\cap X_{v_1}$ to $\mathbf{p}_g$; green denotes online computation and blue offline costs. Left: $h_\text{mot}$ uses only motion constraints. Middle: $h_\text{tri}$ sums independently optimized triplet costs along a minimum-cost suffix, including an online goal transition. Right: $h_\text{tab}$ combines a precomputed interface-to-set trajectory cost with an online goal correction.}
    \label{fig:heur_2d}
\end{figure}

Fig.~\ref{fig:heur_2d} shows the complementary lower bounds supplied by the three heuristics on an example query. The maximum heuristic $h_\text{max}$ retains the strongest bound at each search node by taking their pointwise maximum:
\begin{align*}
h_{\text{\normalfont max}}(N.\pi) = 
\max\left\{
h_\text{mot}(N.\pi),
h_\text{tri}(N.\pi),
h_\text{tab}(N.\pi)
\right\}.
\end{align*}

\subsection{Joint Search-Key Evaluation with $h_{\mathrm{mot}}$}
\label{subsec:joint_key}

Separate evaluation of $N.g$ and $h_{\mathrm{mot}}(N.\pi)=\rho(I_N,\mathbf{p}_g)$ minimizes the prefix cost and the heuristic lower bound independently over the prefix interface $I_N$. However, the state at which an $N.\pi$-consistent solution enters $X_{N.v}$ depends on the complete path, whose suffix has not yet been determined. The two minimizations may therefore select different interface states, without accounting for the motion needed to connect these choices within $X_{N.v}$, leaving a gap in the cost estimate, as illustrated in Fig.~\ref{fig:counter-example-point-to-set-heur}.

Jointly optimizing the prefix cost and a heuristic at the same interface state can tighten this bound, but requires a suitable state-dependent heuristic that can be incorporated into the path-conditioned convex program. The precomputed scalar costs $q(u,v,w)$ and $d(I,w)$ used by $h_{\mathrm{tri}}$ and $h_{\mathrm{tab}}$ have already minimized over interface states, while the heuristics also optimize over relaxed suffix paths. Coupling these heuristics with the prefix would generally require additional, potentially expensive nonconvex optimization over suffix paths and interface states online, rather than directly reusing the precomputed values.

\begin{figure*}[t]
    \centering
    \includegraphics[width=\linewidth]{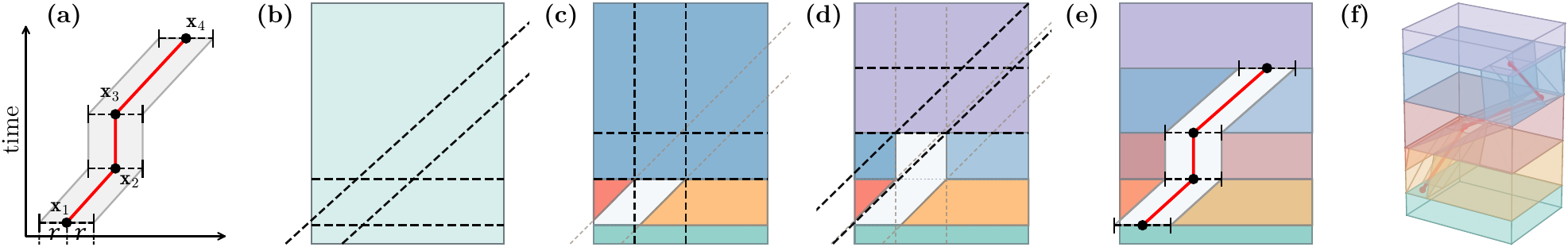}
    \caption{Reserved trajectory occupancy $\Omega(\tau, r)$ in Eqn.~(\ref{eqn:traj_occ}) and the ECD scheme. (a) A piecewise-linear trajectory $\tau=\langle\mathbf{x}_1,\mathbf{x}_2,\mathbf{x}_3,\mathbf{x}_4\rangle$ in 2D space-time and its occupancy $\Omega(\tau, r)$, represented as a union of parallelotopes. (b) The input 2D ST-GCS, together with the first separating hyperplanes induced by the occupancy parallelotopes. (c)--(d) Progressive ECD slicing of $X_v$. Previously used separating hyperplanes are shown in light dashed lines, and newly introduced hyperplanes are shown in black dashed lines. (e) The updated ST-GCS showing the retained convex sets, the trajectory, and the outline of $\Omega(\tau, r)$. (f) A 3D ST-GCS after ECD reservation, with a trajectory in red and its reserved occupancy overlaid.}
    \label{fig:occ}
\end{figure*}

A tractable case arises when the graph-path and set-membership constraints on the suffix are relaxed, yielding a convex lower bound from a given interface state. Our motion-only bound $\rho(\{\mathbf{x}_v\},\mathbf{p}_g)$ has this property and can be jointly optimized with the prefix, as in GCS$^*$~\citep{chia2024gcs}. For $v=N.v$, define
\begin{equation}\label{eqn:joint_key}
N.f=\min_{\tau}\left[c(\tau)+\epsilon\,\rho(\{\mathbf{x}_v\},\mathbf{p}_g)\right],
\end{equation}
subject to the path-conditioned constraints in Eqns.~(\ref{eqn:stgcs}) with $\pi_\Phi=N.\pi$ and $\mathbf{y}_v=\mathbf{x}_v$. Therefore, the relaxed continuation starts at the prefix entry state $\mathbf{x}_v$, jointly optimized with $\tau$ rather than fixed to the stored witness $N.\mathbf{x}$.
In Fig.~\ref{fig:counter-example-point-to-set-heur}, joint evaluation shifts the interface state from $N.\mathbf{x}$ to $\mathbf{x}'$, yielding the optimal $N.\pi$-consistent trajectory (blue). The motion-only term is a weighted $L_\infty$ norm, so the program is convex. The inflation factor $\epsilon$ weights only the motion-only cost-to-go term.

The joint optimizer need not minimize the prefix cost alone, so dominance requires a separate $g$-only solve. Joint evaluation thus incurs an additional convex optimization: Eqn.~(\ref{eqn:joint_key}) computes $N.f$, while \textsc{PathOptimize} separately supplies $N.g$ and $N.\tau$.

\begin{lemma}[Joint Search-Key Bounds]\label{lem:joint_key}
The joint evaluation in Eqn.~(\ref{eqn:joint_key}) is admissible when $\epsilon=1$, with
\[
N.g+h_{\mathrm{mot}}(N.\pi)\leq N.f\leq J^*(N.\pi).
\]
For any $\epsilon\geq1$, $N.f\leq\epsilon J^*(N.\pi)$ and $N.f=N.g$ at a goal node, preserving the $\epsilon$-optimality guarantee of Theorem~\ref{thm:cost_optimality}.
\end{lemma}
\begin{sketch_proof}{App.~\ref{app:joint_key}}
Separate evaluation allows prefix and heuristic to choose different interface states, while joint evaluation requires the same state, giving the first inequality.
The motion-only term underestimates the remaining duration, and any $N.\pi$-consistent solution supplies feasible prefix variables.
Therefore, we have $N.f\leq\epsilon J^*(N.\pi)$. With $N.f=N.g$ at goals, the frontier argument of Theorem~\ref{thm:cost_optimality} applies.
\end{sketch_proof}

\section{Trajectory Occupancy Reservation}\label{sec:ecd}

This section presents the Exact Convex Decomposition (ECD) scheme for reserving piecewise-linear trajectory occupancies in an ST-GCS. 
Given the spatiotemporal occupancy of a moving object, such as a robot or a dynamic obstacle, ECD removes the occupied region from the relevant space-time convex sets and decomposes the remaining free space into convex subsets. 
The resulting ST-GCS can then be used for subsequent planning queries that must avoid the reserved trajectory in space-time.

\subsection{Spatiotemporal Trajectory Occupancy}\label{subsec:trajectory_occupancy}
We assume that each moving object, including a robot or a dynamic obstacle, is represented by a piecewise-linear center trajectory $\tau=\langle\mathbf{x}_1,\ldots,\mathbf{x}_l\rangle$ in $(m+1)$-dimensional space-time with radius $r > 0$.
At each time, the object is centered at the position $\mathbf{x}.\mathbf{p}$ of state $\mathbf{x}$ along $\tau$ and occupies a spatial collision body of hypercube
\begin{align}\label{eqn:hypercube_occ}
\boxdot (\mathbf{x}.\mathbf{p}, r)=\{\mathbf{p} \mid ||\mathbf{p}-\mathbf{x}.\mathbf{p}||_\infty \leq r\}\subseteq \mathbb{R}^{m}.
\end{align}
Let $\Psi(\mathbf{x}_i,\mathbf{x}_{i+1},r)$ be the parallelotope obtained by sweeping the spatial $\boxdot (\mathbf{x}.\mathbf{p}, r)$ along each segment $\mathbf{x}_i\rightarrow \mathbf{x}_{i+1}$ of $\tau$.
A piecewise-convex spatiotemporal occupancy for trajectory $\tau$ with object radius $r$ can be defined by
\begin{align}\label{eqn:traj_occ}
\Omega(\tau, r)=\bigcup_{i=1}^{l-1} \Psi (\mathbf{x}_i,\mathbf{x}_{i+1},r)\subseteq \mathbb{R}^{m+1}.
\end{align}

For a robot of radius $r$ and a moving object of radius $r'$, a collision occurs when their spatial interiors overlap at the same time. Boundary contact is allowed, so the minimum permitted $\ell_\infty$ distance between centers is $R=r+r'$.
When strict separation is required, we choose $R>r+r'$ to provide a positive clearance margin.
Fig.~\ref{fig:occ} illustrates two piecewise-linear trajectories and their reserved occupancies in 1D and 2D spaces.

\subsection{Exact Convex Decomposition (ECD)}\label{subsec:ecd}
ECD reserves trajectory occupancy on an arbitrary ST-GCS $G=(V,E,\mathcal{X})$ using a piecewise-linear trajectory $\tau=\langle\mathbf{x}_{1},\ldots,\mathbf{x}_{l}\rangle$ and a safe clearance parameter $R$ as input.
Each parallelotope piece $\Psi (\mathbf{x}_i,\mathbf{x}_{i+1},R)\subseteq\Omega(\tau, R)$ may intersect multiple convex sets of $G$.
On the other hand, a convex set of $G$ might also contain multiple parallelotope occupancy pieces.
To account for above cases, ECD subdivides $\tau$ and constructs an ordered sequence of vertex--segments tuples $L_v=\langle(\mathbf{x}_j,\mathbf{y}_j)\rangle_{j=1}^{\eta_v}$ for each $v\in V$, such that for every $j=1,\ldots,\eta_v$, it satisfies that $\Psi (\mathbf{x}_j,\mathbf{y}_j,R)\cap X_v\neq\emptyset$ and that $\mathbf{y}_{j}.t\leq \mathbf{x}_{j+1}.t$ if $j\neq \eta_v$.
Intuitively speaking, each sequence $L_v$ collects the time-sorted subdivided segments of $\tau$ whose occupancies intersect with $X_v\in\mathcal{X}$.
Writing $\mathcal{L}$ for all collected vertex--segment tuples, we have
\begin{align*}
\bigcup_{(v, \mathbf{x},\mathbf{y})\in \mathcal{L}} \Psi (\mathbf{x},\mathbf{y},R)=\Omega(\tau, R)
\end{align*}

Alg.~\ref{alg:ecd} summarizes the ECD scheme. 
It starts by iterating each vertex $v\in V$ and collects the corresponding vertex--segments tuple sequence $L_v$ from $\tau$ (Lines~\ref{alg:ecd:iterate_V}-\ref{alg:ecd:build_occ}).
If $L_v$ is not empty, ECD decomposes the part of $X_v$ satisfying the clearance $R$ (Lines~\ref{alg:ecd:iterate_occ}--\ref{alg:ecd:add_residuals}) and removes the original $v$ and $X_v$ from $G'$ (Line~\ref{alg:ecd:remove}).
For each segment $\mathbf{x}_j\rightarrow\mathbf{y}_j$ from $L_v$, it slices only the time-bounded subset $Y\subseteq X_v$ (Line~\ref{alg:ecd:time_restricted_Y}).
For each lateral facet of $\Psi(\mathbf{x}_j,\mathbf{y}_j,R)$ (Line~\ref{alg:ecd:iterate_facet}), ECD retains the closed outer halfspace in $Y$ (Line~\ref{alg:ecd:slice_side}) and updates the residual on the occupancy side (Line~\ref{alg:ecd:exclude_outside}).
By convention, each facet $(\mathbf{A},\mathbf{b})$ is formed such that the convex set $\{\mathbf{x}\mid\mathbf{A}\mathbf{x}\preceq \mathbf{b}\}$ contains parallelotope $\Psi(\mathbf{x}_j,\mathbf{y}_j, R)$.
After the subdivision by each time-bounded $Y$, there remain three types of $X_v$ residuals that are outside the interiors of the time-bounded subsets $Y$ and thus need to be added back to $G'$ if nonempty (Line~\ref{alg:ecd:add_residuals}). 
We define the three types of $X_v$ residuals as follows:
\begin{subequations}\label{eqn:residuals_Xv}
\begin{align}
&X_v\cap\{\mathbf{x}\mid \mathbf{x}.t\leq\mathbf{x}_1.t\},\\
&X_v\cap\{\mathbf{x}\mid \mathbf{y}_{j-1}.t\leq\mathbf{x}.t\leq\mathbf{x}_{j}.t\}, &&\forall j=2,\ldots,\eta_v,\\
&X_v\cap\{\mathbf{x}\mid \mathbf{x}.t\geq\mathbf{y}_{\eta_v}.t\}
\end{align}
\end{subequations}

Finally, ECD rebuilds the adjacency of $G'$ by checking set intersections (Line~\ref{alg:ecd:update_edges}), and returns the updated ST-GCS $G'$ (Line~\ref{alg:ecd:return}).
The ECD scheme is quite general and applies to various other motions that can be represented by a piecewise-linear space-time trajectory.
For example, waiting or in-place turning can be encoded as a segment $(\mathbf{p},t)\to (\mathbf{p},t')$ along $\tau$ and being reserved in the input ST-GCS.
More importantly, as we will see in Sec.~\ref{sec:mrmp}, pre-start and post-arrival waiting segments should be inserted into the beginning and end of every single-robot trajectory $\tau$, respectively.
As a result, when planning on the ST-GCS with $\tau$ reserved by ECD, the other robots must avoid the robot that stays at the start and goal positions before and after executing $\tau$.
\section{Multi-Robot Motion Planning}\label{sec:mrmp}

This section integrates spatiotemporal planning on ST-GCSs and ECD reservation for MRMP. The low-level solver solves single-robot queries one at a time with the BFS solver in Sec.~\ref{sec:st_search}, while the high-level coordinator resolves inter-robot conflicts by deciding which robot trajectories are reserved in each low-level call.

\subsection{Problem Definition}\label{subsec:mrmp_def}

We consider $n$ robots with the same radius $r$ sharing a given base ST-GCS $G_0=(V_0,E_0,\mathcal{X}_0)$, which represents the collision-free space-time region before inter-robot reservations and may be constructed, for example, by extruding a spatial GCS through time and then applying the ECD scheme to remove occupancies of known dynamic obstacles.
Each robot $i=1,2,\ldots,n$ has a query $(\mathbf{x}_{s,i},\mathbf{p}_{g,i})$ with $\mathbf{x}_{s,i}=(\mathbf{p}_{s,i},t_{s,i})\in\mathbb{R}^{m+1}$ and prescribed start time $t_{s,i}\in[0,t_{\max}]$. The arrival time at $\mathbf{p}_{g,i}$ is unconstrained.

A feasible solution of any query $(\mathbf{x}_{s,i},\mathbf{p}_{g,i})$ on ST-GCS is a piecewise-linear trajectory $\tau_i=\langle\mathbf{x}_1,\ldots,\mathbf{x}_l\rangle$, where $\mathbf{x}_1=\mathbf{x}_{s,i}$ and $\mathbf{x}_l.\mathbf{p}=\mathbf{p}_{g,i}$.
When $t_{s,i}\neq 0$ or $\mathbf{x}_{l}.t\neq t_{\max}$, the pre-start or post-arrival staying motions are absent in $\tau_i$.
Planning on an ECD-reserved ST-GCS with such $\tau_i$ could lead to collisions on $\mathbf{p}_{s,i}$ during $t\in[0,t_{s,i}]$ or on $\mathbf{p}_{g,i}$ during $t\in[\mathbf{x}_{l}.t,t_{\max}]$.
Therefore, we let $\tilde{\tau}_i$ denote its \textit{endpoint-augmented} variant, defined as follows:
\begin{align*}
\tilde{\tau}_i=\Big\langle(\mathbf{x}_1.\mathbf{p},0),\mathbf{x}_1,\ldots,\mathbf{x}_l,(\mathbf{x}_l.\mathbf{p},t_{\max})\Big\rangle,
\end{align*}
Let $\mathbf{p}_i(t)$ denote the center of robot $i$ along $\tilde{\tau}_i$. A trajectory set $\mathcal{T}=\{\tau_i\}_{i=1}^n$ is feasible if and only if each $\tau_i$ is a feasible single-robot trajectory for query $(\mathbf{x}_{s,i},\mathbf{p}_{g,i})$ on $G_0$, and every robot pair satisfies
\begin{align*}
\|\mathbf{p}_i(t)-\mathbf{p}_j(t)\|_\infty \geq R,\quad \forall\, i<j,\ \forall\, t\in[0,t_{\max}].
\end{align*}
We use the same clearance $R$ for conflict detection and reservation, with $R=2r$ by default.

\begin{algorithm}[t]
\DontPrintSemicolon
\small
\caption{Exact Convex Decomposition (ECD)}\label{alg:ecd}
\SetKwInput{KwInput}{Input}
\SetKwInput{KwOutput}{Output}
\KwInput{ST-GCS $G=(V,E,\mathcal{X})$, piecewise-linear trajectory $\tau$, safe clearance $R$}
\KwOutput{updated ST-GCS removing the occupancy of $\tau$}
$G'=(V',E',\mathcal{X}')\gets$ a copy of $G=(V,E,\mathcal{X})$\;
\ForEach{$v\in V$}{\label{alg:ecd:iterate_V}
    $L_v\gets$ vertex--segments tuple sequence of $\tau$ for $v$\;\label{alg:ecd:build_occ}
    \If{$L_v$ is empty}{
        \textbf{continue}\;
    }
    \ForEach{$(\mathbf{x}_j,\mathbf{y}_j)\in L_v$}{\label{alg:ecd:iterate_occ}
        $Y\gets X_v\cap\{\mathbf{x}\mid \mathbf{x}_j.t\leq\mathbf{x}.t\leq\mathbf{y}_j.t\}$\;\label{alg:ecd:time_restricted_Y}
        \ForEach{$(\mathbf{A},\mathbf{b})\in\text{the facets of }\Psi(\mathbf{x}_j,\mathbf{y}_j,R)$}{\label{alg:ecd:iterate_facet}
                $X\gets Y\cap \{\mathbf{x}\mid \mathbf{A}\mathbf{x}\succeq \mathbf{b}\}$\;\label{alg:ecd:slice_side}
                $Y\gets Y\cap \{\mathbf{x}\mid \mathbf{A}\mathbf{x}\preceq \mathbf{b}\}$\;\label{alg:ecd:exclude_outside}
                add vertex $v'$ to $V'$ and set $X_{v'}=X$ to $\mathcal{X}'$\;\label{alg:ecd:add_vert}
            }
    }
    add the $X_v$ residuals defined in Eqn.~(\ref{eqn:residuals_Xv}) to $G'$\;\label{alg:ecd:add_residuals}
    remove $v$ and $X_v$ from $G'$\;\label{alg:ecd:remove}
}
$E'\gets \{(u,w)\mid u,w \in V',\; u\neq w,\; X_u\cap X_w\neq\emptyset\}$\;\label{alg:ecd:update_edges}
\Return updated ST-GCS $G'=(V',E',\mathcal{X}')$\;\label{alg:ecd:return}
\end{algorithm}

The MRMP problem seeks to find such a feasible trajectory set described above. 
We evaluate MRMP solution quality using sum-of-costs (SOC) and makespan:
\begin{align*}
\mathrm{SOC}(\mathcal{T})=\sum_{i=1}^n c(\tau_i),\,
\mathrm{Makespan}(\mathcal{T})=\max_{i=1,\ldots, n} c(\tau_i),
\end{align*}
Here, $c(\tau_i)=t_{g,i}-t_{s,i}$ includes all waiting between the prescribed start and goal arrival, where $t_{g,i}$ denotes the goal arrival time. With the common start time $t_{s,i}=0$, makespan equals $\max_i t_{g,i}$, the elapsed time until all robots reach their goals. For unequal start times, the overall schedule duration is instead $\max_i t_{g,i}-\min_i t_{s,i}$, which can differ from the maximum individual duration.

For notation, let $\operatorname{ECD}(G,\tau,R)$ denote the ST-GCS returned by Alg.~\ref{alg:ecd}.
For an ordered list of trajectories, $\operatorname{ECD}(G,\langle \tau_1,\ldots,\tau_l\rangle,2r)$ denotes the ST-GCS obtained by applying Alg.~\ref{alg:ecd} successively to $\tau_1,\ldots,\tau_l$ in that order, with $\operatorname{ECD}(G,\langle\rangle,2r)=G$.
It is worth noting that, for MRMP where the robots have different radii, we only need to construct an individual base $G_0$ for each robot and change the ECD clearance parameter accordingly, while the low-level single-robot solver and the following high-level coordination algorithms remain unchanged.

\subsection{Prioritized Planning with ST-GCS and ECD}
We instantiate Prioritized Planning (PP)~\citep{erdmann1987multiple} in the ST-GCS setting using the BFS solver in Alg.~\ref{alg:st_search} for low-level single-robot planning and ECD for trajectory reservation.
We consider a total priority order $\pmb{\sigma}=\langle \sigma_1, \dots, \sigma_n\rangle$, where $\pmb{\sigma}$ is a permutation of the robot indices $1,2,\ldots,n$.
Robot $\sigma_i$ has higher priority than robot $\sigma_j$ if $i<j$.
Let $G_{0,i}$ denote the base graph for robot $i$. Ordinarily $G_{0,i}=G_0$. For the completeness guarantees below, we choose a base graph that reserves every other robot's start and goal from time zero through $t_{\max}$. Let $\mathcal S_i(0)$ contain the stationary segment $\langle(\mathbf p,0),(\mathbf p,t_{\max})\rangle$ for each $\mathbf p\in\{\mathbf p_{s,j},\mathbf p_{g,j}\}$ and each $j\ne i$. The resulting graph is
\[
\widehat G_i=\operatorname{ECD}(G_0,\mathcal S_i(0),2r).
\]
These reservations apply even while an endpoint is unoccupied. With either choice of base graph, PP solves the query of robot $\sigma_k$ at step $k$ on
\begin{align}\label{eqn:def_Gk}
G^{(k)}=\operatorname{ECD}\bigl(G_{0,\sigma_k},
\langle\tilde{\tau}_{\sigma_1},\ldots,\tilde{\tau}_{\sigma_{k-1}}\rangle,2r\bigr).
\end{align}
The first robot uses its base graph directly. If any BFS call fails, PP reports failure for this order. Otherwise it returns all $n$ trajectories.

\begin{definition}[Well-Formedness]\label{def:mrmp_well_formed}
An MRMP instance is \emph{well-formed} if there is a time $t_0\in[\max_i t_{s,i},t_{\max}]$ such that the spatial free space represented by $G_0$ is time-invariant on $[t_0,t_{\max}]$ and the following conditions hold.
\begin{enumerate}[(i)]
\item\label{wellform:separation} Any two endpoints belonging to different robots have $\ell_\infty$ distance greater than $2r$.
\item\label{wellform:staying} For each robot, both its start and its goal stay collision-free on $G_0$ throughout $[0,t_{\max}]$.
\item\label{wellform:feasibility} For each robot $i$, the query $((\mathbf{p}_{s,i},t_0),\mathbf{p}_{g,i})$ is feasible on $\widehat G_i$. Let $C_i^*$ be its minimum travel time.
\end{enumerate}
\end{definition}

For a spatial GCS extruded through time, one can take $t_0=\max_i t_{s,i}$ in Definition~\ref{def:mrmp_well_formed}. If dynamic obstacles settle later, one can take $t_0$ when they reach their fixed positions, provided the endpoint and route conditions hold.
Since the free space is time-invariant after $t_0$, robot $i$ can delay its endpoint-avoiding route until higher-priority robots reach their goals and still traverse it in time $C_i^*$.

Fix a time $t_0$ satisfying Definition~\ref{def:mrmp_well_formed} for the following bounds. Under the assumptions of Theorem~\ref{thm:cost_optimality}, our BFS solver with inflation $\epsilon\geq1$ returns a trajectory for every feasible query. Its elapsed time, including waiting, is at most $\epsilon$ times the minimum. Applying this guarantee in priority order bounds the arrival times of the first $k$ robots by $B_k(\epsilon)$, where $C_{\max}^*=\max_i C_i^*$ and $0\leq k\leq n$:
\[
B_k(\epsilon)=
\left\{\begin{array}{@{}l@{\;}l@{}}
t_0+kC_{\max}^*, &\epsilon=1,\\[2pt]
\epsilon^k t_0+(\epsilon^k-1)\bigl(\dfrac{\epsilon C_{\max}^*}{\epsilon-1}-\min_i t_{s,i}\bigr), &\epsilon>1.
\end{array}\right.
\]
Thus, $t_{\max}\geq B_n(\epsilon)$ allows enough time for all $n$ robots to reach their goals.

\begin{theorem}[Completeness of PP]\label{thm:pp_complete}
Consider a well-formed instance. Suppose PP uses $G_{0,i}=\widehat G_i$ for each robot $i$, and BFS returns an $\epsilon$-optimal solution for every feasible low-level query. If $t_{\max}\geq B_n(\epsilon)$, PP returns a feasible MRMP solution for every total priority order.
\end{theorem}
\begin{sketch_proof}{App.~\ref{app:pp_complete}}
Every higher-priority trajectory avoids the current robot's start. The robot can therefore wait until all higher-priority robots are at their goals and the free space is time-invariant, then take its endpoint-avoiding route. Applying the low-level cost bound to this feasible trajectory bounds the $k$th robot's actual arrival time by $B_k(\epsilon)$.
\end{sketch_proof}

\subsection{PBS with ST-GCS and ECD}
We instantiate Priority-Based Search (PBS)~\citep{ma2019searching} in the ST-GCS setting by using the BFS solver in Alg.~\ref{alg:st_search} for low-level single-robot planning and ECD for reserving spatiotemporal occupancies.
PBS searches over partial priority orders instead of committing to one total priority order. 
A PBS node $N$ stores a set $\boldsymbol{\pmb\prec}_N$ of ordered robot pairs and a trajectory set $N.\mathcal{T}$.
Let ${\prec}_N^+$ denotes the \textit{transitive closure} of ${\pmb\prec}_N$ such that if $i\prec^+_N j$, then robot $j$ must avoid robot $i$.
Namely, its trajectory $N.\tau_j$ can be obtained by planning on
\begin{align}\label{eqn:reserve_high_priority}
G^{(j)}_N=\operatorname{ECD}(G_{0,j},\langle N.\tilde{\tau}_{i_1},\ldots,N.\tilde{\tau}_{i_l}\rangle,2r),
\end{align}
where $\langle i_1,\ldots,i_l\rangle$ is any fixed topological ordering of the set $\{i\mid i\prec_N^+ j\}$ of all high-priority robots.
Here, $N.\tilde{\tau}_i$ denotes the endpoint-augmented reservation trajectory corresponding to the stored trajectory $N.\tau_i$.
At the root, robot $j$ plans on $G_{0,j}$ as it has no higher-priority robots. In a child node, replanning starts from the same $G_{0,j}$ and adds the trajectories of all current ancestors of $j$.

Alg.~\ref{alg:wpbs} summarizes PBS, starting with an empty partial priority set and independently planned trajectories (Lines~\ref{alg:wpbs:init_root}-\ref{alg:wpbs:root_plan}). PBS performs a depth-first search on a priority tree using STACK. 
When PBS selects a node $N$ from STACK, if the endpoint-augmented trajectories corresponding to $N.\mathcal{T}$ are pairwise collision-free, PBS returns it (Line~\ref{alg:wpbs:sol_found}).
Otherwise, it identifies a colliding robot pair $i_1,i_2$ whose endpoint-augmented trajectories $N.\tilde{\tau}_{i_1}$ and $N.\tilde{\tau}_{i_2}$ collide (Line~\ref{alg:wpbs:1st_col}).
PBS then creates two child nodes by orienting this collision in both directions. One child adds $i_1\prec i_2$, and the other adds $i_2\prec i_1$ (Lines~\ref{alg:wpbs:for_child}-\ref{alg:wpbs:update_prec}). 
For any branch $(i,j)$, the child node $N'$ adds $i\prec j$ and calls \textsc{UpdateNode}$(N',j)$, where $j$ is the newly lower-priority robot.
\textsc{UpdateNode} builds the replanning set $K$, containing $j$ and all robots whose higher-priority constraints may change after adding $i\prec j$ (Line~\ref{alg:wpbs:replan_list}).
It then processes each robot $k\in K$ in a topological ordering of $\boldsymbol{\pmb\prec}_{N'}$.
If $N'.\tilde{\tau}_k$ collides with $N'.\tilde{\tau}_{i'}$ of some higher-priority robot $i'\prec_{N'}^+ k$, \textsc{UpdateNode} replans $\tau_k$ to respect all of higher-priority reservations of robot $k$ (Lines~\ref{alg:wpbs:ecd}--\ref{alg:wpbs:replan}).
If \textsc{UpdateNode} succeeds, PBS adds the child node $N'$ to STACK (Lines~\ref{alg:wpbs:update_node}--\ref{alg:wpbs:add_child}).
PBS returns failure if no valid solution can be found after exploring all nodes (Line~\ref{alg:wpbs:fail}).

\begin{theorem}[Completeness of PBS]\label{thm:pbs_complete}
Consider a well-formed instance. Suppose PBS uses $G_{0,i}=\widehat G_i$ for each robot $i$ at the root and at every replan, and BFS returns an $\epsilon$-optimal trajectory for every feasible low-level query. If $t_{\max}\geq B_n(\epsilon)$, PBS returns a feasible MRMP solution.
\end{theorem}
\begin{sketch_proof}{App.~\ref{app:pbs_complete}}
If the longest priority chain ending at a robot contains $d$ robots, the waiting argument bounds its arrival time by $B_d(\epsilon)$. This bound holds for both replanned and inherited trajectories. Each child resolves a conflict between previously incomparable robots, so the finite priority search reaches a collision-free node.
\end{sketch_proof}

\begin{algorithm}[t]
\DontPrintSemicolon
\small
\caption{PBS with ST-GCS}\label{alg:wpbs}
\SetKwInput{KwInput}{Input}
\SetKwInput{KwOutput}{Output}
\renewcommand{\texttt}[1]{\scshape{\textsc{#1}}}
\KwInput{queries and base ST-GCSs $\{(\mathbf{x}_{s,i},\mathbf{p}_{g,i},G_{0,i})\}_{i=1}^n$}
\KwOutput{collision-free trajectory set}
create root node $N_\text{root}$ with $\boldsymbol{\pmb\prec}_{N_\text{root}}\gets\emptyset$\;\label{alg:wpbs:init_root}
$N_\text{root}.\mathcal{T}\gets$ solve each query $i$ on $G_{0,i}$ using Alg.~\ref{alg:st_search}\;\label{alg:wpbs:root_plan}
STACK $\gets\{N_\text{root}\}$\;
\While{\text{\normalfont STACK} $\neq\emptyset$}{
    $N\gets$ STACK.pop()\;
    \If{$\{N.\tilde{\tau}_i\}_{i=1}^{n}$ is pairwise collision-free}{\label{alg:wpbs:col_check}
        \Return $N.\mathcal{T}$\;\label{alg:wpbs:sol_found}
    }
    $i_1,i_2\gets$ robot pair with colliding $N.\tilde{\tau}_{i_1}$ and $N.\tilde{\tau}_{i_2}$\;\label{alg:wpbs:1st_col}
    \For{$(i,j)\in\{(i_1,i_2),(i_2,i_1)\}$}{\label{alg:wpbs:for_child}
        $N'\gets$ a copy of $N$ with $\boldsymbol{\pmb\prec}_{N'}\gets\boldsymbol{\pmb\prec}_{N}\cup\{i\prec j\}$\;\label{alg:wpbs:update_prec}
        \If{\text{\normalfont \texttt{UpdateNode}}$(N', j)$}{\label{alg:wpbs:update_node}
             STACK.add($N'$)\;\label{alg:wpbs:add_child}
        }
    }
}
\Return ``\textit{no solution found}''\;\label{alg:wpbs:fail}
\SetKwFunction{FMain}{UpdateNode}
\SetKwProg{Fn}{Function}{:}{}
\Fn{\FMain{$N', j$}}{
    $K\gets\{k \mid k=j \text{ or } j\prec_{N'}^+ k\}$\;\label{alg:wpbs:replan_list}
    \ForEach{$k\in \text{\normalfont TopologicalSort}(K,\boldsymbol{\pmb\prec}_{N'})$}{
        \If{$\exists\, i'\prec_{N'}^+ k: N'.\tilde{\tau}_{i'}$\text{ colliding with} $N'.\tilde{\tau}_k$}{
            construct $G^{(k)}_{N'}$ (Eqn.~(\ref{eqn:reserve_high_priority})) via ECD\Comment{Alg.~\ref{alg:ecd}}\;\label{alg:wpbs:ecd}
            solve query $(\mathbf{x}_{s,k},\mathbf{p}_{g,k})$ on $G^{(k)}_{N'}$ using Alg.~\ref{alg:st_search}\;\label{alg:wpbs:replan}
            \eIf{solving finds a feasible solution $\tau$}{
                $N'.\tau_{k}\gets\tau$\;
            }{
                \Return \textbf{False}\;
            }
        }
    }
\Return \textbf{True}\;
}
\end{algorithm}

\subsubsection{Node Evaluation and Ordering Rules}\label{subsec:child-node-exp}
PBS can vary how child nodes are generated and expanded (Lines~\ref{alg:wpbs:for_child}--\ref{alg:wpbs:add_child} of Alg.~\ref{alg:wpbs}). A node is \emph{generated} once its new priority relation is added, and it is \emph{expanded} once \textsc{UpdateNode} has replanned the affected robots under that relation. 
With lazy evaluation, both child nodes are generated before calling \textsc{UpdateNode}; expansion is delayed until a generated child is selected for evaluation. This avoids spending low-level planning effort on generated nodes that may never be expanded.

Alternatively, PBS can expand both child nodes immediately and order the expanded children by a metric. 
A metric of sum-of-costs or makespan favors branches with better current solution quality, while a metric of the number of pairwise conflicts favors less congested branches.
These metric-based rules require immediate node expansion via \textsc{UpdateNode} and therefore are not compatible with lazy evaluation. 
Since Alg.~\ref{alg:wpbs} uses a stack, the favored expanded child is pushed after the other child so that it is selected first.

\subsection{Reuse of Base ST-GCS Heuristic Values}\label{subsec:mrmp_heuristic_reuse}

The low-level solver in PP or PBS is repeatedly invoked on ECD-updated ST-GCSs. 
The motion-only $h_\text{mot}$ is evaluated directly on any ST-GCSs using the actual prefix interface $I_N$. 
In contrast, $h_\text{tri}$ and $h_\text{tab}$ rely on heuristic values computed on the base ST-GCS $G_0=(V_0,E_0,\mathcal{X}_0)$, and we reuse these values on ECD-updated ST-GCSs via the corresponding base vertices and interfaces.
Let $G=(V,E,\mathcal{X})$ be any ST-GCS obtained from $G_0$ by a finite sequence of ECD reservations.
Consider any query-updated $(G^\sharp,v_s,v_g)=\textsc{QueryAugment}(G,\mathbf{x}_s,\mathbf{p}_g)$.
For any non-query vertex $v\in V$, let $\operatorname{base}(v)\in V_0$ denote the base vertex of $G_0$ from which $v$ descends, so that $X_v\subseteq X_{\operatorname{base}(v)}$. 
This map is inherited through ECD subdivisions. 
Initially $\operatorname{base}(v)=v$ for $v\in V_0$, and when ECD subdivides a vertex $v$ into new vertices, each new vertex $w$ is assigned $\operatorname{base}(w)=\operatorname{base}(v)$. 
Specifically, we define $\operatorname{base}(v_s)=v_s$ and $\operatorname{base}(v_g)=v_g$ for the auxiliary query vertices.

Let $h\in\{h_{\mathrm{tri}},h_{\mathrm{tab}}\}$ and evaluate it on the base graph $G_0^\sharp$ augmented with the same query. For a prefix ending in a non-query base edge $u\to v$, the value $h(\langle\ldots,u,v\rangle)$ from Sec.~\ref{subsec:st_planning_heur} depends only on that edge and the query: it lower-bounds the cost from $X_u\cap X_v$ to the goal, without restricting previously visited base vertices. The root and the first edge out of $v_s$ use the query-start rules of Sec.~\ref{subsec:st_planning_heur}, and the value at $v_g$ is zero.
For any other non-goal prefix ending in $u\to v$, we use $h(\langle\ldots,\operatorname{base}(u),\operatorname{base}(v)\rangle)$ if $\operatorname{base}(u)\ne\operatorname{base}(v)$. Otherwise, the prefix has already entered $\operatorname{base}(v)$. We omit the cost of reaching the next base exit and use
\[
\begin{dcases}
0, & (\operatorname{base}(v),v_g)\in E_0^\sharp,\\
\min_w h(\langle\ldots,\operatorname{base}(v),w\rangle), & \text{otherwise},
\end{dcases}
\]
where $E_0^\sharp$ is the edge set of $G_0^\sharp$ and the minimum ranges over all $w$ with $(\operatorname{base}(v),w)\in E_0$. An empty minimum is $+\infty$. Lemma~\ref{lem:mrmp_heuristic_reuse} in App.~\ref{app:heuristic_admissibility} establishes admissibility.

\subsection{Windowed Coordination}\label{subsec:windowed_coord}
Full-horizon coordination can enlarge ECD-updated ST-GCSs by reserving long trajectory portions unrelated to near-term interactions. Windowed coordination limits this growth by restricting PP or PBS reservations as well as PBS conflict detection to a planning window $W=[t,t+\Delta t_{\mathrm{plan}}]$, with span $\Delta t_{\mathrm{plan}}>0$ and execution horizon $0<\Delta t_{\mathrm{exec}}\leq\Delta t_{\mathrm{plan}}$.

At time $t$, PP or PBS solves each robot's query $(\mathbf{x}_{s,i}^{W},\mathbf{p}_{g,i})$ from its current window-start state, initially its original start. 
Each successful low-level call returns a trajectory to the original goal, possibly arriving beyond $W$, while inter-robot coordination is restricted to $W$. 
Goal-reached robots remain in planning and may temporarily leave their goals and return if new priorities ask it to make way for other higher-priority robots.

Let $\tau|_W$ denote a trajectory clipped to $W$, with segment endpoints clipped at the window boundaries as needed. 
Robots $i\neq j$ conflict within $W$ if and only if
\[
\|\mathbf{p}_i(s)-\mathbf{p}_j(s)\|_\infty < R \quad \text{for some } s\in W.
\]
Windowed ECD still refines $G_0$ by excluding reserved occupancy, so base-graph heuristic reuse remains admissible by Lemma~\ref{lem:mrmp_heuristic_reuse}.

Following Eqn.~(\ref{eqn:def_Gk}), windowed-PP starts with $G_W^{(1)}=G_0$ and plans robot $\sigma_k$ on $G_W^{(k)}$ in the fixed priority order $\pmb{\sigma}$. After each successful plan, it updates
\[
G_W^{(k+1)}=\operatorname{ECD}(G_W^{(k)},\tilde{\tau}_{\sigma_k}|_W,2r).
\]
Following Eqn.~(\ref{eqn:reserve_high_priority}), windowed-PBS replans robot $i$ at node $N$ on
\[
\operatorname{ECD}(G_0,\langle N.\tilde{\tau}_{j_1}|_W,\ldots,N.\tilde{\tau}_{j_l}|_W\rangle,2r),
\]
where $\langle j_1,\ldots,j_l\rangle$ is any fixed topological ordering of $\{j\mid j\prec_N^+ i\}$. 
We also introduce an additional branching rule for PBS to dynamically adjust the search direction. 
For a conflict between a goal-reached robot and an unreached robot, it first considers the child prioritizing the unreached robot, and otherwise follows Sec.~\ref{subsec:child-node-exp}.

Once PP or PBS finds trajectories that are collision-free within $W$, each robot's prefix over $[t,t+\Delta t_{\mathrm{exec}}]$ is committed and appended to its global trajectory, and the remaining suffix is discarded. The prefix endpoint becomes its next window-start state, from which the coordinator replans with updated reservations or dynamic obstacles. 
The overall coordination process terminates successfully only when all robots are at their goals at the end of a committed prefix. 
This receding-horizon cycle commits the entire window when $\Delta t_{\mathrm{exec}}=\Delta t_{\mathrm{plan}}$; otherwise, the uncommitted lookahead suffix guides the current solve. For fixed $\Delta t_{\mathrm{plan}}$, a smaller $\Delta t_{\mathrm{exec}}$ allows earlier adjustments that may improve subsequent coordination with more frequent planning.

When $t+\Delta t_{\mathrm{plan}}\geq t_{\max}$, reservations and conflict detection cover the entire remaining horizon, recovering the full-horizon coordination constraints for the current queries.
For shorter windows, coordination is incomplete in general and does not guarantee eventual goal reaching since conflicts beyond $W$ are deferred.
However, reserving fewer trajectory segments per low-level call keeps ECD-updated ST-GCSs smaller and greatly reduces computation.
This in practice can lead to higher success rates under limited runtime budgets (Sec.~\ref{subsec:windowed_ablation}).

\subsubsection{Dynamic Window Adjustment}\label{subsec:dynamic_window_adj}
To expose deferred conflicts before committing another prefix, we enlarge the planning window when a coordination call fails or, for PBS, repeatedly returns the same topological priority order among the same unreached robots without meaningful progress. The coordinator doubles $\Delta t_{\mathrm{plan}}$ and retries the same step without committing new prefixes, exposing more future interactions and reserving longer trajectory portions. Windowed-PP retains its fixed priority order during the retry. After a successful step, $\Delta t_{\mathrm{plan}}$ is reset to its nominal value.

\section{Numerical Results}\label{sec:exp}
This section shows our numerical results for spatiotemporal planning and MRMP on ST-GCS.
We implemented the proposed planners and algorithmic components in \textit{C++} and evaluated them on an \textit{Apple}\textsuperscript{\textregistered} M4 CPU machine with 16GB RAM.
The GCS-related trajectory optimization used \textit{Drake}~\citep{drake} library and the \textit{Mosek} solver~\citep{mosek}.
The source code and numerical results are publicly available on \url{https://github.com/reso1/stgcs}.
More detailed visualizations and simulation videos of the proposed planners can be found at \url{https://sites.google.com/view/stgcs}.

\subsection{Experiment Setup}
We considered three two-dimensional base domains, \textit{rand}, \textit{maze}, and \textit{iris}, to define the problem domains for benchmarking the spatiotemporal and MRMP planners.
For each ST-GCS instance, we initialized a spatial GCS as in Sec.~\ref{subsec:gcs_generation}, and extruded the spatial sets through time $t\in[0,100]$ (i.e., $t_\text{max}=100$).
We precomputed the heuristics $h_\text{tri}$ and $h_\text{tab}$ on these base ST-GCSs.

In this section, violin plots use shaded shapes to show estimated distributions, dots to show per-instance values, and thick black vertical bars to mark medians.

\subsubsection{Spatial GCS Generation}\label{subsec:gcs_generation}
For \textit{rand}, we generated a square grid with $4$ cells along each axis, sampled a random convex polygon around each grid cell, connected intersecting convex sets, and retained the largest connected component.
For \textit{maze}, we first generated a random $12 \times 12$ maze via recursive division~\citep{buck2011recursivedivision} and then merged adjacent free cells into larger axis-aligned rectangles whenever possible using a bipartite matching procedure similar to that of~\cite{lu2023tmstc}.
For \textit{iris}, we sampled $6$ polygonal static obstacles in a square workspace and grew a connected cover of collision-free convex sets using visibility clique cover~\citep{werner2024approximating} with IRIS region inflation~\citep{deits2015computing} and a free-space coverage threshold of $90\%$.
This produced a set of more realistic GCS commonly seen in practice.

\begin{figure}
\centering
\includegraphics[width=\linewidth]{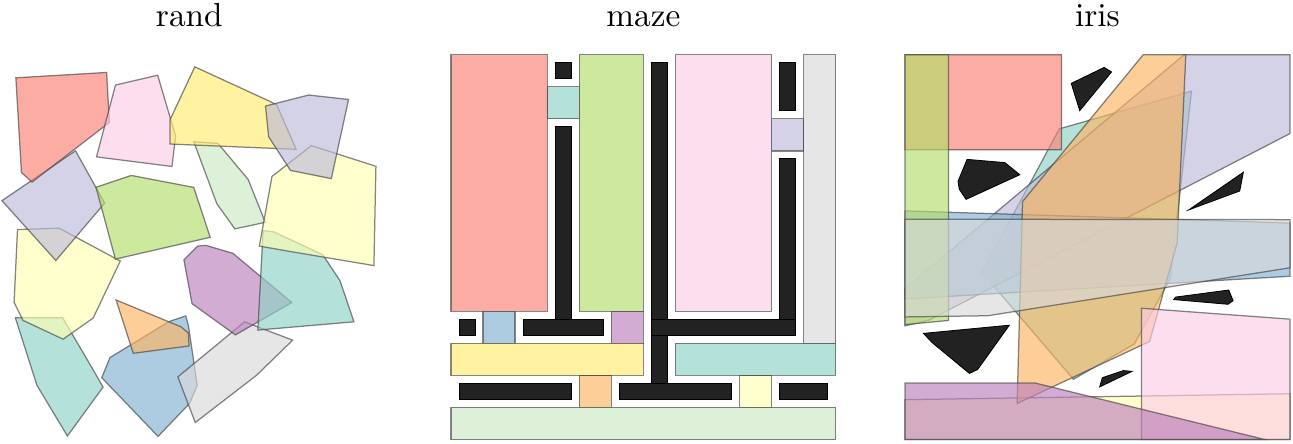}
\caption{Example instances of the generated spatial GCS. Colors indicate convex regions in the spatial decomposition. Black regions indicate environment static obstacles.}
\label{fig:exp_domains}
\end{figure}

At last, we affinely rescaled each domain’s spatial GCS regions and static obstacles to a \(5\times5\) workspace.
We set the robot radius $r=0.1$ for all domain instances.
Fig.~\ref{fig:exp_domains} shows examples of the generated spatial GCS.

\subsection{Spatiotemporal Motion Planning}
We ablated the BFS solvers (Sec.~\ref{sec:st_search}) on spatiotemporal planning instances in Sec.~\ref{subsec:st_instances}, and compared their performance with other planners.
Heuristic inflation in the BFS ablations was $\epsilon=1$ unless otherwise stated.

\subsubsection{Benchmark Instances}\label{subsec:st_instances}
We generated $150$ single-robot spatiotemporal planning instances, with $50$ instances from each domain of \textit{rand}, \textit{maze}, and \textit{iris}.
Each instance contained $10$ moving obstacles with axis-aligned square footprints of side $2r$.
Each dynamic obstacle moved periodically at constant speed along a sampled line segment within one convex region for the first $20$s, then remained at its final position until $t_\text{max}=100$s.
We applied ECD with clearance $2r$ to remove their space-time occupancies from the base ST-GCS.
Each query had random start position at time $0$ and random goal position, with velocity bound $\mathbf{v}_{\mathrm{lim}}=[1.0,1.0]$.
The resulting ST-GCSs had a median of $334$ vertices and $1901$ directed edges.
We reused the heuristic values precomputed on the base ST-GCSs before adding the dynamic obstacles.

\begin{figure}[t]
    \centering
    \includegraphics[width=\linewidth]{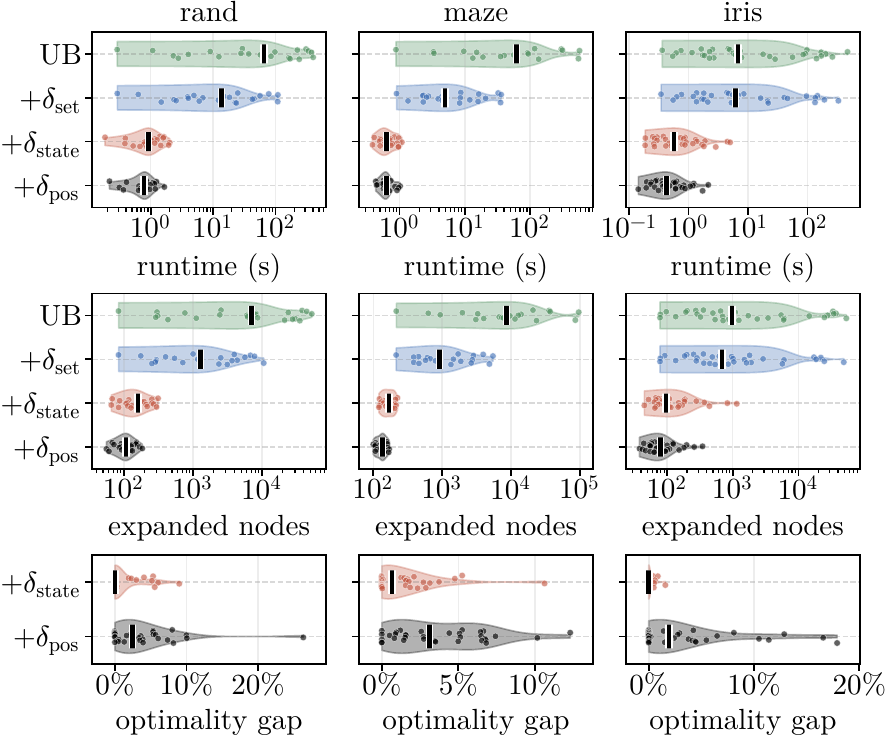}
    \caption{BFS dominance check ablation ($600$s budget). Rows show runtime, expanded-node, and optimality-gap distributions on common-success instances.}
    \label{fig:dc_ablation}
\end{figure}

\subsubsection{Ablation of BFS Dominance Checks}\label{subsec:dc_ablation}
We isolated the pairwise dominance check (Sec.~\ref{subsec:dc}) with $h_\text{max}$ as the heuritic and upper-bound pruning (UB) across variants.
We used UB alone without pairwise dominance checks as a baseline.
The alternatives were additions of sound $\delta_\text{set}$, as well as the two approximate $\delta_\text{pos}$ and $\delta_\text{state}$.
Within the $600$s budget, $\delta_\text{state}$ solved all instances, followed by $\delta_\text{pos}$ ($132$), $\delta_\text{set}$ ($126$), and UB ($86$).
On the $77$ common-success instances, Fig.~\ref{fig:dc_ablation} shows substantial runtime and node-expansion reductions with the approximate checks.
The median runtime decreased from $6.312$s with $\delta_\text{set}$ to $0.770$s with $\delta_\text{state}$ and $0.633$s with $\delta_\text{pos}$.
The corresponding median expanded-node counts decreased from $960$ to $147$ and $107$, respectively.
Compared with $\delta_\text{state}$, $\delta_\text{pos}$ reduced node expansions more substantially than runtime on these instances.
Its scalability benefit became clearer on the larger instances in App.~\ref{app:st_stress_test}, where it achieved a higher success rate.

For solution quality, we compared the $108$ instances solved by all three dominance-check variants, measuring cost increases relative to the optimal $\delta_\text{set}$ solution costs.
Optimality gaps occurred less frequently and were generally smaller with $\delta_\text{state}$ than with $\delta_\text{pos}$.
On these spatiotemporal instances, $\delta_\text{state}$ had smaller quality losses and a higher success rate, while $\delta_\text{pos}$ achieved a lower median runtime on common-success instances.

\subsubsection{Ablation of BFS Heuristics}\label{subsec:heur_ablation}
We compared the heuristics (Sec.~\ref{subsec:st_planning_heur}) with the same UB and $\delta_\text{state}$.
We used the zero heuristic $h_\text{zero}(N.\pi)=0$ for every search node $N$ as a baseline, so the search priority reduces to $N.f=N.g$.
We also included the Joint Search-Key Evaluation variant (\textit{joint}) from Sec.~\ref{subsec:joint_key}.

\begin{figure}[t]
    \centering
    \includegraphics[width=\linewidth]{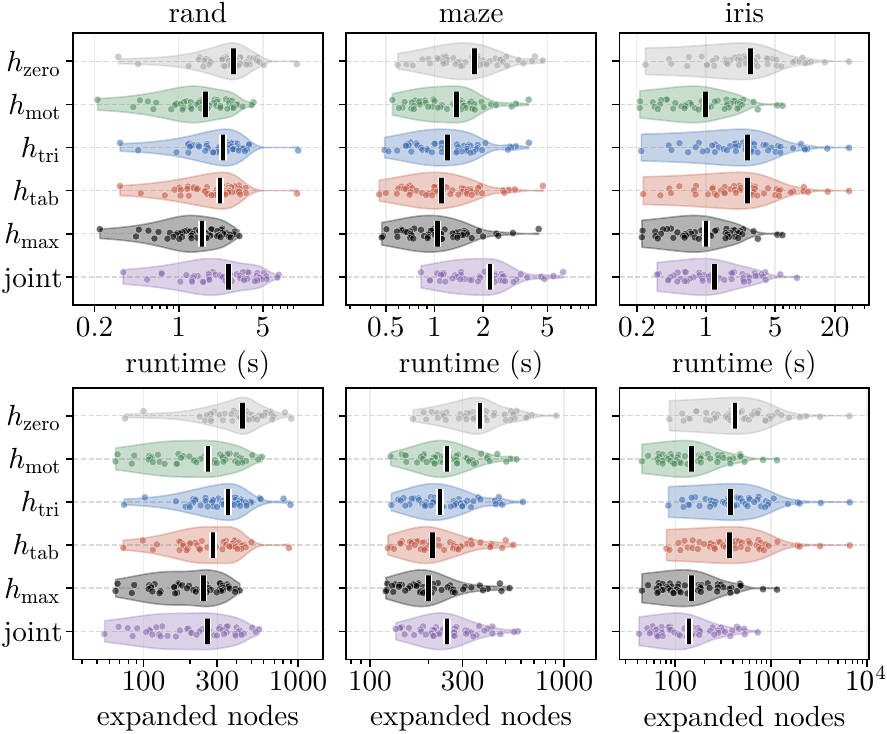}
    \caption{BFS heuristic ablation ($30$s budget), with Joint Search-Key Evaluation (\textit{joint}). Rows show runtime and expanded-node distributions on common-success instances.}
    \label{fig:heur_ablation}
\end{figure}

Within $30$s budget, all variants solved all instances in Fig.~\ref{fig:heur_ablation}.
$h_\text{max}$, which combines complementary bounds, achieved the lowest overall median runtime ($1.14$s) and expanded-node count ($188$), compared with $2.40$s and $400$ for $h_\text{zero}$.
$h_\text{mot}$ was particularly competitive in \textit{rand} and \textit{iris}, achieving the lowest median runtime and expanded-node count among the individual heuristics.
In \textit{maze}, $h_\text{tab}$ achieved the lowest values on both metrics.
Joint optimization reduced the overall median expanded-node count slightly, from $227$ with $h_\text{mot}$ to $223$, but increased median runtime from $1.37$s to $1.97$s.

We also evaluated the sensitivity of heuristic inflation in App.~\ref{app:heuristic_inflation} for all heuristics except $h_\text{zero}$ with the two approximate checks.
Among all tested combinations of heuristics and inflation factors, $h_\text{max}$ with $\epsilon=8$ achieved the lowest mean runtime for both $\delta_\text{state}$ ($0.89$s) and $\delta_\text{pos}$ ($0.90$s).
Both solved all instances within $30$s, with mean cost gaps of $9.77\%$ and $19.68\%$, respectively, relative to the lowest solution costs found.
Further inflation increased both mean runtime and cost gap.

In App.~\ref{app:heuristic_precomputation}, we separately evaluated how heuristic precomputation costs scaled with base ST-GCS size on a distinct size sweep.
The results showed that the cost grew approximately linearly for $h_\text{tri}$ but approximately exponentially for $h_\text{tab}$.
These trends support including $h_\text{tri}$ in $h_\text{max}$ while limiting $h_\text{tab}$ to relatively small ST-GCSs.
In MRMP, these costs are amortized by reusing the precomputed values across multiple rounds of low-level planning on the same base ST-GCS.

\subsubsection{Benchmark Planners}\label{subsec:st_baselines}
We evaluated the spatiotemporal planners with a $30$s runtime budget on all instances from Sec.~\ref{subsec:st_instances}.
We evaluated four BFS variants as follows.
B1 served as the optimal reference, using $h_\text{max}$ with sound $\delta_\text{set}$ and $\epsilon=1$.
B2 and B3 used $h_\text{max}$ with $\epsilon=8$ and the approximate checks $\delta_\text{state}$ and $\delta_\text{pos}$, respectively, which achieved the best empirical scalability as observed in Tab.~\ref{tab:heuristic_inflation}.
B4 used Joint Search-Key Evaluation (\textit{joint}; Sec.~\ref{subsec:joint_key}) with $\delta_\text{pos}$ and $\epsilon=2$, providing an alternative to separate prefix-cost and heuristic evaluation.
All BFS variants shared the same UB pruning.

We excluded direct optimization of the MICP in Eqns.~(\ref{eqn:stgcs}) from the comparison since it did not solve any instances in our experiments due to the large sizes of ST-GCSs.
Instead, we evaluated MICP(g) as an optimization baseline described in Sec.~\ref{subsec:solution_approaches}.
It first solves the linear relaxation of the MICP and generates $\lceil 1000\log |E| \rceil$ candidate paths through stochastic rounding.
It then invokes \textsc{PathOptimize} to compute a trajectory for each rounded path.

\begin{figure}[t]
    \centering
    \includegraphics[width=\linewidth]{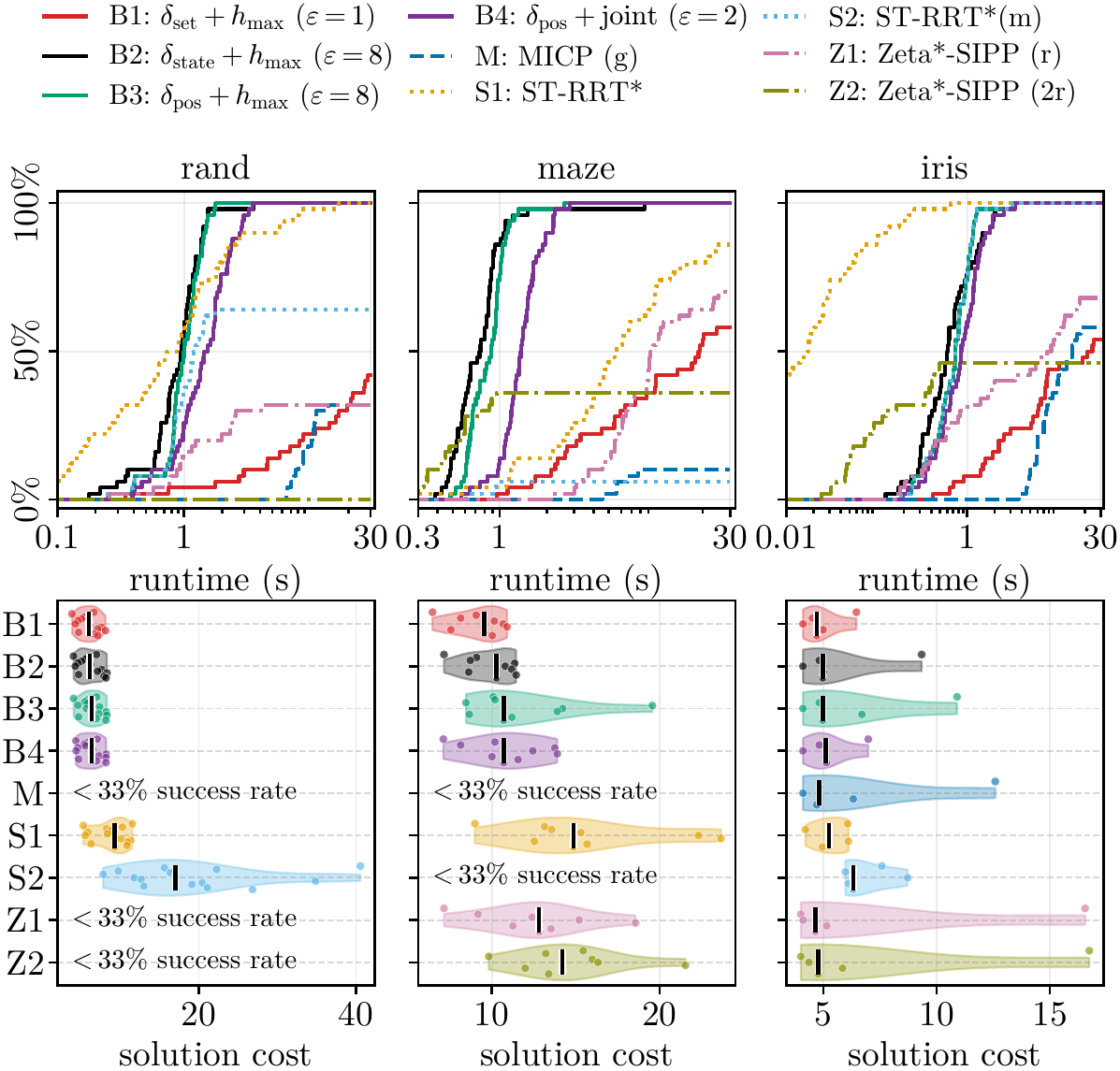}
    \caption{Spatiotemporal planner comparison ($30$s budget). The top row shows success rate over runtime. The bottom row shows solution cost distributions on common-success instances among planners with at least $33\%$ success in each domain.}
    \label{fig:st_performance_comparison}
\end{figure}

We also compared against the sampling-based baseline ST-RRT$^*$~\citep{grothe2022st} from OMPL~\citep{sucan2012the-open-motion-planning-library}, reporting its final incumbent after $30$s budget alongside the runtime-matched variant ST-RRT$^*$(m) in Fig.~\ref{fig:st_performance_comparison}.
ST-RRT$^*$(m) used the smaller of B3's runtime and $30$s as its per-instance budget.
Given the robot radius $r$, the discrete search-based Zeta$^*$-SIPP~\citep{zou2024zeta} used cell sizes of $r$ and $2r$ in all three domains, conservatively discretizing both static and dynamic obstacles on the same grid.
Query coordinates were retained without snapping to grid centers, and queries in blocked cells were rejected.

ST-RRT$^*$ and Zeta$^*$-SIPP used the same spatiotemporal planning queries, component-wise velocity limits, and planning horizon ($t_\text{max}=100$s) as the ST-GCS solvers.
For both baselines, the underlying spatial free space in \textit{rand} is the union of the convex sets.
ST-RRT$^*$ connections may cross boundaries between these sets, but any connection leaving their union is infeasible.
In \textit{maze} and \textit{iris}, the spatial free space is defined by the obstacle polygons and the robot's square footprint.

\subsubsection{Scalability}
The top row of Fig.~\ref{fig:st_performance_comparison} shows that the BFS variants B2--B4 with the approximate dominance checks solved all instances within budget.
B2 and B3 had similar runtime profiles, with B2 achieving a lower median runtime in each domain.
Joint optimization in B4 retained full success but generally took longer, consistent with the heuristic ablations in Sec.~\ref{subsec:heur_ablation}.
The optimal reference B1 and relaxation-based MICP(g) solved much fewer instances with longer runtimes on successful runs.

ST-RRT$^*$ found feasible trajectories quickly in \textit{rand} and especially \textit{iris}, but required more time in \textit{maze} and did not solve all instances there.
Note that its success rate curves reflect the time to the first feasible solutions.
However, when its runtime budget was matched to B3, success in \textit{maze} decreased considerably, although it still solved all \textit{iris} instances and most \textit{rand} instances.
This pattern is consistent with narrow maze passages making collision-free space-time connections harder to find through sampling within a limited runtime budget.

Zeta$^*$-SIPP solved more instances with the finer grid of $r$, while the coarser grid of $2r$ was faster whenever both succeeded.
Only one run with cell size $r$ timed out, while all other failures terminated before the time limit.
This suggests that conservative static occupancy and cell-based dynamic safe intervals limited success more than the runtime budget.

We further stress-tested B2 and B3 on another $150$ spatiotemporal planning instances with more dynamic obstacles and larger ST-GCSs (App.~\ref{app:st_stress_test}).
B3 ($\delta_\text{pos}$) showed better scalability than B2 ($\delta_\text{state}$), achieving higher success within the same budget and lower mean runtime on common-success instances.
This contrasted with B2's slightly lower median runtimes on the original benchmark, favoring $\delta_\text{pos}$ for larger ST-GCSs.

\begin{figure}[t]
    \centering
    \includegraphics[width=\linewidth]{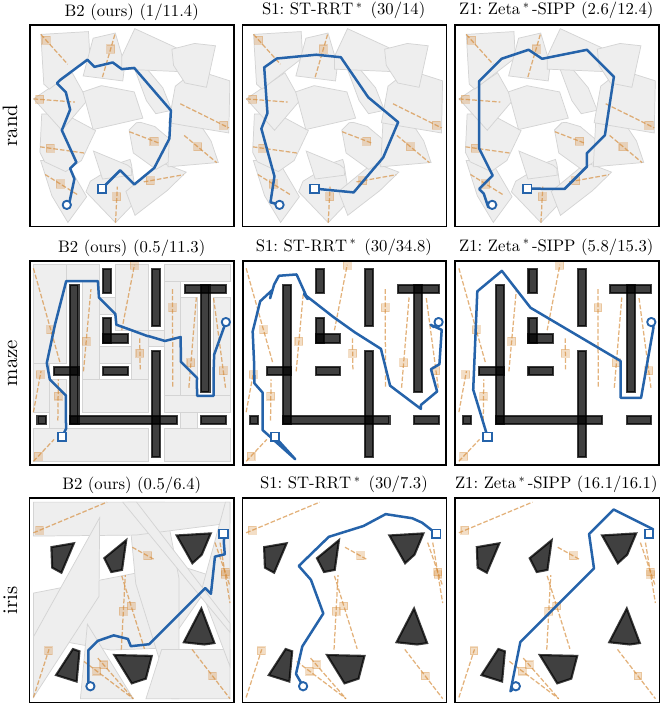}
    \caption{Qualitative spatiotemporal planning comparison. Each subplot title reports runtime and arrival time in seconds. Gray and black regions show the spatial GCS and static obstacles, respectively. Blue curves show robot trajectories; orange squares and dashed lines show moving obstacle motions. For S1 and Z1, the spatial GCS is visualized only in \textit{rand}, where it defines the explicit spatial collision-free space.}
    \label{fig:st_qualitative_comparison}
\end{figure}

\subsubsection{Solution Quality}\label{subsec:st_solution_quality}
The bottom row of Fig.~\ref{fig:st_performance_comparison} compares solution costs on common-success instances among planners with at least $33\%$ success in each domain.
This gives $15$, $9$, and $5$ common-success instances in \textit{rand}, \textit{maze}, and \textit{iris}, respectively.
B1 had the lowest median cost among the included ST-GCS solvers in each domain.
MICP(g) had a median cost close to B1 in \textit{iris}, but with a wider cost spread.
Among the faster BFS variants, B2 had lower median costs than B3 and B4 in \textit{rand} and \textit{maze}, and matched B3 in \textit{iris}.
B4 did not improve median cost over B3 despite its longer runtimes.

ST-RRT$^*$ benefited from the full $30$s runtime budget, returning lower-cost solutions than its runtime-matched variant ST-RRT$^*$(m) on every common-success instance in \textit{rand} and \textit{iris}.
However, its final-incumbent median cost remained higher than B2's in each domain.
The finer grid of $r$ gave Zeta$^*$-SIPP lower or equal costs than $2r$ on every common-success instance, alongside its higher success rate.
On the five common-success \textit{iris} instances, it achieved the lowest median cost among the included planners, slightly below B1.

In general, our BFS solvers achieved lower median solution costs than baselines in \textit{rand} and \textit{maze}, where all planners shared the same underlying space-time collision-free regions.
Even in \textit{iris}, where the ST-GCS solvers were restricted to the initial convex sets covering only about $90\%$ of the spatial free space, they remained competitive with ST-RRT$^*$ and Zeta$^*$-SIPP, which were not restricted to this cover.
Fig.~\ref{fig:st_qualitative_comparison} shows a qualitative comparison of B2, S1, and Z1 in each domain.

\subsection{Multi-Robot Motion Planning}
We ablated the proposed MRMP planners (Sec.~\ref{sec:mrmp}) on the MRMP instances in Sec.~\ref{subsec:mrmp_instances}, and compared their performance with other planners.
All ST-GCS-based MRMP planners share the same low-level BFS solver with $h_\text{max}$, inflation factor $\epsilon=8$, UB pruning, and the $\delta_\text{pos}$ dominance check.
We chose this configuration since it was empirically the most scalable option (Tab.~\ref{tab:heuristic_inflation} and App.~\ref{app:st_stress_test}), making it well suited to the large ECD-updated ST-GCSs commonly encountered in MRMP.

\begin{figure}[t]
    \centering
    \includegraphics[width=\linewidth]{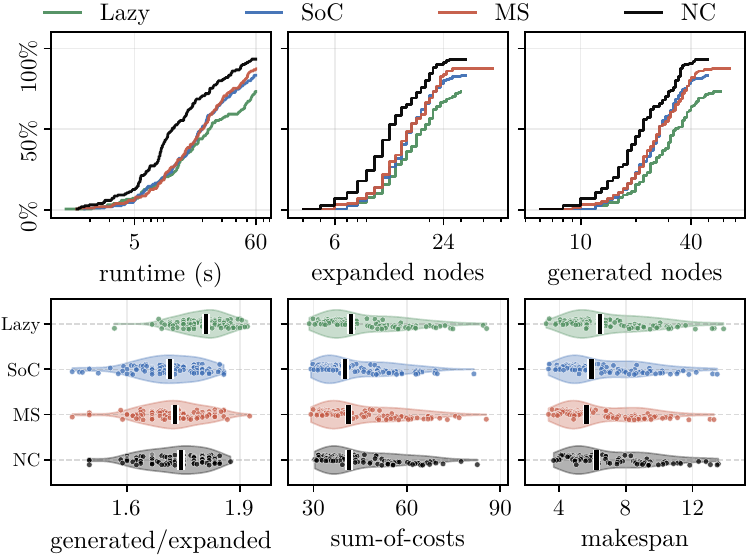}
    \caption{PBS expansion rules ablation on $150$ $10$-robot MRMP instances ($60$s budget). Top row shows success rate over runtime, expanded, and generated PBS nodes. Bottom row shows distributions of generated/expanded ratio, sum-of-costs, and makespan on common-success instances.}
    \label{fig:mrmp_pbs_node_expansion}
\end{figure}

\subsubsection{Benchmark Instances}\label{subsec:mrmp_instances}
We generated $150$ MRMP instances, with $50$ instances from each domain of \textit{rand}, \textit{maze}, and \textit{iris}.
Each instance contained queries for $n>1$ robots on a seeded base ST-GCS. 
Each query started at time $0$ with a velocity bound $\mathbf{v}_{\mathrm{lim}}=[1.0,1.0]$.
We sampled start and goal positions with pairwise $\ell_\infty$ separation of at least $2r$ between different robots, and retained instances whose independently planned trajectories conflicted. This emphasized inter-robot coordination instead of isolated single-robot planning.

\subsubsection{Ablation of PBS Expansion Rules}\label{subsec:pbs_node_expansion_ablation}
This ablation examines the node expansion rule in full-horizon PBS in Alg.~\ref{alg:wpbs}.
As introduced in Sec.~\ref{subsec:child-node-exp}, we compared the four rules of lazy evaluation (Lazy), the sum-of-costs (SoC) metric, the makespan (MS) metric, and the number of conflicts (NC) metric.
Recall that a PBS node is said to be generated once created by adding a new partial priority order, while it is expanded once the robots are replanned via \textsc{UpdateNode}.

Fig.~\ref{fig:mrmp_pbs_node_expansion} shows that NC offered the best scalability, solving more instances within shorter runtime budgets.
NC also had the lowest median counts of expanded and generated PBS nodes on common-success instances, indicating more efficient high-level search.
Lazy had the highest median generated/expanded ratio but the lowest success rate, suggesting that deferring replanning alone was insufficient to improve scalability.

\begin{figure}[t]
    \centering
    \includegraphics[width=\linewidth]{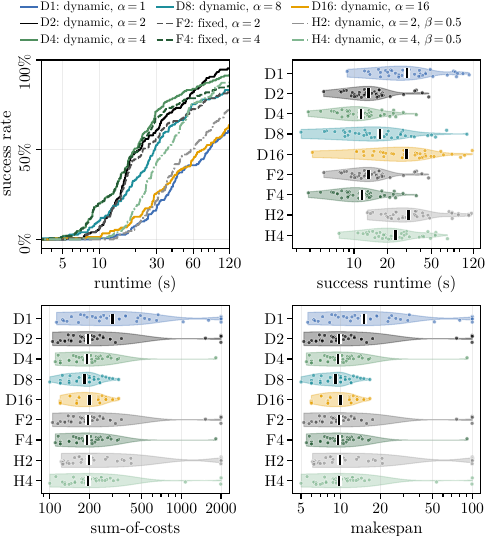}
    \caption{Windowed-PBS ablation on $150$ $20$-robot MRMP instances ($120$s budget). The distributions of success runtime, sum-of-costs, and makespan use common-success instances.}
    \label{fig:mrmp_windowed_coordination}
\end{figure}

\begin{figure*}[t]
    \centering
    \includegraphics[width=\linewidth]{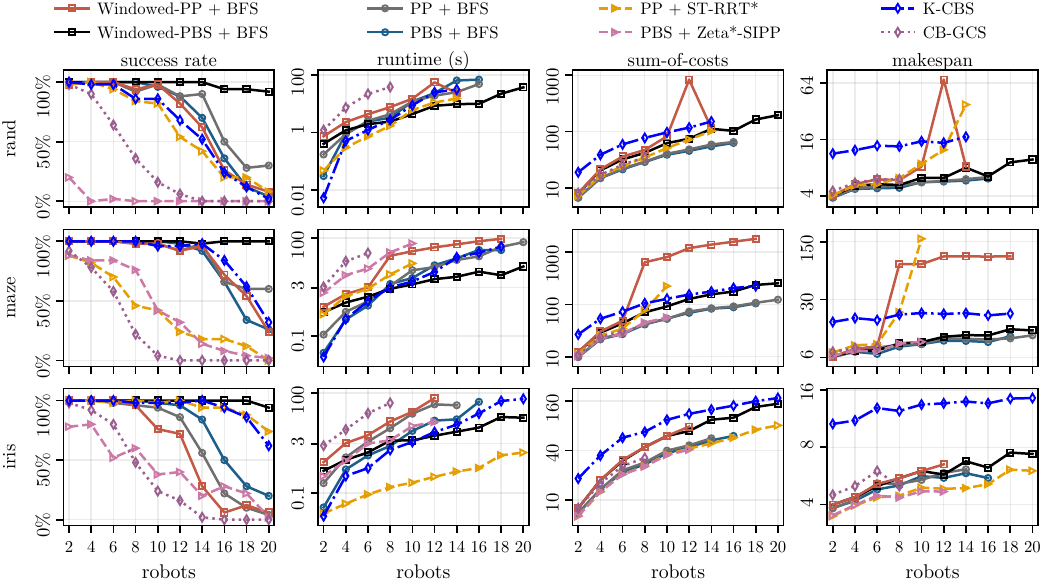}
    \caption{MRMP performance comparison ($120$s budget). Runtime, sum-of-costs, and makespan are plotted in log scale and report medians over common-success instances among planners with at least $33\%$ success rate in each domain and robot count.}
    \label{fig:mrmp_performance_comparison}
\end{figure*}

On common-success instances, SoC and MS yielded the lowest median values of their respective objectives.
NC remained competitive in both metrics, supporting its use as the default expansion rule.
This comparison measured sensitivity to the child-node expansion rule; sensitivity to robot indexing and tied branch order was not separately evaluated.

\subsubsection{Ablation of Windowed Coordination}\label{subsec:windowed_ablation}
This ablation isolated the nominal planning span $\Delta t_{\mathrm{plan}}=\alpha r/\|\mathbf{v}_{\mathrm{lim}}\|_\infty$ and the execution span $\Delta t_{\mathrm{exec}}=\beta\Delta t_{\mathrm{plan}}$ for windowed-PBS as described in Sec.~\ref{subsec:windowed_coord}.
We compared dynamic windows with $\alpha\in\{1,2,4,8,16\}$ and $\beta=1$.
At the more promising planning spans of $\alpha\in\{2,4\}$, we also evaluated fixed windows with $\beta=1$ and dynamic windows with $\beta=0.5$.
All variants used the NC child-node expansion rule.

Fig.~\ref{fig:mrmp_windowed_coordination} shows that moderate planning spans gave the best scalability, with D2 (dynamic window with $\alpha=2$ and $\beta=1$) achieving the highest success rate.
Shorter windows required more coordination calls, whereas longer windows made each call more expensive.
Dynamic window adjustment improved success relative to fixed windows at both tested spans, with similar median runtimes on common-success instances.

On common-success instances, dynamic adjustment preserved the solution quality of the corresponding fixed windows, while D8 (dynamic window with $\alpha=8$ and $\beta=1$) yielded the lowest median SoC and makespan.
Halving the execution span increased median runtime and reduced success at both tested planning spans without improving solution quality.
Overall, D2 gave the best scalability in this ablation, while D4 (dynamic window with $\alpha=4$ and $\beta=1$) achieved the lowest median runtime on common-success instances.

\subsubsection{Benchmark Planners}
We evaluated various MRMP planners on $n$-robot MRMP instances, with $150$ instances for each $n\in\{2,4,\ldots,20\}$ generated as in Sec.~\ref{subsec:mrmp_instances}.
Following Sec.~\ref{subsec:windowed_ablation}, the proposed Windowed-PBS+BFS used dynamic window with $\alpha=2$ and $\beta=1$ as its high-level coordinator for its best scalability.
We also compared several composed baselines with different high-level coordinators, namely, PP, PBS with NC node expansion rule, and windowed-PP with the same $\alpha=2$ and $\beta=1$.

To test the alternative low-level planners in Sec.~\ref{subsec:st_baselines}, we evaluated PBS+Zeta$^*$-SIPP with discretization of $r$, and PP+ST-RRT$^*$.
The latter used ST-RRT$^*$ as the single-robot solver in an anytime PP baseline.
In the first pass, robots were planned sequentially in a fixed priority order, with each robot treating previously planned trajectories, including their endpoint-staying portions, as moving obstacles.
After this pass produced a feasible joint solution, the remaining budget was used for improvement rounds that revisited the robots in the same order and replaced an incumbent trajectory only when ST-RRT$^*$ found a shorter trajectory that avoided all other robots' current trajectories.
We report its first feasible solution for runtime and its final incumbent solution for solution quality.

In addition, we compared with two standalone MRMP baselines. K-CBS~\citep{kottinger2022conflict} is a fast feasibility-oriented and sampling-based MRMP planner. CB-GCS~\citep{zhao2025cbgcs} similarly solves MRMP on GCSs under a fixed time-step representation.

\subsubsection{Scalability}\label{subsec:mrmp_scalability}
The success-rate column of Fig.~\ref{fig:mrmp_performance_comparison} shows that the proposed Windowed-PBS+BFS had the highest overall success rate.
Over the $n\ge 10$ instances, it achieved a success rate of $98.4\%$, compared with $69.4\%$ for the next-best baseline, K-CBS.
At $n=20$, its success rate was $95.3\%$, compared with $32.0\%$ for K-CBS.

These comparisons highlight the complementary roles of the low-level BFS solver, windowed planning, and local priority search.
BFS achieved higher overall success than ST-RRT$^*$ under PP and Zeta$^*$-SIPP under PBS.
Windowing alone was not sufficient. Windowed-PP+BFS used the same low-level solver and nominal window settings as Windowed-PBS+BFS, but its success rate declined sharply at larger team sizes.
PBS alone was also not sufficient. Full-horizon PBS+BFS solved all instances with $n\le 8$, but its success rate fell to $16.7\%$ at $n=20$.
These results support the proposed combination, which couples BFS with local priority search inside each window while committing only the next execution interval.

On common-success instances, full-horizon BFS solvers and K-CBS were faster at small team sizes, while Windowed-PBS+BFS achieved lower median runtimes than the other included planners in \textit{rand} and \textit{maze} as team size increased.
PP+ST-RRT$^*$ remained particularly strong in \textit{iris}, finding feasible solutions much faster than Windowed-PBS+BFS, but its success rate decreased at larger team sizes and was substantially lower in \textit{rand} and \textit{maze}.
PBS+Zeta$^*$-SIPP and CB-GCS also lost success at smaller team sizes and generally had longer median runtimes on the plotted common-success subsets.
Thus, Windowed-PBS+BFS provided the most consistent scalability across all three domains.

Search and trajectory optimization accounted for most of the Windowed-PBS+BFS runtime, while ECD became substantial and the share of other costs decreased at larger team sizes.
App.~\ref{app:mrmp_runtime_profile} gives the complete runtime breakdown and query densities.

\begin{figure}[t]
    \centering
    \includegraphics[width=\linewidth]{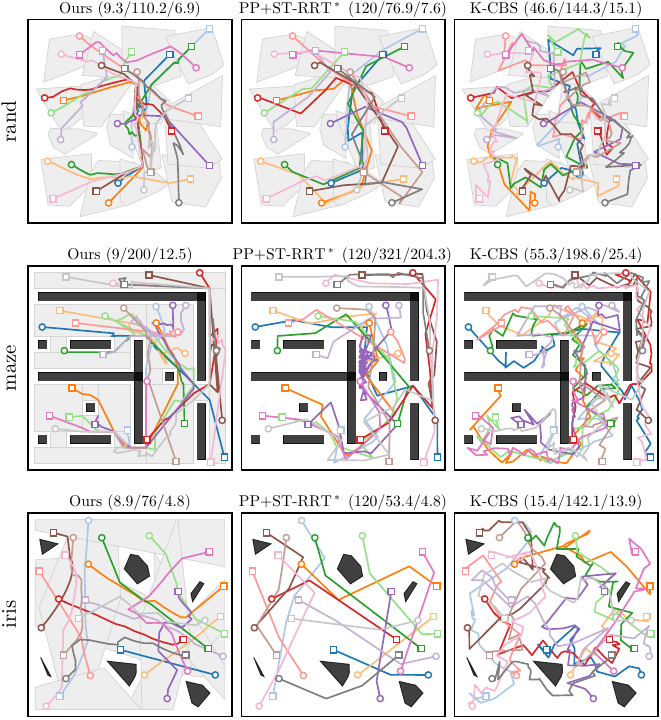}
    \caption{Qualitative MRMP comparison. Each subplot title reports the runtime, sum-of-costs, and makespan in seconds. Gray and black regions show the spatial GCS and static obstacles, respectively. Colors distinguish robots, and circles and squares mark starts and goals. For PP+ST-RRT$^*$ and K-CBS, the spatial GCS is visualized only in \textit{rand}, where it defines the explicit spatial collision-free space.}
    \label{fig:mrmp_qualitative_comparison}
\end{figure}

\subsubsection{Solution Quality}
The SoC and makespan columns of Fig.~\ref{fig:mrmp_performance_comparison} show that the proposed Windowed-PBS+BFS retained competitive solution quality while achieving the scalability reported above.
On common-success instances, full-horizon PBS+BFS and PP+BFS generally had lower median SoC and makespan.
Thus, windowed coordination sacrificed some SoC since it plans and commits only within a finite horizon, while keeping the makespan closer to the full-horizon BFS solvers.
Windowed-PP+BFS gave a similar tradeoff in \textit{iris}, but had much larger SoC and makespan in \textit{maze} for $n\ge 8$.
This further supported resolving local priority conflicts within each window.

In contrast, the feasibility-oriented K-CBS had substantially larger makespans across all domains and generally higher SoC than Windowed-PBS+BFS.
PP+ST-RRT$^*$ was a stronger solution-quality baseline in \textit{iris} after using the full runtime budget, achieving lower median SoC and makespan than Windowed-PBS+BFS.
As in Sec.~\ref{subsec:st_solution_quality}, this is consistent with the ST-GCS solvers being restricted to an IRIS-generated convex cover of only about $90\%$ of the spatial free space, while ST-RRT$^*$ could also explore the uncovered regions.
However, its quality advantage was domain-dependent. In \textit{rand} and \textit{maze}, its makespans grew more sharply with team size and exceeded those of Windowed-PBS+BFS despite the improvement rounds.
Fig.~\ref{fig:mrmp_qualitative_comparison} shows a qualitative comparison of Windowed-PBS+BFS, PP+ST-RRT$^*$, and K-CBS on $16$-robot MRMP instances with a $120$s runtime budget.
\section{Applications and Demonstrations}
This section complements the quantitative comparisons above with several demonstration scenarios.
We first describe a trajectory-optimization postprocessing step for smoothing the piecewise-linear trajectories produced by our planner, then demonstrate the solutions on large-scale simulated problems and real-robot deployments.

\subsection{Global Trajectory Optimization}\label{subsec:trajopt}

\begin{figure}[t]
    \centering
    \includegraphics[width=\linewidth]{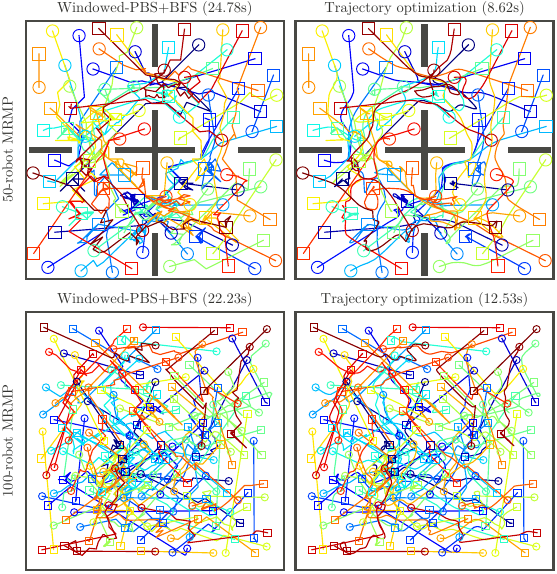}
    \caption{Large-scale 2D coordination in a four-room workspace for $50$ robots (top) and an empty workspace for $100$ robots (bottom). Left and right columns show MRMP and smoothed solutions, respectively, with runtimes in parentheses. Hollow circles/squares mark starts/goals.}
    \label{fig:demo_large_2d}
\end{figure}

The piecewise-linear trajectories produced by our MRMP planner can be smoothed by solving a Quadratic Program (QP) over spatial positions at fixed time samples.
The objective penalizes velocity variation, displacement from the reference trajectories, and differences in the rate of change of this displacement across consecutive intervals.
The constraints preserve start and goal positions and enforce component-wise velocity limits.
For larger instances, a rolling-horizon variant optimizes successive blocks of samples.
App.~\ref{app:trajectory_optimization} gives the sampling scheme, objective, and constraints.

The set-containment constraints keep entire optimized segments inside collision-free convex sets.
Pairwise separation constraints use a common separating normal at both endpoints of each interval, ensuring separation throughout the interval under linear interpolation.
Together, these constraints ensure that the optimized trajectories remain continuously collision-free for disk or spherical robots.

\begin{figure}[t]
    \centering
    \includegraphics[width=\linewidth]{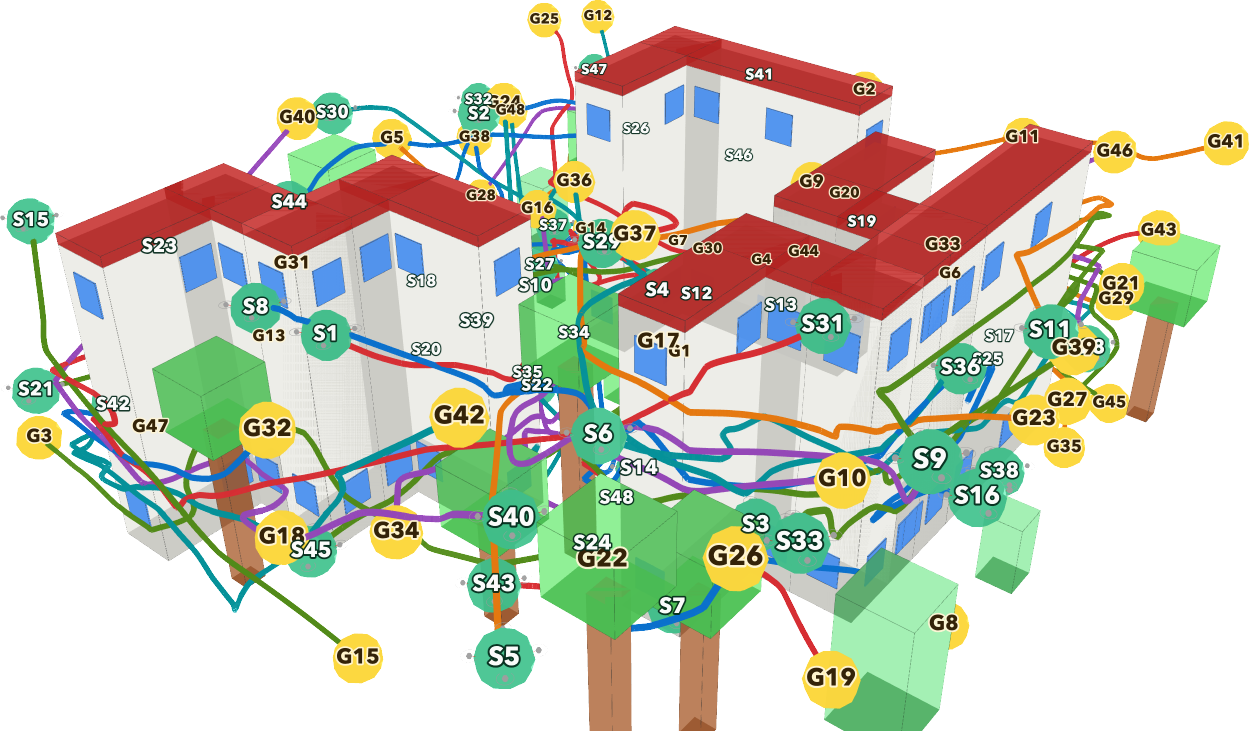}
    \caption{48-UAV MRMP in a cluttered village with trees and buildings. The trajectory-optimized result is shown and solved in 12.78s after our planner found a solution in 172.62s.}
    \label{fig:uav-demo}
\end{figure}

\subsection{Large-Scale Coordination}

We demonstrate in large-scale MRMP instances that the proposed Windowed-PBS+BFS scales to confined and highly congested workspaces, with the trajectory optimization in Sec.~\ref{subsec:trajopt} for post-processing smoothing.

\begin{figure*}[t]
    \centering
    \includegraphics[width=0.98\linewidth]{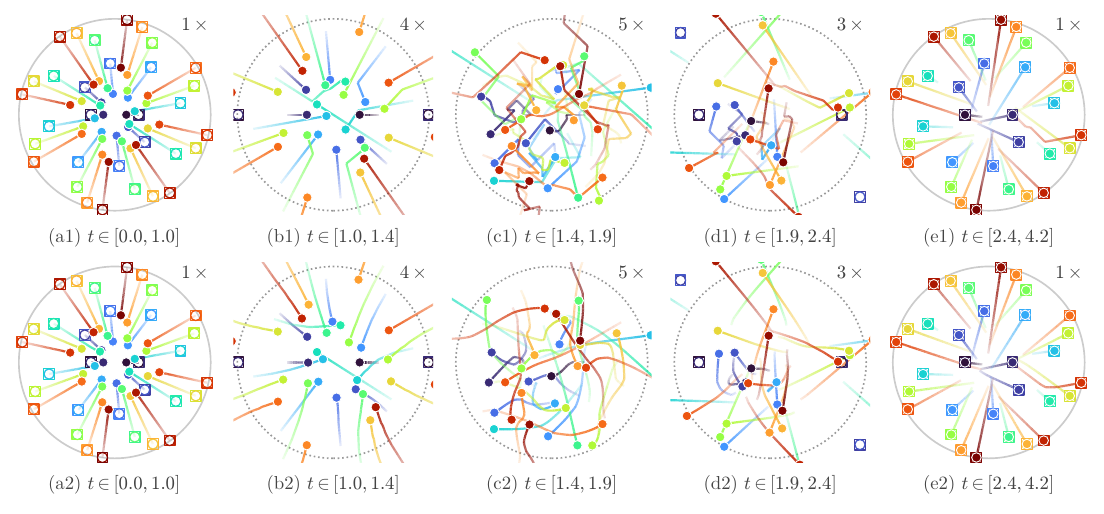}
    \caption{Top-down snapshots of a 32-UAV position exchange between antipodal points on a sphere. Top and bottom rows show our MRMP solution and trajectory-optimized result solved in 15.67s and 5.35s, respectively. Hollow circles/squares mark starts/targets, filled circles mark window-end positions, and dotted circles and labels mark zoom bounds and factors.}
    \label{fig:demo_sphere_3d_traj}
\end{figure*}

Fig.~\ref{fig:demo_large_2d} shows two planar coordination settings with complementary sources of difficulty.
The $50$-robot and $100$-robot instances have query densities of $25.86\%$ and $18.00\%$, respectively, both much higher than the benchmark's highest median query density of $11.53\%$ reported in Tab.~\ref{tab:mrmp-wpbs-runtime-breakdown}.
In the $50$-robot four-room instance, the workspace funnels traffic through narrow inter-room passages, so difficulty comes from geometric bottlenecks and the need to sequence many robots through shared doorways.
In the $100$-robot empty-workspace instance, the independent trajectories of $96$ robots are involved in initial collisions, with $93$ robots in the largest connected conflict component, so difficulty comes from dense pairwise interaction rather than static obstacles.
These results show that the planner remains practical both when difficulty comes from geometric bottlenecks and when it comes from dense pairwise robot interactions.

Fig.~\ref{fig:demo_sphere_3d_traj} further stresses coordination with a $32$-UAV position-exchange problem, with starts on a sphere and goals at their antipodes.
Planning takes place in an obstacle-free enclosing cube represented by one convex polyhedron, with robot occupancies defined in Sec.~\ref{subsec:trajectory_occupancy}.
The independent straight paths all pass through the sphere center, creating dense conflicts that the coordinator resolves before trajectory optimization smooths the resulting paths.
Fig.~\ref{fig:uav-demo} tests a highly challenging $48$-UAV instance in a 3D village with dense buildings and trees~\citep{marcucci2024fast}.
The static obstacle collision-free space in the scene is decomposed into an ST-GCS with $119$ convex sets, similarly to the \textit{maze} domain.
The MRMP queries are sampled as long-distance trips across this fragmented free space, so many UAVs must simultaneously route around vertical clutter while avoiding one another in narrow local passages.

We solved all the large-scale MRMP instances in this subsection without instance-specific parameter tuning.
We used the low-level BFS solver with $h_\text{max}$, $\epsilon=8$, UB pruning, and the $\delta_\text{pos}$ dominance check, following the scalability results in App.~\ref{app:st_stress_test}.
The high-level PBS coordinator used NC child-node expansion based on Sec.~\ref{subsec:pbs_node_expansion_ablation}.
We retained dynamic windows with $\alpha=2$ and $\beta=1$, which achieved the highest success rate in Sec.~\ref{subsec:windowed_ablation}.
Guided by the window ablation in Fig.~\ref{fig:mrmp_windowed_coordination}, we regard the coordination-window span factor $\alpha$ as the primary parameter for further problem-specific adjustment, including for large-scale MRMP.
It controls how far ahead robot interactions are coordinated and the size of reservation-updated ST-GCSs, balancing per-call cost against the number of replanning calls.

\subsection{Mobile Robots Deployment}

We validate our MRMP planner in an open indoor environment with a system of $9$ differential drive mobile robots. 
The system is centrally controlled by an Intel Core Ultra 5 125U laptop with 16GB RAM, ROS2 Humble~\citep{ros2_humble}, and \textit{Vicon}~\citep{vicon_tracker_software} for robot localization.
Each robot has a bounding radius of 7cm and is exposed to a velocity command $(v,\omega)$, with linear velocity $v\in [0, 0.6]$ m/s and angular velocity $\omega\in[-2.5, 2.5]$ rad/s.
Note that this velocity limit is set conservatively smaller than the physical robot limits for system robustness.

We construct a rearrange task consisting of a sequence of MRMP queries within a 3m$\times$3m bounding square.
The goal positions in each query consecutively form a configuration of five letters: S, T, G, C, and S.
All the robots start simultaneously from their previous goal configurations.
Specifically, they initialize and terminate in a $3\times 3$ grid configuration.
We first compute piecewise-linear solution trajectories via Windowed-PBS+BFS (in $5.97$s), and then smooth them via the trajectory optimization described in Sec.~\ref{subsec:trajopt} (in $1.59$s).
The safe clearance between robots is set to $20$cm during planning, which is $6$cm larger than the robot diameter.

\begin{figure}[t]
\centering
\includegraphics[width=\linewidth]{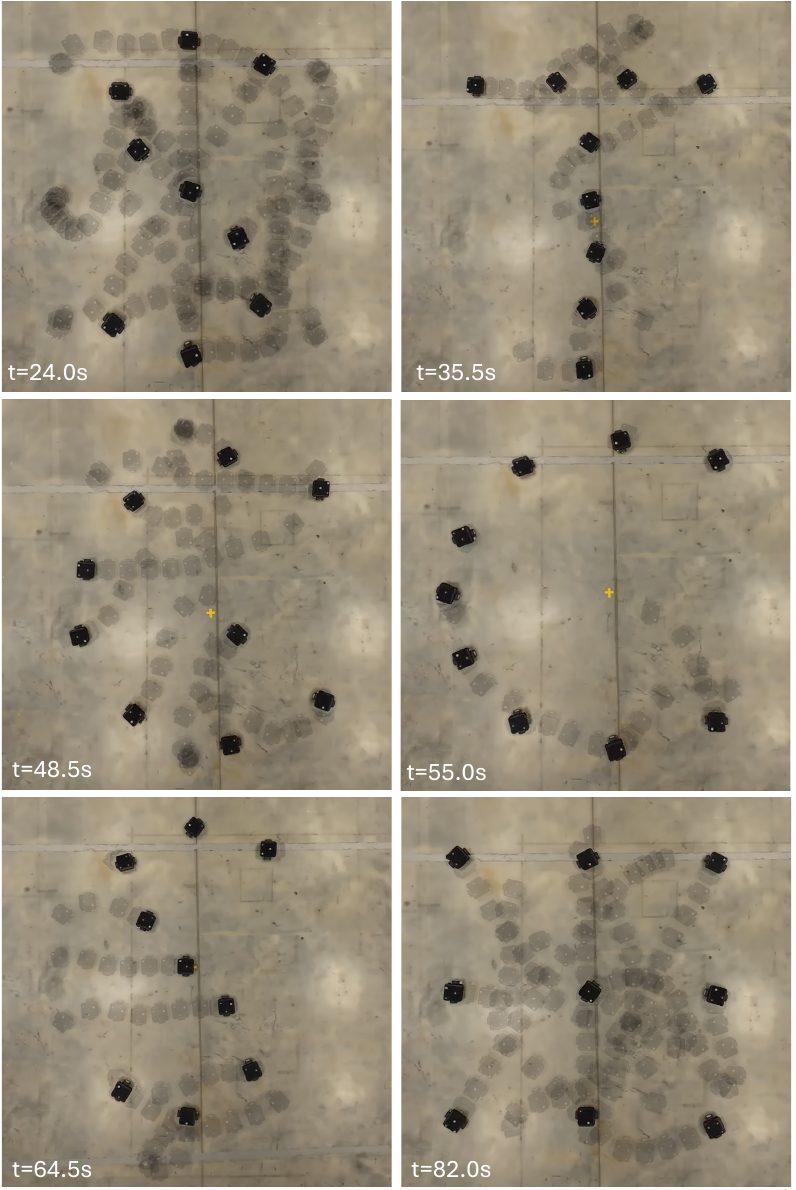}
\caption{Real-robot demonstration of rearrange task for the letters S, T, G, C, S. Opaque robots mark captured poses, and translucent ghosts show preceding motion traces.}
\label{fig:real_exp}
\end{figure}

During online execution, the smoothed trajectories are discretized with a $0.05$s time step into time-indexed waypoints, and the tracker precomputes segment-wise reference linear and angular velocities from consecutive samples.
All robots receive a common start time and evaluate their own trajectory at a system-wide shared elapsed time, preserving the simultaneity encoded by the offline solution.
We use a simple trajectory tracking controller for the robots with a 60-Hz frequency.
In each control cycle, the tracker first identifies the active trajectory segment whose time interval contains the system time. 
The final published velocity command combines the active segment's reference velocities as feedforward terms with body-frame feedback corrections computed from the interpolated desired pose error.
In our deployments, the conservative velocity limits and additional planned inter-robot clearance provided enough margin for the observed tracking errors, and the robots completed the rearrangement without inter-robot collisions.
Fig.~\ref{fig:real_exp} demonstrates the actual robot trajectories during the rearrange task.

\section{Conclusions}
We presented ST-GCS for time-optimal planning among moving obstacles, combining path-based search with ECD trajectory reservations for prioritized and windowed MRMP.
Experiments demonstrate improved scalability and competitive solution quality, including coordination of up to $100$ robots within a minute.

Future work will explore richer motion models, including nonholonomic and kinodynamic constraints, using geometrically certified convex envelopes for trajectory reservations.
A key goal is to establish corresponding planning guarantees while retaining the scalability of the ST-GCS framework.
We will also investigate adaptive or approximate decompositions for efficient online replanning and window-selection rules with completeness guarantees.

\begin{funding}
This work was supported by the NSERC under grant number RGPIN2020-06540 and a CFI JELF award.
\end{funding}


\bibliographystyle{plainnat}
\bibliography{ref}

\titleformat{\section}
  {\normalfont\large\bfseries}
  {Appendix~\thesection}{0.5em}{}
\clearpage
\appendix
\counterwithin{table}{section}
\counterwithin{figure}{section}
\renewcommand{\thesection}{\Alph{section}}
\makeatletter
\renewcommand{\@seccntformat}[1]{%
  \ifstrequal{#1}{section}{Appendix~}{}\csname the#1\endcsname\quad}
\makeatother
\renewcommand{\thesubsection}{\Alph{section}.\arabic{subsection}}
\renewcommand{\thetable}{\Alph{section}\arabic{table}}
\renewcommand{\thefigure}{\Alph{section}\arabic{figure}}

\section{Theoretical Results}
\label{app:theoretical_proofs}

This appendix collects supporting lemmas and full proofs for the theoretical results in the main text.

\subsection{Lemma~\ref{lemma:no_set_revisit} Proof (No Set Revisitation)}
\label{app:no_set_revisit}

\begin{myproof}
Let the feasible trajectory $\tau$ conditioned on $\pi$ have entry and exit states $(\mathbf{x}_k,\mathbf{y}_k)$ for the $k$-th vertex occurrence, where $\mathbf{x}_k,\mathbf{y}_k\in X_{v_k}$, and continuity gives $\mathbf{y}_k=\mathbf{x}_{k+1}$ for $k=1,\ldots,l-1$. Since $v_i=v_j$, both $\mathbf{x}_i$ and $\mathbf{y}_j$ lie in the same convex set $X_{v_i}$. Hence the straight shortcut segment $\mathbf{x}_i\rightarrow \mathbf{y}_j$ lies entirely in $X_{v_i}$.

It remains to check time monotonicity and velocity feasibility for this shortcut segment. For each original segment $k=i,\ldots,j$, let $\Delta t_k$ and $\Delta\mathbf{p}_k$ denote its elapsed time and spatial displacement as defined in Sec.~\ref{subsec:st_trajectory_representation}, indexed here by vertex occurrence. Each original segment satisfies Eqn.~(\ref{eqn:st_local_motion}). Summing the velocity bounds over $k=i,\ldots,j$ gives
\[-\mathbf{v}_{\mathrm{lim}}\sum_{k=i}^{j}\Delta t_k \preceq \sum_{k=i}^{j}\Delta \mathbf{p}_k\preceq \mathbf{v}_{\mathrm{lim}} \sum_{k=i}^{j}\Delta t_k.\]
By continuity, the sums are
\[\sum_{k=i}^{j}\Delta t_k=\mathbf{y}_j.t - \mathbf{x}_i.t, \quad \sum_{k=i}^{j}\Delta \mathbf{p}_k=\mathbf{y}_j.\mathbf{p} - \mathbf{x}_i.\mathbf{p}.\] Since $\sum_{k=i}^{j}\Delta t_k\geq 0$, the shortcut segment $\mathbf{x}_i\rightarrow \mathbf{y}_j$ satisfies both time monotonicity and the velocity bounds in Eqn.~(\ref{eqn:st_local_motion}).

Replacing the portion of the trajectory $\tau$ from $\mathbf{x}_i$ to $\mathbf{y}_j$ by this shortcut yields a feasible trajectory on $\pi'$, while preserving continuity with the preceding segment and, if $j<l$, the following segment. By the cost definition in Eqn.~(\ref{eqn:traj_cost}), the shortcut cost $\mathbf{y}_j.t - \mathbf{x}_i.t=\sum_{k=i}^{j}\Delta t_k$ equals the cost of the removed portion. Thus, the shortened path has a feasible trajectory with no larger cost. Repeating this shortcut operation removes all repeated vertices.
\end{myproof}

\subsection{Search Optimality and Sound Pruning}
\label{app:search_guarantees}

We prove the search guarantee under sound pruning and show that $\delta_\text{set}$ satisfies the required soundness condition.

\subsubsection{Theorem~\ref{thm:cost_optimality} Proof (\texorpdfstring{$\epsilon$}{epsilon}-Optimality)}
\label{app:cost_optimality}

\begin{myproof}
Let $c^*$ be the optimal trajectory cost.
Both key evaluations satisfy $N.f\leq\epsilon J^*(N.\pi)$ and $N.f=N.g$ at a goal node. For separate evaluation, admissibility and $N.g\geq0$ give
\begin{align*}
N.f&=N.g+\epsilon h(N.\pi)\\
&\leq\epsilon J^*(N.\pi)-(\epsilon-1)N.g
\leq\epsilon J^*(N.\pi),
\end{align*}
and $h=0$ at a goal. For joint evaluation, these properties follow from Lemma~\ref{lem:joint_key}, proved in App.~\ref{app:joint_key}.

Because the graph is finite and Lemma~\ref{lemma:no_set_revisit} lets us restrict attention to simple prefixes, the search space is finite. If $\tau_\text{ub}\neq\emptyset$ and $c_\text{ub}\leq \epsilon c^*$, returning $\tau_\text{ub}$ is already $\epsilon$-optimal. Any goal returned before the incumbent has passed the upper-bound test, so its key, and hence its cost, is smaller than $c_\text{ub}\leq\epsilon c^*$. 
Thus, it remains to consider the complementary case. Either no incumbent is available or $c_\text{ub}>\epsilon c^*$. In both cases, upper-bound pruning keeps every node with key at most $\epsilon c^*$.

We maintain the following invariant: until termination, OPEN contains a representative node $N$ with $J^*(N.\pi)=c^*$ and $N.f\leq \epsilon c^*$. Initially this holds for the root because $\langle v_s\rangle$ prefixes every feasible trajectory and the key bound gives $N_r.f\leq\epsilon J^*(\langle v_s\rangle)=\epsilon c^*$. If OPEN expands a non-representative node, the representative remains in OPEN unless the expanded node is a goal; in that case the minimum-key rule and equality of the goal key and cost imply that the returned trajectory has cost at most $\epsilon c^*$.

It remains to show that the invariant is preserved when a representative node $N$ is expanded. If $N.v=v_g$, then $N.\tau$ is an optimal solution. Otherwise, take an optimal $N.\pi$-consistent solution of cost $c^*$, and let $w$ be its next vertex, giving the feasible prefix $\pi'=N.\pi\oplus\langle w\rangle$. Then \textsc{PathOptimize} creates a child $M$ with $M.\pi=\pi'$ and $J^*(M.\pi)=c^*$. The key bound gives $M.f\leq\epsilon J^*(M.\pi)=\epsilon c^*$.
By the case assumption above, upper-bound pruning keeps $M$. If $M$ is inserted into OPEN, the invariant follows. If $M$ is pruned by a sound dominance check, some $M'\in S(w)$ satisfies $J^*(M'.\pi)\leq J^*(M.\pi)=c^*$. Optimality of $c^*$ gives $J^*(M'.\pi)=c^*$, and the same key bound gives $M'.f\leq\epsilon c^*$. Thus $M'$ is a valid representative if it is in OPEN; if it has already been expanded, the same argument applied at its expansion transfers the representative to an OPEN descendant or to a retained node dominating such a descendant. Likewise, if a retained representative is later removed from $S(w)$, soundness implies that the newly accepted node has $J^*=c^*$ and key at most $\epsilon c^*$. Sound dominance pruning therefore transfers, but never eliminates, all representatives before termination.

The incumbent-return case was handled above. In the complementary case, the invariant prevents OPEN from becoming empty before termination, so Alg.~\ref{alg:st_search} must return a goal node $N_g$ expanded from OPEN. The invariant and the minimum-key rule give $N_g.f\leq\epsilon c^*$; since the goal key equals its cost, $c(N_g.\tau)=N_g.g=N_g.f\leq\epsilon c^*$. Setting $\epsilon=1$ gives cost optimality.
\end{myproof}

\subsubsection{Lemma~\ref{lem:esc_sound} Proof (\texorpdfstring{$\delta_\text{set}$}{delta-set} Soundness)}
\label{app:dominance_soundness}

\begin{myproof}
Let $N'\in S(N.v)$ satisfy Eqn.~(\ref{eqn:delta_esc}). Consider any $N.\pi$-consistent solution trajectory $\bar{\tau}$. Let $\mathbf{x}$ and $\mathbf{y}$ be the states where $\bar\tau$ enters and exits $X_{N.v}$, respectively. By construction, $\mathbf{x}\in R_N$. Moreover, the segment $\mathbf{x}\rightarrow\mathbf{y}$ is part of $\bar{\tau}$ within $X_{N.v}$, so $\mathbf{y}\in C(\mathbf{x})$. Eqn.~(\ref{eqn:delta_esc}) gives $C(\mathbf{x})\subseteq C(N'.\mathbf{x})$, and therefore $\mathbf{y}\in C(N'.\mathbf{x})$.

Since $N'.\mathbf{x}, \mathbf{y} \in X_{N.v}$, the straight segment $N'.\mathbf{x}\rightarrow\mathbf{y}$ remains in $X_{N.v}$ and satisfies the time-monotonicity and velocity constraints. Replacing the zero-duration terminal segment of $N'.\tau$ with this segment, and then concatenating the suffix of $\bar\tau$ after $\mathbf{y}$, yields a solution trajectory with the same time cost as $\bar\tau$. Since this holds for every $N.\pi$-consistent solution trajectory, we have $J^*(N'.\pi)\leq J^*(N.\pi)$. Thus, $N'$ dominates $N$.
\end{myproof}

\subsection{Heuristic Admissibility Lemmas and Proofs}
\label{app:heuristic_admissibility}

We state and prove the admissibility of $h_{\mathrm{mot}}$, $h_{\mathrm{tri}}$, $h_{\mathrm{tab}}$, and $h_{\mathrm{max}}$, as defined in Sec.~\ref{subsec:st_planning_heur}. We then establish the admissibility of the heuristic reuse described in Sec.~\ref{subsec:mrmp_heuristic_reuse}.

\begin{lemma}\label{lem:hmot_admissible}
The heuristic $h_\text{mot}$ is admissible.
\end{lemma}

\phantomsection\label{proof:hmot_admissible}
\begin{myproof}
Consider any $N.\pi$-consistent trajectory $\bar\tau$. Let $\mathbf{x} \in I_N$ be the state through which $\bar\tau$ passes before continuing beyond the prefix. The prefix portion of $\bar\tau$ has cost at least $N.g$ as $N.g$ is the optimal cost conditioned on $N.\pi$. The remaining portion from $\mathbf{x}$ to $\mathbf{p}_g$ satisfies the the velocity limits, so its cost is at least $\rho(\{\mathbf{x}\}, \mathbf{p}_g)$, and therefore at least $\rho(I_N, \mathbf{p}_g)$. Thus, $c(\bar\tau) \geq N.g + h_\text{mot}$ holds for any $N.\pi$-consistent solution trajectory $\bar\tau$.
\end{myproof}

\begin{lemma}\label{lem:htri_admissible}
The heuristic $h_{\text{tri}}$ is admissible.
\end{lemma}

\phantomsection\label{proof:htri_admissible}
\begin{myproof}
Fix any $N.\pi$-consistent trajectory. Its suffix induces a sequence of triplets. For each triplet, the actual segment through the middle convex set is feasible for the corresponding triplet subproblem, so the triplet cost is at most the actual segment cost. Omitting query-specific start triplets only further relaxes the cost. Summing these inequalities over the suffix lower bounds the optimal remaining solution cost. Since $h_{\text{tri}}$ takes the minimum over all suffixes in the relaxed triplet graph, it is at most the remaining cost of any $N.\pi$-consistent trajectory beyond the prefix represented by $N.\pi$.
\end{myproof}

\begin{lemma}\label{lem:htab_admissible}
The heuristic $h_\text{tab}$ is admissible.
\end{lemma}

\phantomsection\label{proof:htab_admissible}
\begin{myproof}
If $N.v \in \operatorname{adj}(v_g)$, or $N.v=v_s$ and $\operatorname{adj}(v_s)\cap
\operatorname{adj}(v_g)\cap V\ne\emptyset$, then $h_{\mathrm{tab}}(N.\pi) = \rho(I_N, \mathbf{p}_g)$ is admissible by the same argument as Lemma~\ref{lem:hmot_admissible}. Now consider a non-goal node $N$ for which the table-based expression applies, and any $N.\pi$-consistent solution trajectory $\bar\tau$. Let $v_p\in \operatorname{adj}(v_g)\cap V$ be the input-graph vertex through which the vertex path of $\bar\tau$ reaches $v_g$, and let $u\in\operatorname{adj}(v_p)\cap V$ be the input-graph vertex preceding $v_p$ on that path. When $N.v=v_s$, we have $\operatorname{adj}(v_s)\cap \operatorname{adj}(v_g)\cap V=\emptyset$, so every $N.\pi$-consistent solution must cross at least one interface between input-graph vertices. If $I_N$ is an input-graph interface, the portion from $I_N$ to $X_{v_p}$ has cost at least $d(I_N,v_p)$. If $I_N$ is query-specific because $N.v = v_s$ or $N.\pi = \langle v_s, N.v\rangle$, then $\hat{d}_N(v_p)$ starts from a later input-graph interface and omits only nonnegative cost needed to reach that interface, so it remains a lower bound. The final portion from $X_u\cap X_{v_p}$ to $\mathbf{p}_g$ satisfies the velocity limits and is lower-bounded by $\rho(X_u\cap X_{v_p},\mathbf{p}_g)$. Thus, the corresponding term in the minimization defining $h_\text{tab}$ is no larger than the remaining cost of $\bar\tau$. Since $h_\text{tab}$ minimizes over all such pairs $(u,v_p)$, it is also no larger than the remaining cost of $\bar\tau$, for any $N.\pi$-consistent solution trajectory.
\end{myproof}

\begin{lemma}\label{lem:hmax_admissible}
The heuristic $h_\text{max}$ is admissible.
\end{lemma}

\phantomsection\label{proof:hmax_admissible}
\begin{myproof}
Each component heuristic is no larger than the optimal remaining solution cost from $N.\pi$ by Lemmas~\ref{lem:hmot_admissible}, \ref{lem:htri_admissible}, and \ref{lem:htab_admissible}. The maximum of several lower bounds is still a lower bound, so $h_\text{max}$ is admissible.
\end{myproof}

\begin{lemma}\label{lem:mrmp_heuristic_reuse}
On an ECD-updated graph $G^\sharp$, the reuse rule in Sec.~\ref{subsec:mrmp_heuristic_reuse} preserves admissibility of $h_\text{tri}$ and $h_\text{tab}$.
\end{lemma}

\phantomsection\label{proof:mrmp_heuristic_reuse}
\begin{myproof}
Fix $h\in\{h_{\mathrm{tri}},h_{\mathrm{tab}}\}$ and an $N.\pi$-consistent solution $\bar\tau$. Its cost up to entry into $N.v$ is at least $N.g$, so it suffices to show that the reused
heuristic lower-bounds the remaining cost. Removing reservations gives a feasible base continuation from the same entry state because each refined set is contained in its base set, and consecutive segments within a common base set can be joined by a straight segment with the same endpoint times and velocity bounds.

At the root or first query edge, the continuation starts at $\mathbf{x}_s$. If the prefix ends in $u\to v$ with $\operatorname{base}(u)\ne\operatorname{base}(v)$, its entry state lies in $X_{\operatorname{base}(u)}\cap X_{\operatorname{base}(v)}$. In both cases the base bound applies by Lemmas~\ref{lem:htri_admissible} and~\ref{lem:htab_admissible}.

For a same-base transition, the zero case is immediate. Otherwise $\bar\tau$ must leave $\operatorname{base}(v)$ before reaching $v_g$. Let $\operatorname{base}(v)\to w$ be its first base exit. The remaining cost from this exit is at least $h(\langle\ldots,\operatorname{base}(v),w\rangle)$ on $G_0^\sharp$ by the same lemmas. Travel from the current refined interface to this exit has nonnegative cost. Taking the minimum over all exits therefore gives $N.g+h(N.\pi)\le c(\bar\tau)$. This holds for every consistent solution, and the heuristic is zero at the goal.
\end{myproof}

\subsection{Lemma~\ref{lem:joint_key} Proof (Joint Search-Key Bounds)}
\label{app:joint_key}

\begin{myproof}
Fix a generated node $N$ with terminal vertex $v=N.v$. Every feasible trajectory $\tau$ in Eqn.~(\ref{eqn:joint_key}) has $c(\tau)\geq N.g$, since \textsc{PathOptimize} minimizes the prefix cost. For a non-goal node, its terminal entry state satisfies $\mathbf{x}_v\in I_N$, so $\rho(\{\mathbf{x}_v\},\mathbf{p}_g)\geq\rho(I_N,\mathbf{p}_g)=h_{\mathrm{mot}}(N.\pi)$. Minimizing their sum proves $N.f\geq N.g+h_{\mathrm{mot}}(N.\pi)$ when $\epsilon=1$.

To establish the upper bound, consider any $N.\pi$-consistent solution trajectory $\bar\tau$, and let $\mathbf{x}$ be its entry state in $X_v$. Truncate $\bar\tau$ at $\mathbf{x}$ and give its terminal segment zero duration. This yields a feasible prefix $\tau$ for Eqn.~(\ref{eqn:joint_key}), with $\mathbf{x}_v=\mathbf{y}_v=\mathbf{x}$. Let $C=c(\tau)$ and let $R=c(\bar\tau)-C$ be the remaining duration, including motion in $X_v$. Summing the velocity bounds along this remaining trajectory gives
\[
\rho(\{\mathbf{x}\},\mathbf{p}_g)\leq R.
\]
For the root, the same construction uses the zero-duration prefix at $\mathbf{x}_s$. Consequently, for any $\epsilon\geq1$,
\begin{align*}
N.f&\leq C+\epsilon\rho(\{\mathbf{x}\},\mathbf{p}_g)\\
&\leq C+\epsilon R\\
&=\epsilon c(\bar\tau)-(\epsilon-1)C\\
&\leq\epsilon c(\bar\tau)-(\epsilon-1)N.g,
\end{align*}
where the last inequality uses $C\geq N.g$. Taking the infimum over all $N.\pi$-consistent solutions yields
\[
N.f\leq\epsilon J^*(N.\pi)-(\epsilon-1)N.g
\leq\epsilon J^*(N.\pi).
\]
If no such solution exists, $J^*(N.\pi)=+\infty$ and the upper bound is immediate. Setting $\epsilon=1$ proves $N.f\leq J^*(N.\pi)$.

At a goal node, $\mathbf{x}_v\in X_{v_g}$ implies $\mathbf{x}_v.\mathbf{p}=\mathbf{p}_g$, so the continuation term is zero. A prefix-cost optimum already has $\mathbf{y}_v=\mathbf{x}_v$; hence the joint optimum equals $N.g$ for every $\epsilon\geq1$. Since $h_{\mathrm{mot}}=0$ at a goal, the lower bound above also holds there.

The separate prefix-only solve preserves $N.g$, $N.\tau$, and the earliest arrival time used by $\delta_{\mathrm{set}}$, so its soundness proof is unchanged. The bound $N.f\leq\epsilon J^*(N.\pi)$ and equality of the goal key and cost are precisely the key properties used in App.~\ref{app:cost_optimality}. Thus, its frontier and incumbent-pruning arguments establish Theorem~\ref{thm:cost_optimality} for joint evaluation as well.
\end{myproof}

\subsection{Theorem~\ref{thm:pp_complete} Proof (Completeness of PP)}
\label{app:pp_complete}

\begin{myproof}
Fix any total order $\pmb{\sigma}$ and the BFS inflation factor $\epsilon$. The bounds defined before Theorem~\ref{thm:pp_complete} have initial value $B_0(\epsilon)=t_0$ and satisfy
\[
B_k(\epsilon)=\epsilon(B_{k-1}(\epsilon)+C_{\max}^*)-(\epsilon-1)\min_j t_{s,j}.
\]
Since $B_0(\epsilon)\geq\min_j t_{s,j}$ and $\epsilon\geq1$, we have $B_{k-1}(\epsilon)+C_{\max}^*\leq B_k(\epsilon)\leq B_n(\epsilon)$ for $1\leq k\leq n$.

Consider robot $i=\sigma_k$ and suppose each earlier robot $\sigma_j$ arrived by $B_j(\epsilon)$. Let $T$ be the maximum of $t_0$ and their actual arrival times, taking $T=t_0$ for the first robot. Then $T\leq B_{k-1}(\epsilon)$. Following the waiting construction of \citet[Theorem~3]{ma2019searching}, consider a trajectory that waits at $\mathbf{p}_{s,i}$ until $T$ and then follows the route of duration $C_i^*$ from condition~(\ref{wellform:feasibility}). The wait is safe from time zero since every higher-priority robot planned with $\mathbf{p}_{s,i}$ reserved throughout the horizon and conditions~(\ref{wellform:separation}) and~(\ref{wellform:staying}) protect the endpoint stays. At $T$ all higher-priority robots are at their goals. The route avoids these goals on $\widehat G_i$, and time invariance after $t_0$ permits shifting its departure to $T$. It reaches $\mathbf{p}_{g,i}$ by $T+C_i^*\leq B_k(\epsilon)\leq t_{\max}$ and can stay there until $t_{\max}$.

This trajectory is feasible for the current query with its prescribed start time $t_{s,i}$ and the actual higher-priority reservations. By the assumed $\epsilon$-optimality of BFS, its returned arrival time satisfies
\begin{align*}
t_{g,i}
&\leq t_{s,i}+\epsilon(T+C_i^*-t_{s,i})\\
&\leq \epsilon(B_{k-1}(\epsilon)+C_{\max}^*)-(\epsilon-1)\min_j t_{s,j}=B_k(\epsilon),
\end{align*}
where the second inequality uses $\epsilon\geq1$ and $t_{s,i}\geq\min_j t_{s,j}$. BFS may return a trajectory that moves concurrently with higher-priority robots. Its motion and goal stay avoid their reserved trajectories, and their endpoint reservations protect its pre-start stay. By induction every call succeeds and the returned endpoint-augmented trajectories are pairwise collision-free.
\end{myproof}

\subsection{Theorem~\ref{thm:pbs_complete} Proof (Completeness of PBS)}
\label{app:pbs_complete}

\begin{myproof}
Let $d_N(i)$ be the number of robots in a longest priority chain ending at $i$ in node $N$. We show that every stored trajectory avoids the other robots' endpoints, robot $i$ arrives by $B_{d_N(i)}(\epsilon)$, and the endpoint-augmented trajectories of comparable robots are collision-free. At the root every depth is one. Each robot plans independently on $\widehat G_i$, so the first-step argument of App.~\ref{app:pp_complete} gives arrival by $B_1(\epsilon)$.

Suppose the properties hold at $N$. A collision can only involve incomparable robots, so adding either orientation preserves acyclicity. Every priority chain in $N$ remains a chain in the child $N'$, hence $d_{N'}(k)\geq d_N(k)$ for every robot $k$. Every inherited trajectory thus meets its new arrival bound and still avoids the other robots' endpoints. If the new relation is $i\prec j$, only $j$ and its descendants can acquire new ancestors or have an ancestor whose trajectory changes. \textsc{UpdateNode} processes these robots in topological order.

When a robot $k$ of depth $d=d_{N'}(k)$ is processed, all its ancestors have their final trajectories for this child. Their depths are at most $d-1$, so they arrive by $B_{d-1}(\epsilon)$. If a collision requires replanning, the argument of App.~\ref{app:pp_complete} applies with these ancestors as the higher-priority robots. Waiting until the later of $t_0$ and their last arrival gives a feasible continuation, and BFS on $G^{(k)}_{N'}$ returns a trajectory arriving by $B_d(\epsilon)\leq B_n(\epsilon)\leq t_{\max}$. Otherwise the inherited trajectory already avoids every ancestor. Thus every child update succeeds and preserves all three properties.

Each branch orders a previously incomparable pair, so a branch has at most $n(n-1)/2$ such steps. At every node with a collision, both children can be generated. The finite search therefore reaches a node with pairwise collision-free endpoint-augmented trajectories and returns a feasible solution.
\end{myproof}

\section{Additional Experiments}
\label{app:additional_experiments}
\makeatletter
\setlength{\@fptop}{0pt}
\makeatother

This appendix reports the heuristic-inflation sensitivity, offline heuristic precomputation cost, and MRMP runtime profile supporting the conclusions in Sec.~\ref{sec:exp}.

\begin{table}[t]
\centering
\caption{BFS heuristic-inflation sensitivity ($30$s budget). Runtime and cost gap report mean values of successful runs.}
\label{tab:heuristic_inflation}
\small
\setlength{\tabcolsep}{5pt}
\begin{tabular}{@{}cccccccc@{}}
\toprule
 & \multirow{2}{*}{$\epsilon$} & \multicolumn{2}{c}{\shortstack{Succ. rate (\%)}} & \multicolumn{2}{c}{\shortstack{Runtime (s)}} & \multicolumn{2}{c}{\shortstack{Cost gap (\%)}} \\
\cmidrule(lr){3-4}\cmidrule(lr){5-6}\cmidrule(lr){7-8}
 & & $\delta_\text{state}$ & $\delta_\text{pos}$ & $\delta_\text{state}$ & $\delta_\text{pos}$ & $\delta_\text{state}$ & $\delta_\text{pos}$ \\
\midrule
\multirow{6}{*}{\rotatebox{90}{$h_\text{mot}$}} & 1 & $100$ & $84.0$ & $1.515$ & $1.066$ & $0.230$ & $3.720$ \\
 & 2 & $100$ & $100$ & $1.201$ & $1.050$ & $1.894$ & $9.580$ \\
 & 4 & $100$ & $99.3$ & $1.265$ & $1.079$ & $4.178$ & $12.68$ \\
 & 8 & $100$ & $99.3$ & $1.480$ & $1.172$ & $8.159$ & $18.46$ \\
 & 16 & $100$ & $99.3$ & $1.783$ & $1.277$ & $13.10$ & $22.86$ \\
 & 32 & $100$ & $99.3$ & $2.038$ & $1.391$ & $15.09$ & $27.69$ \\
\midrule
\multirow{6}{*}{\rotatebox{90}{$h_\text{tri}$}} & 1 & $100$ & $81.3$ & $2.589$ & $1.218$ & $0.275$ & $4.592$ \\
 & 2 & $100$ & $99.3$ & $2.127$ & $1.166$ & $0.889$ & $9.270$ \\
 & 4 & $100$ & $100$ & $1.514$ & $1.102$ & $3.496$ & $13.62$ \\
 & 8 & $100$ & $100$ & $1.374$ & $1.035$ & $6.805$ & $17.14$ \\
 & 16 & $100$ & $100$ & $1.334$ & $1.014$ & $8.926$ & $21.57$ \\
 & 32 & $100$ & $100$ & $1.279$ & $0.994$ & $9.202$ & $22.79$ \\
\midrule
\multirow{6}{*}{\rotatebox{90}{$h_\text{tab}$}} & 1 & $100$ & $82.7$ & $2.464$ & $1.165$ & $0.275$ & $4.446$ \\
 & 2 & $100$ & $99.3$ & $1.646$ & $1.071$ & $1.947$ & $10.42$ \\
 & 4 & $100$ & $100$ & $1.327$ & $1.034$ & $6.053$ & $16.15$ \\
 & 8 & $100$ & $100$ & $1.237$ & $0.977$ & $8.338$ & $20.29$ \\
 & 16 & $100$ & $100$ & $1.176$ & $0.957$ & $9.670$ & $22.02$ \\
 & 32 & $100$ & $100$ & $1.174$ & $0.955$ & $10.66$ & $24.79$ \\
\midrule
\multirow{6}{*}{\rotatebox{90}{$h_\text{max}$}} & 1 & $100$ & $88.7$ & $1.393$ & $1.009$ & $0.246$ & $3.307$ \\
 & 2 & $100$ & $100$ & $0.945$ & $0.916$ & $2.149$ & $9.523$ \\
 & 4 & $100$ & $100$ & $0.901$ & $0.905$ & $5.689$ & $14.26$ \\
 & 8 & $100$ & $100$ & $0.887$ & $0.903$ & $9.766$ & $19.68$ \\
 & 16 & $100$ & $100$ & $0.888$ & $0.926$ & $15.83$ & $24.63$ \\
 & 32 & $100$ & $100$ & $0.954$ & $0.948$ & $17.58$ & $28.65$ \\
\midrule
\multirow{6}{*}{\rotatebox{90}{\textit{joint}}} & 1 & $100$ & $86.0$ & $2.298$ & $1.413$ & $0.232$ & $3.658$ \\
 & 2 & $100$ & $100$ & $1.879$ & $1.337$ & $1.586$ & $9.576$ \\
 & 4 & $100$ & $99.3$ & $1.882$ & $1.363$ & $4.324$ & $13.77$ \\
 & 8 & $100$ & $99.3$ & $2.322$ & $1.462$ & $13.32$ & $23.48$ \\
 & 16 & $100$ & $99.3$ & $2.795$ & $1.591$ & $21.28$ & $31.20$ \\
 & 32 & $99.3$ & $99.3$ & $3.359$ & $1.787$ & $32.03$ & $38.78$ \\
\bottomrule
\end{tabular}
\end{table}

\subsection{BFS Heuristic Inflation}
\label{app:heuristic_inflation}

We evaluated heuristic inflation for the four BFS heuristics and \textit{joint} (Sec.~\ref{subsec:joint_key}) on all spatiotemporal instances from Sec.~\ref{subsec:st_instances}.
All variants used UB pruning with either $\delta_\text{state}$ or $\delta_\text{pos}$.
We tested $\epsilon\in\{1,2,4,8,16,32\}$ with a $30$s runtime budget per instance.
Tab.~\ref{tab:heuristic_inflation} reports success rates over all instances and mean runtime and cost gap over each configuration's successful runs.
The cost gap was computed as $100(c_\epsilon/c_\text{bk}-1)\%$, where $c_\epsilon$ was the returned solution cost and $c_\text{bk}$ was the lowest found for the same instance across all configurations.

All configurations with $\delta_\text{state}$ solved every instance except \textit{joint} at $\epsilon=32$, which solved all but one ($99.3\%$).
With $\delta_\text{pos}$, success rates increased from $81.3$--$88.7\%$ at $\epsilon=1$ to $99.3$--$100\%$ at $\epsilon\geq2$.

$h_\text{max}$ with $\epsilon=8$ achieved the lowest mean runtime among all tested configurations under each dominance check, but moderate inflation captured most of the runtime benefit.
With $\delta_\text{state}$, increasing $\epsilon$ from $1$ to $2$ reduced mean runtime from $1.39$s to $0.94$s, with a mean cost gap of $2.15\%$.
Increasing $\epsilon$ to $8$ lowered mean runtime to $0.89$s but raised the mean cost gap to $9.77\%$.
With $\delta_\text{pos}$, increasing $\epsilon$ from $2$ to $8$ reduced mean runtime by only $1.4\%$ ($0.916$s to $0.903$s), while the mean cost gap rose from $9.52\%$ to $19.68\%$.
Beyond $\epsilon=8$, both mean runtime and cost gap increased for $h_\text{max}$ under either check.

The inflation factor minimizing mean runtime varied across heuristics.
With both checks, $h_\text{mot}$ and \textit{joint} were fastest at $\epsilon=2$, whereas $h_\text{tri}$ and $h_\text{tab}$ reached lowest mean runtimes at $\epsilon=32$.
Mean cost gaps increased with inflation for all variants, and stronger inflation can also increase runtime.
For example, increasing $\epsilon$ from $2$ to $32$ raised mean runtime of \textit{joint} with $\delta_\text{state}$ from $1.88$s to $3.36$s and its mean cost gap from $1.59\%$ to $32.03\%$.
Such slowdowns can arise when inflation overemphasizes optimistic heuristic estimates and triggers additional \textsc{PathOptimize} on infeasible extensions.

\subsection{BFS Heuristic Offline Precomputation}
\label{app:heuristic_precomputation}

This scaling study used base ST-GCSs of varying sizes that differed from those in Sec.~\ref{subsec:st_instances}.
Fig.~\ref{fig:heur_precomputation_scaling} (left) reports median preprocessing runtime at each observed base ST-GCS size $|V_0|+|E_0|$.
Throughout the sweep, $h_\text{tri}$ took less than $0.2$s, whereas $h_\text{tab}$ peaked at about $2.3$ hours.
In the evaluated sweep, $h_\text{tri}$ scaled approximately linearly through small local triplet computations, whereas $h_\text{tab}$ exhibited approximately exponential growth as it repeatedly optimizes competing prefixes for each starting interface.

\subsection{BFS Solver Stress Test}
\label{app:st_stress_test}

\begin{figure}[t]
\centering
\includegraphics[width=\linewidth]{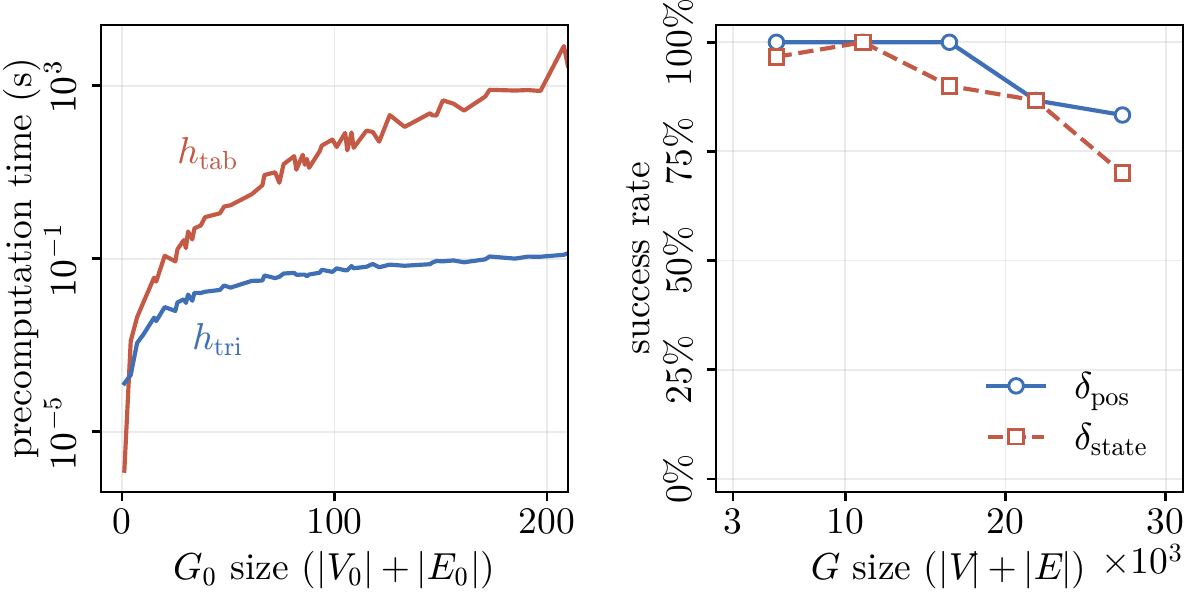}
\caption{Left: median offline precomputation runtime for $h_\text{tab}$ and $h_\text{tri}$ versus base ST-GCS size. Right: spatiotemporal stress-test success rates for $\delta_\text{pos}$ and $\delta_\text{state}$ versus ST-GCS size. Each point reports success within $30$s over one size interval of $30$ instances, plotted at the interval midpoint.}
\label{fig:heur_precomputation_scaling}
\end{figure}

We stress-tested the approximate dominance checks by adding more moving obstacles to the $150$ instances in Sec.~\ref{subsec:st_instances}, retaining their base ST-GCSs, queries, and planning horizons while preserving a validated collision-free trajectory for each instance.
The enlarged ST-GCSs were balanced across five size intervals spanning $3{,}000\leq |V|+|E|\leq30{,}000$, with $10$ instances per domain in each interval.
Both variants used $h_\text{max}$, $\epsilon=8$, UB pruning, and a $30$s runtime budget.
In Fig.~\ref{fig:heur_precomputation_scaling} (right), $\delta_\text{pos}$ matched or exceeded the success rate of $\delta_\text{state}$ across all five size intervals.
The gap was widest in the largest interval, where $\delta_\text{pos}$ solved $83.3\%$ of instances compared with $70.0\%$ for $\delta_\text{state}$.
On their $133$ common-success instances, $\delta_\text{pos}$ also reduced mean runtime from $7.43$s to $6.60$s, supporting its use on larger ST-GCSs.

\subsection{MRMP Runtime Profile}
\label{app:mrmp_runtime_profile}

Tab.~\ref{tab:mrmp-wpbs-runtime-breakdown} gives the Windowed-PBS+BFS solving profile for the MRMP benchmark in Sec.~\ref{subsec:mrmp_scalability}.
Query density is the fraction of the spatial GCS area covered by the union of all start and goal robot footprints (Eqn.~(\ref{eqn:hypercube_occ})).
Overlaps are counted only once in both the footprint union and the spatial GCS union.
Trajectory optimization was the largest runtime component for $n\ge4$, while the ECD share increased with team size.
At $n=20$, the reported shares were $23.6\%$ for search, $42.9\%$ for trajectory optimization, and $25.9\%$ for ECD.

\section{Global Trajectory Optimization Formulation}
\label{app:trajectory_optimization}

This appendix details the smoothing procedure summarized in Sec.~\ref{subsec:trajopt}.
The proposed MRMP planner produces piecewise-linear trajectories with fixed vertex paths on ST-GCS.
As a postprocessing step, we can smooth these trajectories by solving a Quadratic Program (QP) over sampled spatial positions along the trajectories at fixed time samples.
Given $n$ robot trajectories $\mathcal{T}^0=\{\tau_i^0\}_{i=1}^{n}$, we choose a shared ordered sample sequence
$\mathcal{K}=\langle t_1,\ldots,t_l\rangle$, with $t_1<\cdots<t_l$, formed from a uniform time grid together with the original trajectory knot times.
We will use $k$ to denote the index of the fixed time sample $t_k$.

\begin{table}[t]
\centering
\caption{The runtime profile of our Windowed-PBS+BFS for $n$-robot MRMP ($120$\,s budget). Query density and total runtime are medians. Component-wise runtime percentages is its median runtime divided by the total runtime. Opt. indicates convex optimization in \textsc{PathOptimize}. Search includes the runtime of both BFS and PBS without Opt.}
\label{tab:mrmp-wpbs-runtime-breakdown}
\small
\setlength{\tabcolsep}{3.5pt}
\begin{tabular}{rcccccc}
\toprule
\multicolumn{1}{c}{\multirow{2}{*}{$n$}} & \multicolumn{1}{c}{\multirow{2}{*}{\begin{tabular}[c]{@{}c@{}}Query\\Density (\%)\end{tabular}}} & \multicolumn{1}{c}{\multirow{2}{*}{\begin{tabular}[c]{@{}c@{}}Runtime\\Total (s)\end{tabular}}} & \multicolumn{4}{c}{Runtime Percentages (\%)} \\
\cmidrule(lr){4-7}
\multicolumn{1}{c}{} & \multicolumn{1}{c}{} & \multicolumn{1}{c}{} & \rule[-0.7ex]{0pt}{2.8ex}Search & Opt. & ECD & Others \\
\midrule
2 & 1.15 & 0.49 & 22.4 & 37.9 & 0.0 & 39.7 \\
4 & 2.34 & 1.08 & 26.2 & 42.6 & 6.0 & 25.2 \\
6 & 3.46 & 1.92 & 25.7 & 44.0 & 11.6 & 18.7 \\
8 & 4.67 & 2.96 & 26.6 & 44.2 & 14.3 & 14.9 \\
10 & 5.81 & 4.43 & 24.5 & 45.1 & 17.3 & 13.1 \\
12 & 7.04 & 7.16 & 24.5 & 44.5 & 20.5 & 10.5 \\
14 & 8.12 & 9.14 & 23.1 & 45.2 & 22.1 & 9.6 \\
16 & 9.27 & 13.4 & 21.9 & 45.7 & 23.8 & 8.6 \\
18 & 10.39 & 18.8 & 21.3 & 45.3 & 24.9 & 8.5 \\
20 & 11.53 & 21.7 & 23.6 & 42.9 & 25.9 & 7.6 \\
\bottomrule
\end{tabular}
\end{table}

We write $\tau_i^0(t)$ for the space-time state obtained by piecewise-linear interpolation along $\tau_i^0$ at time $t$.
Let $\bar{\mathbf{p}}_{i,k}=\tau_i^0(t_k).\mathbf{p}$ be the reference position of robot $i$ at time $t_k$.
The program keeps the times fixed and optimizes only the sampled spatial positions $\mathbf{p}_{i,k}\in\mathbb{R}^{m}$.
Similarly to Sec.~\ref{subsec:mrmp_def}, we assume a common radius $r$ for the robots.
For compactness, we define the sampled velocity and velocity-variation term as follows:
\begin{align*}
\mathbf{v}_{i,k}(\mathbf{p}) &=
\frac{\mathbf{p}_{i,k+1}-\mathbf{p}_{i,k}}{t_{k+1}-t_k},\\
\mathbf{g}_{i,k}(\mathbf{p}) &=
\frac{\mathbf{p}_{i,k+1}-\mathbf{p}_{i,k-1}}{t_{k+1}-t_{k-1}}
-
\frac{\mathbf{p}_{i,k}-\mathbf{p}_{i,k-2}}{t_k-t_{k-2}}.
\end{align*}
Let $\Delta\mathbf{p}_{i,k}=\mathbf{p}_{i,k}-\bar{\mathbf{p}}_{i,k}$ denote the displacement from the reference position.
For smooth motions, we penalize a velocity variation term $J_s$ across neighboring samples:
\begin{align}
J_s(\mathbf{p}) &= \sum_{i=1}^{n}\sum_{k=3}^{l-1}
    \|\mathbf{g}_{i,k}(\mathbf{p})\|_2^2. \notag
\end{align}
To let the smoothed trajectories close to the references, we penalize a spatial deviation term $J_d$:
\begin{align}
J_d(\mathbf{p}) &= \sum_{i=1}^{n}\sum_{k=1}^{l}
    \|\Delta\mathbf{p}_{i,k}\|_2^2. \notag
\end{align}
To reduce jitter in the trajectories, we also introduce an objective term $J_j$ to penalize differences in the rate of change of the displacement from the reference position between consecutive intervals.
Applying the same finite-difference operator to $\Delta\mathbf{p}$ gives:
\begin{align}
J_j(\mathbf{p}) &= \sum_{i=1}^{n}\sum_{k=2}^{l-1}
    \|\mathbf{v}_{i,k}(\Delta\mathbf{p})
      -\mathbf{v}_{i,k-1}(\Delta\mathbf{p})\|_2^2. \notag
\end{align}

Let $\mathcal{F}_i$ denote the samples at which robot $i$ is fixed at its initial or terminal state.
For each interval $[t_k,t_{k+1}]$, let $v_{i,k}$ be the ST-GCS vertex traversed by the input piecewise-linear trajectory at the interval midpoint.
We formulate the trajectory optimization as:
\begin{subequations}\label{eqn:trajopt}
\begin{align}
\min_{\{\mathbf{p}_{i,k}\}}\quad
& \lambda_c J_s(\mathbf{p})+\lambda_d J_d(\mathbf{p})+\lambda_j J_j(\mathbf{p})
    \notag\\
\textbf{s.t.}\quad
& \mathbf{p}_{i,k}=\bar{\mathbf{p}}_{i,k},
    \label{eqn:trajopt:fixed}\\
& -\mathbf{v}_{\mathrm{lim}}\preceq\mathbf{v}_{i,k}(\mathbf{p})
    \preceq\mathbf{v}_{\mathrm{lim}},
    \label{eqn:trajopt:vel}\\
& (\mathbf{p}_{i,k},t_k),\,(\mathbf{p}_{i,k+1},t_{k+1})\in X_{v_{i,k}},
    \label{eqn:trajopt:cell}\\
& \mathbf{n}_{ij,k}^{\top}(\mathbf{p}_{i,k}-\mathbf{p}_{j,k})\geq2r,
    \label{eqn:trajopt:pairwise}\\
& \mathbf{n}_{ij,k}^{\top}(\mathbf{p}_{i,k+1}-\mathbf{p}_{j,k+1})\geq2r,
    \label{eqn:trajopt:interval}
\end{align}
\end{subequations}
where, for every robot, Eqn.~(\ref{eqn:trajopt:fixed}) is enforced for $k\in\mathcal{F}_i$ and Eqns.~(\ref{eqn:trajopt:vel})--(\ref{eqn:trajopt:cell}) are enforced for $1\leq k\leq l-1$.
Eqns.~(\ref{eqn:trajopt:pairwise}) and (\ref{eqn:trajopt:interval}) are enforced for every robot pair $1\leq i<j\leq n$, with $1\leq k\leq l-1$.
The three nonnegative weights $\lambda_c$, $\lambda_d$, and $\lambda_j$ tune the relative emphasis of the corresponding objective terms.
Eqn.~(\ref{eqn:trajopt:fixed}) preserves the prescribed start and goal positions.
Eqn.~(\ref{eqn:trajopt:vel}) enforces the same component-wise velocity limits constraint as Eqn.~(\ref{eqn:st_local_motion}) in the original formulation.
Eqn.~(\ref{eqn:trajopt:cell}) constrains both endpoints of each optimized segment to lie in the same set $X_{v_{i,k}}$ of the original trajectory segment.
Since each optimized segment is a linear interpolation in space-time between the two endpoints, the convexity of $X_{v_{i,k}}$ keeps the entire optimized segment inside this collision-free set.
Eqns.~(\ref{eqn:trajopt:pairwise})--(\ref{eqn:trajopt:interval}) use one common separating normal at both endpoints of each interval.
Let $\bar{\mathbf{d}}^{*}_{ij,k}$ be the point of minimum Euclidean norm on the relative-position segment joining $\bar{\mathbf{p}}_{i,k}-\bar{\mathbf{p}}_{j,k}$ and $\bar{\mathbf{p}}_{i,k+1}-\bar{\mathbf{p}}_{j,k+1}$.
We set $\mathbf{n}_{ij,k}=\bar{\mathbf{d}}^{*}_{ij,k}/\|\bar{\mathbf{d}}^{*}_{ij,k}\|_2$ and hold this normal fixed for both endpoint constraints.
The relative position varies affinely between the endpoints, so its projection onto this unit normal remains at least $2r$ throughout the interval.
For strict separation, we use the prescribed clearance $R>2r$ in both endpoint constraints.
Thus, Eqns.~(\ref{eqn:trajopt:pairwise})--(\ref{eqn:trajopt:interval}) guarantee continuous pairwise separation for disk or spherical robots.

With ST-GCS containment in Eqn.~(\ref{eqn:trajopt:cell}) and interval pairwise separation in Eqns.~(\ref{eqn:trajopt:pairwise})--(\ref{eqn:trajopt:interval}), the resulting trajectories are collision-free throughout each interval.
For larger instances, we improve scalability with a sequential rolling-horizon variant that optimizes a short block of consecutive time samples, fixes the accepted prefix, advances the block, and repeats until all samples have been processed. 
This decomposition keeps each local trajectory-optimization QP small and makes the full problem easier to solve, while retaining the same objective terms and ensuring that the optimized trajectories remain continuously collision-free.

\end{document}